\documentclass[sw,crcready]{iosart2x}

\usepackage{color,soul}

\usepackage{amssymb}
\usepackage{svg}
\usepackage{subfig}

\usepackage{framed}
\usepackage{multirow}
\usepackage{multicol}
\usepackage{booktabs}

\PassOptionsToPackage{hyphens}{url}

\usepackage{colortbl}

\usepackage{xcolor}

\usepackage{dirtytalk}

\usepackage[most]{tcolorbox} 
\definecolor{light_gray}{gray}{0.98}
\definecolor{tan_grey}{HTML}{efeee8}
\definecolor{purple_1}{HTML}{d8ced4}
\definecolor{purple_2}{HTML}{a687a9}
\definecolor{purple_3}{HTML}{805388}
\definecolor{purple_4}{HTML}{450256}
\definecolor{black}{HTML}{000000}
\definecolor{dark_purple}{HTML}{450256}

\newtcolorbox{zitat}[2][]{%
    colback=tan_grey,
    grow to right by=-5mm,
    grow to left by=-5mm, 
    boxrule=0pt,
    boxsep=0pt,
    breakable,
    enhanced jigsaw,
    borderline west={4pt}{0pt}{dark_purple},
    title={#2\par},
    colbacktitle={tan_grey},
    coltitle={black},
    fonttitle={\large\bfseries},
    attach title to upper={},
    #1,
}
\pubyear{0000}
\volume{0}
\firstpage{1}
\lastpage{1}

\begin{document}

\begin{frontmatter}

\title{Is Neuro-Symbolic AI Meeting its Promise in Natural
Language Processing? A Structured Review}
\runtitle{NeSy for NLP: Review}


\begin{aug}
\author{\inits{K.}\fnms{Kyle} \snm{Hamilton}\ead[label=e1]{kyle.i.hamilton@mytudublin.ie}%
\thanks{Corresponding author. \printead{e1}.}}
\author{\inits{A.}\fnms{Aparna} \snm{Nayak}\ead[label=e2]{aparna.nayak@tudublin.ie}}
\author{\inits{B.}\fnms{Bojan} \snm{Božić}\ead[label=e3]{bojan.bozic@tudublin.ie}}
\author{\inits{L.}\fnms{Luca} \snm{Longo}\ead[label=e4]{luca.longo@tudublin.ie}}
\address{SFI Centre for Research Training in Machine Learning, School of Computer Science, \orgname{Technological University Dublin}, \cny{Republic of Ireland}\printead[presep={\\}]{e1,e2,e3,e4}}
\end{aug}


\begin{abstract}
Advocates for Neuro-Symbolic Artificial Intelligence (NeSy) assert that combining deep learning with symbolic reasoning will lead to stronger AI than either paradigm on its own. As successful as deep learning has been, it is generally accepted that even our best deep learning systems are not very good at abstract reasoning. And since reasoning is inextricably linked to language, it makes intuitive sense that Natural Language Processing (NLP), would be a particularly well-suited candidate for NeSy. We conduct a structured review of studies implementing NeSy for NLP, with the aim of answering the question of whether NeSy is indeed meeting its promises: reasoning, out-of-distribution generalization, interpretability, learning and reasoning from small data, and transferability to new domains. We examine the impact of knowledge representation, such as rules and semantic networks, language structure and relational structure, and whether implicit or explicit reasoning contributes to higher promise scores. We find that systems where logic is compiled into the neural network lead to the most NeSy goals being satisfied, while other factors such as knowledge representation, or type of neural architecture do not exhibit a clear correlation with goals being met. We find many discrepancies in how reasoning is defined, specifically in relation to human level reasoning, which impact decisions about model architectures and drive conclusions which are not always consistent across studies. Hence we advocate for a more methodical approach to the application of theories of human reasoning as well as the development of appropriate benchmarks, which we hope can lead to a better understanding of progress in the field. We make our data and code available on github for further analysis.\footnote{\url{https://github.com/kyleiwaniec/neuro-symbolic-ai-systematic-review}}
\end{abstract}


\end{frontmatter}








\section{Introduction}\label{sec:introduction}

\begin{figure}
\includegraphics[scale=0.3]{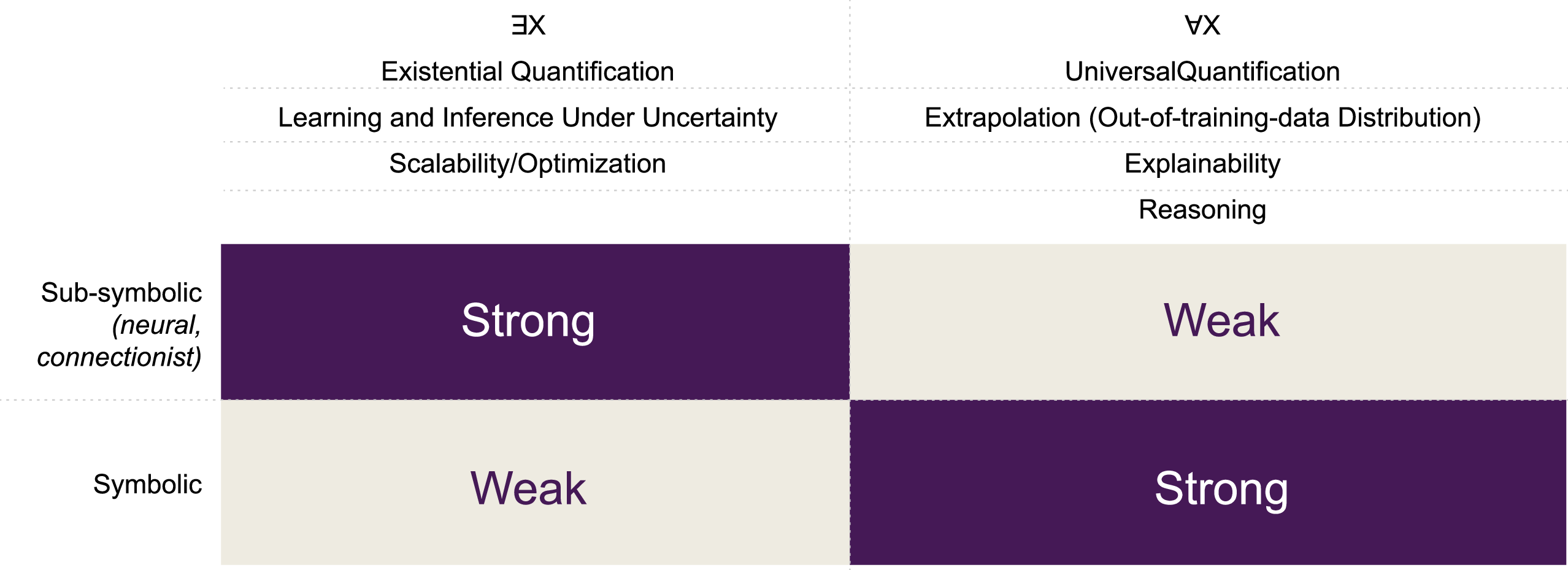}
\caption{Symbolic vs Sub-Symbolic strengths and weaknesses. Based on the work of \cite{Garcez_Gori_Lamb_Serafini_Spranger_Tran_2019} }
\label{fig:capabilities}
\end{figure}

At its core, Neuro-Symbolic AI (NeSy) is \say{the combination of deep learning and symbolic reasoning} \cite{Garcez_Lamb_2020}. The goal of NeSy is to address the weaknesses of each of symbolic and sub-symbolic (neural, connectionist)  approaches while preserving their strengths (see figure \ref{fig:capabilities}). Thus NeSy promises to deliver a best-of-both-worlds approach which embodies the \say{two most fundamental aspects of intelligent cognitive behavior: the ability to learn from experience, and the ability to reason from what has been learned} \cite{Garcez_Lamb_2020, Valiant_2003}.

Remarkable progress has been made on the learning side, especially in the area of Natural Language Processing (NLP) and in particular with deep learning architectures such as the Transformer \cite{vaswani2017attention,devlin-etal-2019-bert}. However, these systems display certain intrinsic weaknesses which some researchers \cite{pearl2018theoretical,marcus2018deep} argue cannot be addressed by deep learning alone and that in order to do even the most basic reasoning, we need rich representations which enable precise, human interpretable inference via mathematical logic.\footnote{See also Besold et al. \cite{besold2017neural}, p.17-18 for additional context.}

\begin{figure}[htbp]
    \centering
    \includegraphics[scale=0.55]{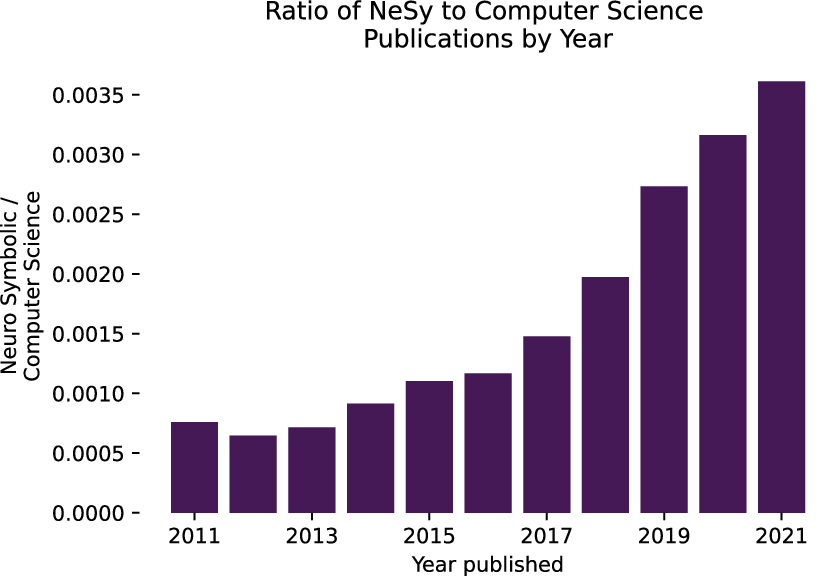}
    \caption{Number of Neuro Symbolic articles published since 2010, normalized by the total number of all Computer Science articles published each year. The figure represents the unfiltered results from Scopus given the search keywords described in section \ref{search_process}.}
    \label{fig:allNeSy}
\end{figure}

Recently, a discussion between Gary Marcus and Yoshua Bengio at the 2019 Montreal AI Debate prompted some passionate exchanges in AI circles, with Marcus arguing that \say{expecting a monolithic architecture to handle abstraction and reasoning is unrealistic}, while Bengio defended the stance that \say{sequential reasoning can be performed while staying in a deep learning framework} \cite{AIDebate2019}.
Spurred by this discussion, and almost ironically, by the success of deep learning (and ergo, the clarity into its limitations), research into hybrid solutions has seen a dramatic increase (see figure \ref{fig:allNeSy}). At the same time, discussion in the AI community has culminated in \say{violent agreement} \cite{Kautz} that the next phase of AI research will be about \say{combining neural and symbolic approaches in the sense of NeSy AI [which] is at least a path forward to much stronger AI systems} \cite{Sarker_Zhou_Eberhart_Hitzler_2021}. Much of this discussion centers around the ability (or inability) of deep learning to reason, and in particular, to reason outside of the training distribution. Indeed, at IJCAI 2021, Yoshua Bengio affirms that \say{we need a new learning theory to deal with Out-of-Distribution generalization} \cite{IJCAI2021}. Bengio's talk is titled \say{System 2 Deep Learning: Higher-Level Cognition, Agency, Out-of-Distribution Generalization and Causality.} Here, System 2 refers to the System 1/System 2 dual process theory of human reasoning explicated by psychologist and Nobel laureate Daniel Kahneman in his 2011 book \say{Thinking, Fast and Slow} \cite{kahneman2011thinking}. AI researchers \cite{Garcez_Lamb_2020,Liu_Wang_Lin_Li_2022,Zhang_Chen_Zhang_Ke_Ding_2021,marcus2018deep,Lamb_Garcez_Gori_Prates_Avelar_Vardi_2020,von_Rueden_Mayer_Beckh_Georgiev_Giesselbach_Heese_Kirsch_Walczak_Pfrommer_Pick_etal_2021,Belle_2020} have drawn many parallels between the characteristics of sub-symbolic and symbolic AI systems and human reasoning with System 1/System 2. Broadly speaking, sub-symbolic (neural, deep-learning) architectures are said to be akin to the fast, intuitive, often biased and/or logically flawed System 1. And the more deliberative, slow, sequential System 2 can be thought of as symbolic or logical. But this is not the only theory of human reasoning as we will discuss later in this paper. It should also be noted that Kahneman himself has cautioned against the over reliance on the System 1/System 2 analogy in a followup discussion at the Montreal AI Debate 2 the following year, stating, \say{I think that this idea of two systems may have been adopted more than it should have been.}\footnote{\url{https://youtu.be/2zNd69ZGZ8o?t=161}}

\subsection{Reasoning \& Language}

\say{Language understanding in the broadest sense of the term, including question answering that requires commonsense reasoning, offers probably the most complete application area of neurosymbolic AI} \cite{Garcez_Lamb_2020}. This makes a lot of intuitive sense from a linguistic perspective. If we accept that language is compositional, with rules and structure, then it should be possible to obtain its meaning via logical reasoning. Compositionality in language was formalized by Richard Montague in the 1970s, in what is now referred to as \textit{Montague grammar}: \say{The key idea is that compositionality requires the existence of a homomorphism between the expressions of a language and the meanings of those expressions.}\footnote{\url{https://plato.stanford.edu/entries/compositionality/\#FormStat}} In other words, there is a direct relationship between syntax and semantics (meaning). This is in line with Noam Chomsky's \textit{Universal grammar}\footnote{\url{https://www.britannica.com/topic/universal-grammar}} which states that there is a structure to natural language which is innate and universal to all humans, and is governed by precise mathematical rules. While an analysis of the study of linguistics is beyond the scope of this paper, the key takeaway is this: what makes such theories so attractive to computational linguists is that meaning can be derived from syntactic structures which can be translated into computer programs. Today, industrial strength tools for extracting these structures (i.e., part-of-speech tagging, constituency parsing, dependency parsing) are readily available, such as for example  NLTK\footnote{\url{https://www.nltk.org/}} or SpaCy\footnote{\url{https://spacy.io/}}. The challenge lies in representing and utilizing these structures in a way that both captures the semantics and is computationally efficient. 

On the one hand, distributed representations are desirable because they can be efficiently processed by gradient descent (the backbone of deep learning). The downside is that the meaning embedded in a distributed representation is difficult if not impossible to decompose. So while a Large Language Model (LLM), a deep learning language model based on the principle of distributional semantics, may be very good at making certain types of predictions, it cannot be queried for answers not present in the training data by way of analogy or logic. We have also seen that even as these models get infeasibly large - the larger the model, the better the predictions \cite{Sharir_Peleg_Shoham_2020} - they still fail on tasks requiring basic commonsense. The example in Figure \ref{fig:gpt3fumble}, given by Marcus and Davis in \cite{Marcus_Davis} is a case in point.  

\begin{figure}[htbp]
    \footnotesize
    \begin{zitat}{}
    You are having a small dinner party. You want to serve dinner in the living room. The dining room table is wider than the doorway, so to get it into the living room, you will have to \textbf{remove the door. You have a table saw, so you cut the door in half and remove the top half.}
    \end{zitat}
    \caption{Third Generation Generative Pre-trained Transformer (GPT3) \cite{gpt3} text completion example. The prompt is rendered in regular font, while the GPT3 response is shown in bold. It is clear that GPT3 is incapable of commonsense.}
    \label{fig:gpt3fumble}
\end{figure}

On the other hand, traditional symbolic approaches have also failed to capture the essence of human reasoning. While we may not yet understand exactly how people reason, it is generally accepted that human reasoning is nothing like the rigorous mathematical logic where the goal is validity. Though not for lack of ambition - Socrates got himself killed trying to get people to reason with logic \cite{farnsworth2021socratic}. In the \textit{Dictionary of Cognitive Science} \cite{engel_2003}, Pascal Engel describes reasoning in a natural setting as \say{ridden with errors and paralogisms.} Engel refers to Daniel Kahneman, Amos Tversky, Philip Wason, among others, who have conducted numerous experiments and written extensively showing how logical fallacies and \say{noise} can lead to those errors \cite{kahneman2011thinking, Kahneman_Sibony_Sunstein_2021}. But even when the objective is not to emulate human thinking, but rather the execution of tasks which require precise, deterministic answers such as expert reasoning or planning, traditional symbolic reasoners are slow, cumbersome, and computationally intractable at scale, \say{typically subject to combinatorial explosions that limit both the number of axioms, the number of individuals and relations described by these axioms, and the depth of reasoning that is possible} \cite{Belle_2020}. For example, Description Logics (DLs) such as OWL\footnote{\url{https://www.w3.org/2007/OWL/wiki/Direct_Semantics}} are used to reason over ontologies and knowledge graphs (KGs). However, one must accept a harsh trade-off between expressivity and complexity when choosing a DL flavor. Improving the performance of reasoning over ontologies and knowledge graphs that power search and information retrieval across the web is particularly relevant to the Semantic Web community. Hitzler et al. \cite{Hitzler_Bianchi_Ebrahimi_Sarker_2020} report on recent research on neuro-symbolic integration in relation to the Semantic Web field, with a focus on the promises and possible benefits for both.

The remainder of this manuscript is structured as follows. Section \ref{sec:preliminaries} offers a brief history of NLP in the context of reasoning. Several recent surveys and their contributions to NeSy are discussed in section \ref{sec:related_work}, and are intended as an introduction to the field. Our contribution is given in section \ref{sec:contributions}, which also details the goals of NeSy selected for this survey. Section \ref{methods} describes the research methods employed for searching and analysing relevant studies. In Section \ref{results} we analyze the results of the data extraction, how the studies reviewed fit into Henry Kautz's NeSy taxonomy \cite{Kautz}, and we propose a simplified nomenclature for describing Kautz's NeSy categories. Section \ref{discussion} discusses the limitations and challenges of the reviewed implementations. Section \ref{future} presents limitations of this work and future directions for NeSy in NLP, followed by the conclusion in Section \ref{conclusion}.

\section{A Brief History of NLP}\label{sec:preliminaries}

\begin{figure}[htbp]
    \centering
    \includegraphics[scale=0.4]{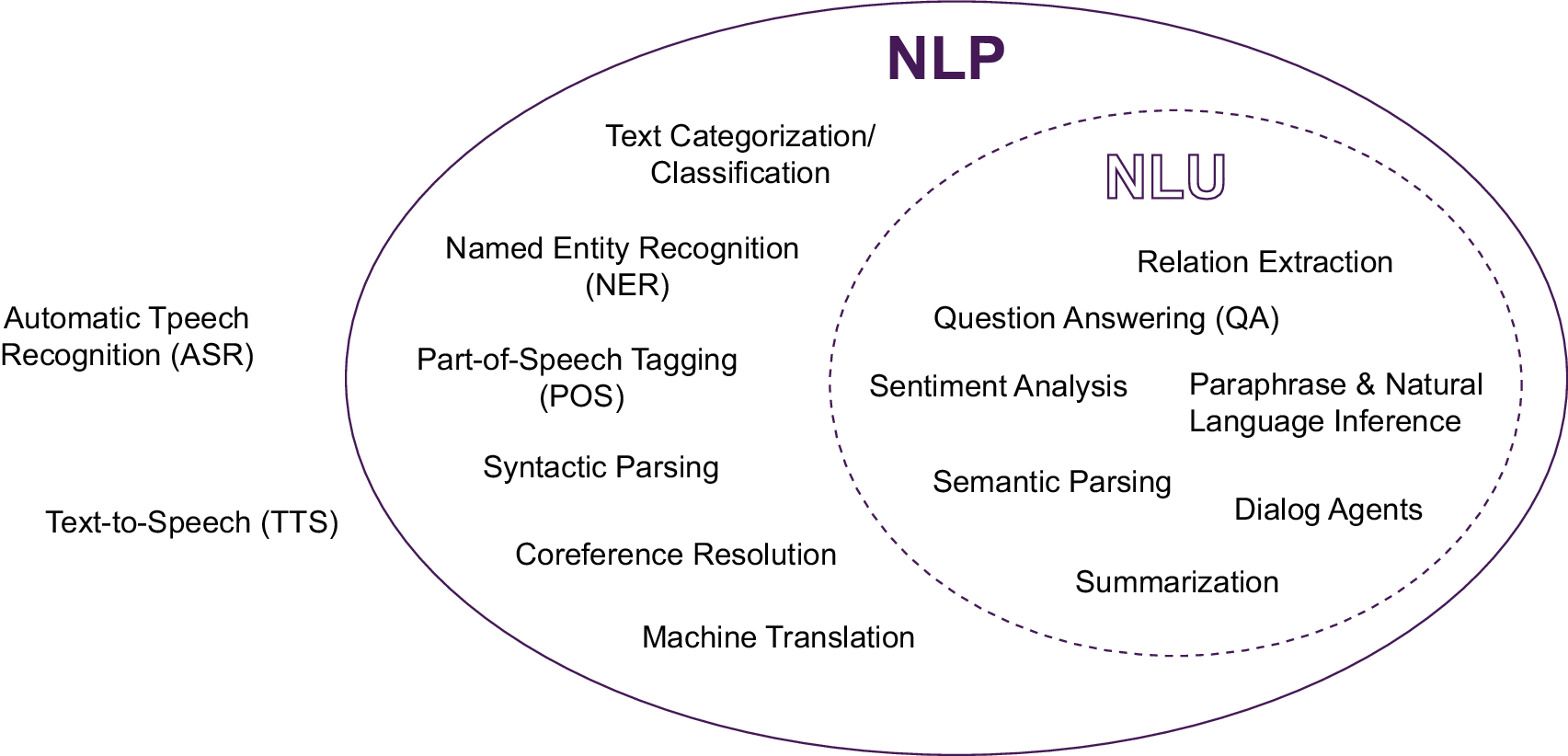}
    \caption{Common Natural Language Processing tasks \cite{Bill_MacCartney_2014}.}
    \label{fig:nlp_tasks}
\end{figure}

The study of language and reasoning goes back thousands of years, but it was not until the 1960's that the first computational models were realized. The Association for Computational Linguistics (ACL)\footnote{\url{https://www.aclweb.org/portal/}} was founded in 1962 for people working on computational problems involving human language, a field often referred to as either computational linguistics or \textbf{Natural Language Processing (NLP)}. Common NLP tasks are illustrated in Figure \ref{fig:nlp_tasks}.

One of the first NLP projects was a chat-bot named ELIZA \cite{weizenbaum1966eliza}, written by Joseph Weizenbaum around 1965. Given a small hand crafted set of rules, ELIZA was able to hold an, albeit superficial, conversation, gaining tremendous popularity. Curiously, despite the program's simplicity those who interacted with it, attributed to it human-like emotions. These early systems were based on pattern matching and small rule-sets, and were very limited for obvious reasons. In the 1970s and 80s linguistically rich, logic-driven, grounded systems, largely influenced by Noam Chomsky's \textit{Universal Grammar}\footnote{\url{https://www.cs.bham.ac.uk/~pjh/sem1a5/pt1/pt1_history.html}} were developed. The 1990s and early 2000s saw the `statistical revolution' and the rise of machine learning, and work on NLP tasks focused on semantics, such as Natural Language Understanding (NLU), diminished for the next decade or so\footnote{\url{https://nlp.stanford.edu/~wcmac/papers/20140716-UNLU.pdf}}. NLU returns to center stage, mixing techniques from previous years sometime around 2010. As a case in point, in 2011 IBM's Watson DeepQA computer system won first place on Jeopardy! for a prize of \$1 million, competing  against champions Brad Rutter and Ken Jennings.\footnote{\url{https://www.youtube.com/watch?v=lI-M7O\_bRNg}} DeepQA is a large ensemble of techniques and models, the vast majority of which was focused on general Information Retrieval (IR), NLP/NLU, Knowledge Representation \& Reasoning (KRR), and Machine Learning (ML) \cite{watson}. Broadly speaking, DeepQA is a large neuro-symbolic question answering software pipeline. In the last decade, and especially in the last few years, the emphasis on deep learning has somewhat overshadowed traditional NLP approaches. The Long Short Term Memory (LSTM) \cite{Hochreiter_1997_LSTM} architecture paved the way for the Transformer, which has generated a huge amount of optimism leading some people to believe that \say{deep learning is going to be able to do everything.}\footnote{\url{https://www.technologyreview.com/2020/11/03/1011616/ai-godfather-geoffrey-hinton-deep-learning-will-do-everything/}} However, as already mentioned, the success of the Transformer and Large Language Models (LLMs) has also served to highlight their inherent shortcomings. This brings us to the present, or the \say{3rd Wave} \cite{Garcez_Lamb_2020}, which seeks to overcome those shortcomings by combining deep learning with symbolic reasoning and knowledge, and by integrating and expanding on the work of previous decades.

Areas of NLP which are said to benefit from this approach are ones which require some form of reasoning or logic. In particular, Natural Language Understanding (NLU), Natural Language Inference (NLI), and Natural Language Generation (NLG).

\textbf{Natural Language Understanding (NLU)} is a large subset of NLP containing topics particularly focused on semantics and meaning. The boundaries between NLP and NLU are not always clear and open to debate, and even when they are agreed upon, they’re somewhat arbitrary, as it’s a matter of convention and a reflection of history \cite{Bill_MacCartney_2014}.

\textbf{Natural Language Inference (NLI)} enables tasks like semantic search, information retrieval, information extraction, machine translation, paraphrase acquisition, reading comprehension, and question answering. It is the problem of determining whether a natural language hypothesis $h$ can reasonably be inferred from a given premise $p$ \cite{maccartney_manning_2009_extended}. For example, the premise \say{Hazel is an Australian Cattle Dog}, entails the hypothesis \say{Hazel is a dog}, and can be expressed in First Order Logic (FOL) by: $p \models h$.

\begin{figure}[htbp]%
    \centering
    \subfloat[\centering Symbolic view - reasoning is performed explicitly via rules and logic]{{\includegraphics[scale=1]{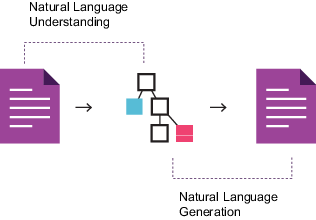} }}%
    \qquad
    \subfloat[\centering Connectionist view - reasoning is performed implicitly inside the neural network]{{\includegraphics[scale=1]{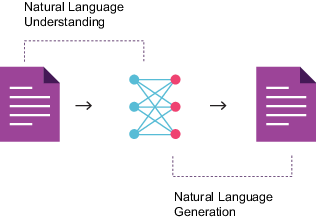} }}%
    \caption{NLU takes as input unstructured text and produces output which can be reasoned over. NLG takes as input structured data and outputs a response in natural language.}%
    \label{fig:nlu-versus-nlg}%
\end{figure}

\textbf{Natural Language Generation (NLG)} is the task of generating text or speech from non-linguistic (structured) input \cite{Gatt_Krahmer_2018}. It can be seen as orthogonal to NLU, where the input is natural language. An end-to-end system can be made up of both NLU and NLG components. When that is the case, what happens in the middle is not always that clear-cut. A neural language model such as GPT3 \cite{gpt3} has no structured component, however, whether it performs \say{understanding} is subject to debate - Figure \ref{fig:nlu-versus-nlg}.

\section{Related Work}\label{sec:related_work}
Several recent surveys \cite{Garcez_Lamb_2020,Sarker_Zhou_Eberhart_Hitzler_2021,Garcez_Gori_Lamb_Serafini_Spranger_Tran_2019,Yu_Yang_Liu_Wang_2021,besold2017neural,von_Rueden_Mayer_Beckh_Georgiev_Giesselbach_Heese_Kirsch_Walczak_Pfrommer_Pick_etal_2021,Zhang_Chen_Zhang_Ke_Ding_2021,Lamb_Garcez_Gori_Prates_Avelar_Vardi_2020,Belle_2020} cover neuro-symbolic architectures in detail. Our aim is not to produce another NeSy survey, but rather to examine whether the promises of NeSy in NLP are materializing. However, for completeness, and by way of introduction to the subject, we briefly summarize each of these surveys and provide references for the architectures under review.

\begin{figure}[htbp]%
    \centering
    \includegraphics[scale=0.75]{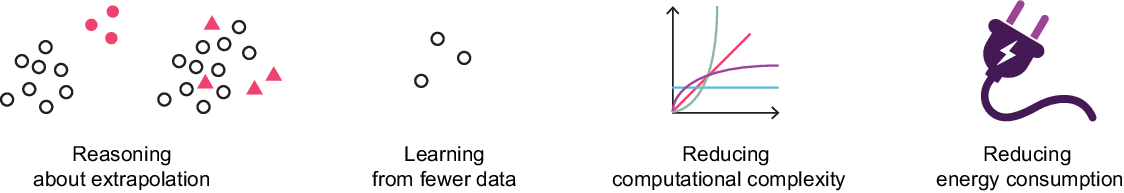} 
    \caption{Neuro-Symbolic Artificial Intelligence promise areas \cite{Garcez_Lamb_2020}}
    \label{fig:promises-garcez}%
\end{figure}

In response to recent discussions in the AI community and the resurgence of interest in NeSy AI, Garcez et al. \cite{Garcez_Lamb_2020} synthesize the last 20 years of research in the field in the context of the aforementioned debate. The authors highlight the need for trustworthiness, interpretability, and accountability in AI systems, which ostensibly, NeSy is most suited to, in particular when it comes to natural language understanding. The authors also emphasize the distinction between commonsense knowledge and expert knowledge, and suggest that these two goals may ultimately lead to two distinct research directions: \say{those who seek to understand and model the brain, and those who seek to achieve or improve AI.} Garcez at al. conclude that \say{Neurosymbolic AI is in need of standard benchmarks and associated comprehensibility tests which could in a principled way offer a fair comparative evaluation with other approaches} with a focus on the following goals: learning from fewer data, reasoning about extrapolation, reducing computational complexity, and reducing energy consumption\footnote{Energy consumption is particularly significant when training Large Language Models which can cost in the thousands if not millions of dollars in electricity \cite{Sharir_Peleg_Shoham_2020}.} - Figure \ref{fig:promises-garcez}.

\begin{figure}[htbp]%
    \centering
    \includegraphics[scale=0.75]{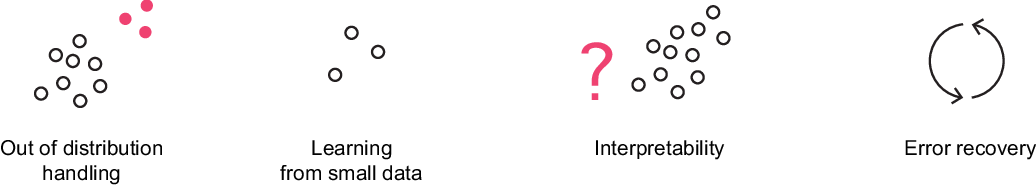} 
    \caption{Neuro-Symbolic Artificial Intelligence promise areas \cite{Sarker_Zhou_Eberhart_Hitzler_2021}}
    \label{fig:promises-sarker}%
\end{figure}

Sarker et al. \cite{Sarker_Zhou_Eberhart_Hitzler_2021} survey recent work in the proceedings of leading AI conferences. The authors review a total of 43 papers and classify them according to Henry Kautz's categories\footnote{Henry Kautz introduced a taxonomy of NeSy types at the Third AAAI Conference on AI \cite{Kautz}. We rely on this taxonomy to classify the studies under review, and  discuss each type in detail in section \ref{sec:nesycategories}}, as well as an earlier categorisation scheme from 2005  \cite{Bader_Hitzler_2005}. Comparing the earlier research to the current trends, the authors confirm advancements on both the neural side, as well as the logic side, with a tendency towards more expressive logics being explored today than was thought tractable in the past, and the influence of the success of neural networks on the rise in interest in NeSy in general. Sarker et al. identify four areas of AI that can benefit from NeSy approaches: Learning from small data, Out of distribution handling, Intepretability, and Error recovery - Figure \ref{fig:promises-sarker}.

The authors conclude that \say{more emphasis is needed, in the immediate future, on deepening the logical aspects in NeSy research even further, and to work towards a systematic understanding and toolbox for utilizing complex logics in this context.} Based on the studies in our review, we come to a similar conclusion.

Garcez et al. \cite{Garcez_Gori_Lamb_Serafini_Spranger_Tran_2019} survey recent accomplishments for integrated machine learning and reasoning motivated by the need for interpretability and accountability in AI systems. According to  \cite{Garcez_Gori_Lamb_Serafini_Spranger_Tran_2019}, there are three main important features of a NeSy system: Representation, Extraction, and Reasoning \& Learning. Symbolic knowledge can also be categorized into three groups: rule-based, formula-based, and embedding-based. The authors categorize and describe the following neuro-symbolic architectures.

\begin{figure}[htbp]%
    \centering
    \includegraphics[scale=0.3]{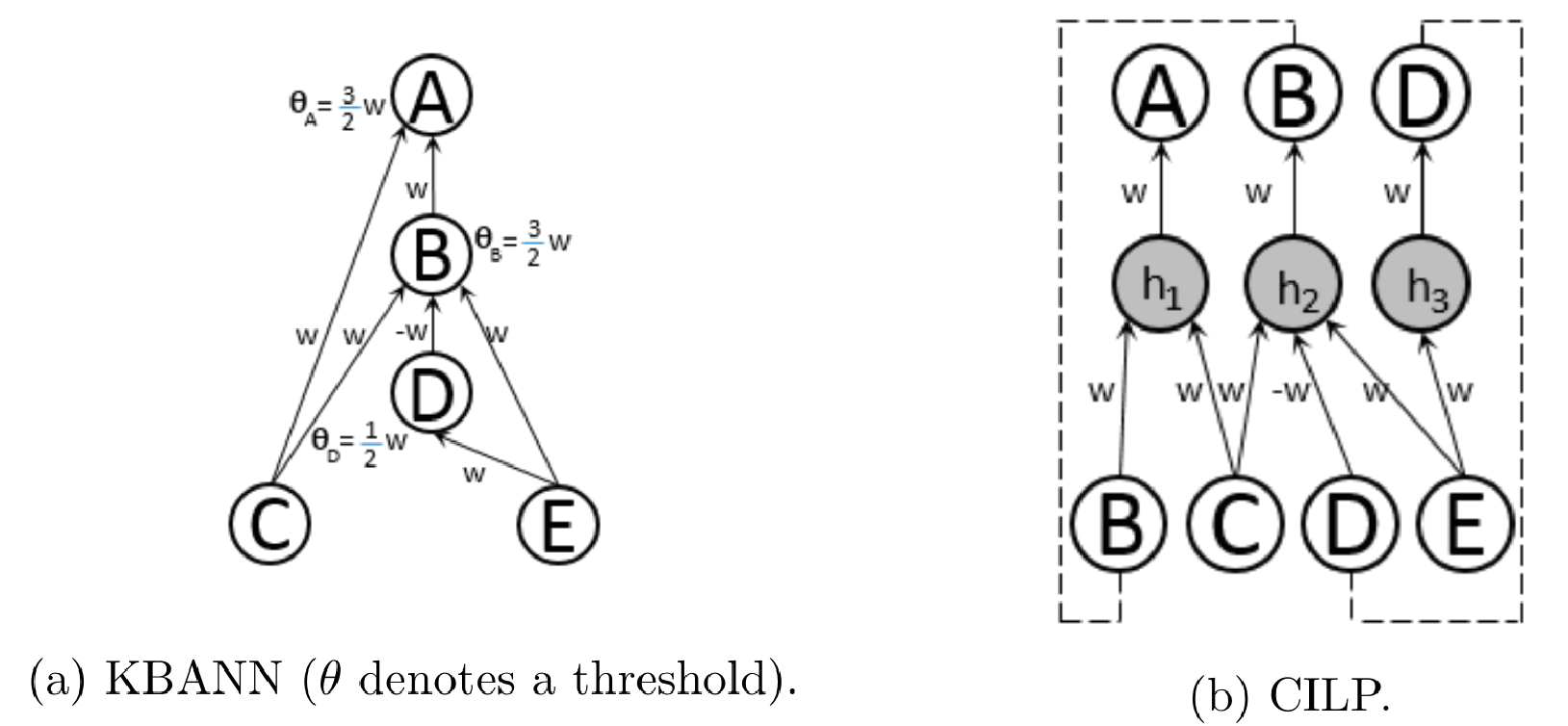} 
    \caption{Knowledge representation of $\phi = \{A \leftarrow B \land C, B \leftarrow C \land 	\neg D \land E, D \leftarrow E\}$ using KBANN and CILP. \cite{Garcez_Gori_Lamb_Serafini_Spranger_Tran_2019}}%
    \label{fig:kbann-cilp}%
\end{figure}

\textbf{Early systems} such as KBANN \cite{towell1994knowledge} and CILP  \cite{garcez2002neural} embed propositional logic in a neural network by constraining the model parameters - Figure \ref{fig:kbann-cilp}.

\begin{figure}[htbp]%
    \centering
    \includegraphics[scale=0.3]{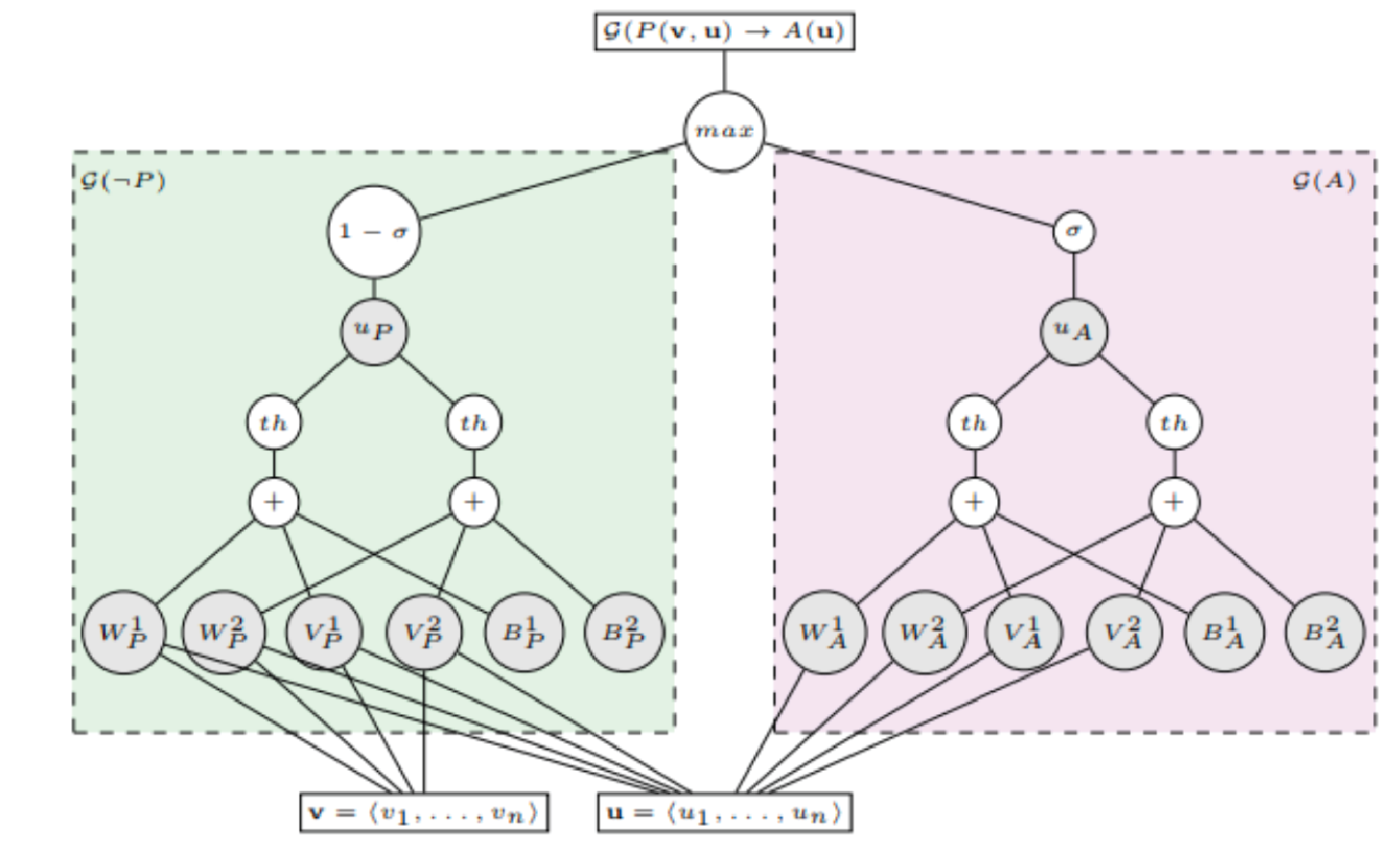} 
    \caption{Logic Tensor Network (LTN) for $P ( x, y ) \rightarrow A ( y )$ with $G ( x ) = v$ and $G ( y ) = u$; G are grounding (vector representation) for symbols in first-order language. \cite{Garcez_Gori_Lamb_Serafini_Spranger_Tran_2019}}%
    \label{fig:ltn}%
\end{figure}

\textbf{Tensorization} is a process that embeds first order logic (FOL) symbols into real-valued tensors. Reasoning is performed through matrix computation. Examples include Logic Tensor Networks (LTNs) \cite{Serafini_dAvila_Garcez_2016} and Neural Tensor Networks (NTLs) \cite{Socher_Chen_Manning_Ng_2013} -  Figure \ref{fig:ltn}.

In \textbf{Neural-Symbolic Learning} the primary goal is learning, with the assistance of rules and logic. Different architectures are characterized by how the logic is incorporated into the network, and how it is translated into differentiable form. 
  \begin{itemize}
    \item \textit{Inductive Logic Programming} (ILP) \cite{Muggleton_1991} is a set of techniques for learning logic programs from examples:
    \begin{itemize}
        \item Neural Logic Programming (NLP) \cite{yang2017differentiable}
        \item Differentiable Inductive Logic Programming (\(\partial\)ILP)  \cite{evans2018learning}
        \item Neural Theorem Prover (NTP) \cite{rocktaschel2016learning}
        \item Neural Logic Machines (NLMs) \cite{dong2019neural}
    \end{itemize}
    \item \textit{Horizontal Hybrid Learning } combines expert knowledge in the form of rules/logic with data, thus are suitable to knowledge transfer learning (horizontally across domains).  
    \item \textit{Vertical Hybrid Learning} combines symbolic and sub-symbolic modules which take inspiration from neuroscience in that certain areas of the brain are responsible for processing input signals, while other areas perform logical thinking and reasoning (vertically for a single domain). 
  \end{itemize}  
  
\textbf{Neural-Symbolic Reasoning} concerns itself with logical reasoning, as the name suggests, powered by neural computation. These consist of model-based, and theorem proving approaches. In early theorem proving systems such as SHRUTI \cite{wendelken2004multiple} learning capability was limited. On the other hand, model-based approaches inside neural networks have been shown to demonstrate nonmonotonic, intuitionistic, abductive, and other forms of human reasoning capability. Hence, rather than attempting to perform both learning and reasoning in a single architecture, more recent designs tend to contain separate learning and reasoning modules which communicate with each other. The authors conclude that combining symbolic and sub-symbolic modules, in other words, the compositionality of neuro-symbolic systems, contributes to the development of explainable and accountable AI \cite{vilone2021notions}.

\begin{figure}[htbp]%
    \centering
    \includegraphics[scale=0.4]{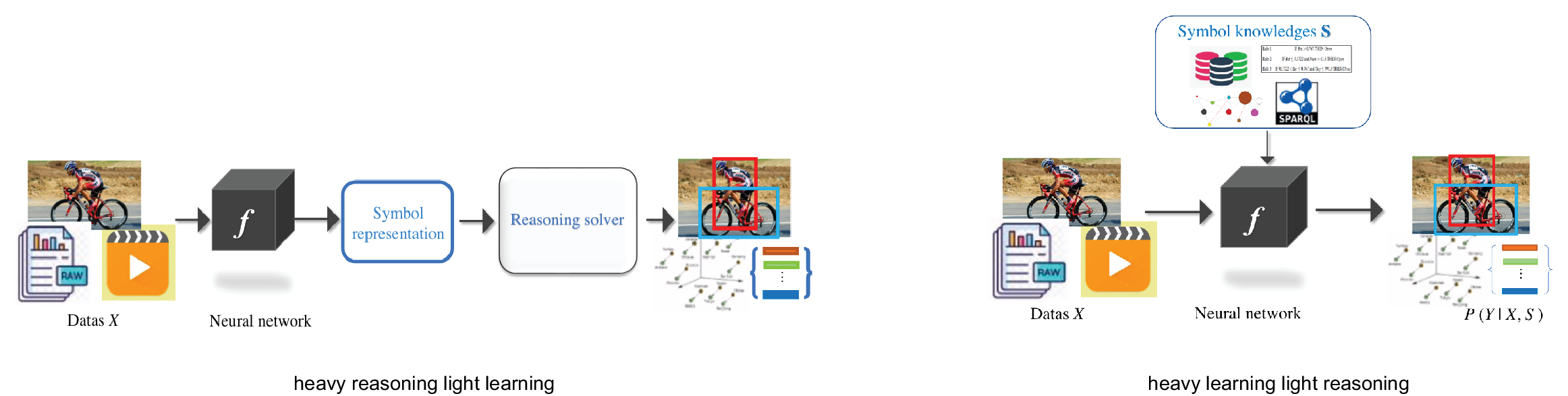} 
    \caption{Two types of neuro-symbolic systems: heavy reasoning light learning, and heavy learning light reasoning \cite{Yu_Yang_Liu_Wang_2021}}%
    \label{fig:yuatal}%
\end{figure}

Yu et al. \cite{Yu_Yang_Liu_Wang_2021} divide neuro-symbolic systems into two types: heavy-reasoning light-learning and heavy-learning light-reasoning (Figure \ref{fig:yuatal}). These are similar to the Neural-Symbolic  Reasoning  and Neural-Symbolic Learning categorization in \cite{Garcez_Gori_Lamb_Serafini_Spranger_Tran_2019} above.
Heavy-reasoning light-learning mainly adopts the methods of the symbolic system to solve the problem in machine reasoning, and introduces neural networks to assist in solving those problems, while heavy-learning light-reasoning mainly applies methods of the neural system to solve the problem in machine learning, and introduces symbolic knowledge in the training process.

\begin{itemize}
    \item Heavy reasoning light learning (based on Statistical Relational Learning (SRL) \cite{koller2007introduction})
    \begin{itemize}
        \item Probabilistic Logic Programming (ProbLog) \cite{de2007problog}
        \item Markov Logic Network (MLN) \cite{Richardson_Domingos_2006}
        \item Inductive Logic Programming (ILP) \cite{Muggleton_1991}
    \end{itemize}
    \item Heavy learning light reasoning
    \begin{itemize}
        \item \textit{Regularization models} add symbols in the form of regular terms to the objective function as a kind of prior knowledge to guide training. 
        \item \textit{Knowledge transfer models} integrate the knowledge graph that represents semantic information into the neural network model, making up for the lack of data by transferring semantic knowledge. Knowledge transfer models are mainly used to solve zero-shot learning and few-shot learning \cite{wang2020generalizing} tasks.
    \end{itemize}
\end{itemize}

Besold et al. \cite{besold2017neural} examine neuro-symbolic learning and reasoning through the lens of cognitive science, cognitive neuro-science, and human-level artificial intelligence. This is a much more theoretical approach. The authors first describe some early systems such as CILP \cite{garcez2002neural} and fibring, introduced by Garcez \& Gabby \cite{garcez2004fibring}. Fibred networks work on the principle of recursion, where multiple neural networks are connected together, such that a fibring function in a network A, determines which neurons should be activated in a network B. A key characteristic of neuro-symbolic systems is modularity, where each network in the ensemble is responsible for a specific logic or task, increasing expressivity and allowing for non-classical logics to be represented such as connectionist modal, intuitionistic, temporal, nonmonotonic, epistemic and relational logic. Neuro-symbolic computation encompasses the integration of cognitive abilities - induction, deduction, abduction - and the study of mental models. The study of mental models has a long history, and the authors reference research from the field of neuro science and cognitive science, including the \say{binding} problem, dual process theory (System 1/System 2), and theories of affect; with the goal of formulating these in a neuro-symbolic system. Of particular interest to our work are the two sections on Syntactic Structures, and Compositionality, as they both deal with modeling language. Psycho-linguists have different theories of language morphology (study of the internal construction of words\footnote{\url{https://www.britannica.com/topic/morphology-linguistics}}), with some arguing for association based explanations (McClelland \cite{joanisse2015connectionist}), while others argue for a rule-based one (Pinker \cite{Pinker_1999}) - the question being whether it is better to model language through a connectionist approach, per McClelland, or a symbolic one, as per Pinker. Whether to model language in a connectionist or symbolic manner hinges also on its inherent compositionality\footnote{According to Noam Chomsky theory of language, language is compositional, in the sense that a sentence is composed of phrases, which are in turn composed of sub-phrases, and so on, in a recursive manner. This idea enables the construction of infinite possibilities from finite means. This seems particularly well suited to a symbolic system which, given a finite set of rules should be capable of constructing/deconstructing (reasoning over) all possibilities. In contrast, a sub-symbolic, or distributional, system can never see the infinite amount of the data in the universe to learn from. For learning in infinite domains, see also \cite{Belle_2020}. \url{https://www.britannica.com/biography/Noam-Chomsky/Rule-systems-in-Chomskyan-theories-of-language}}. 



\begin{figure}[htbp]%
    \centering
    \includegraphics[scale=0.35]{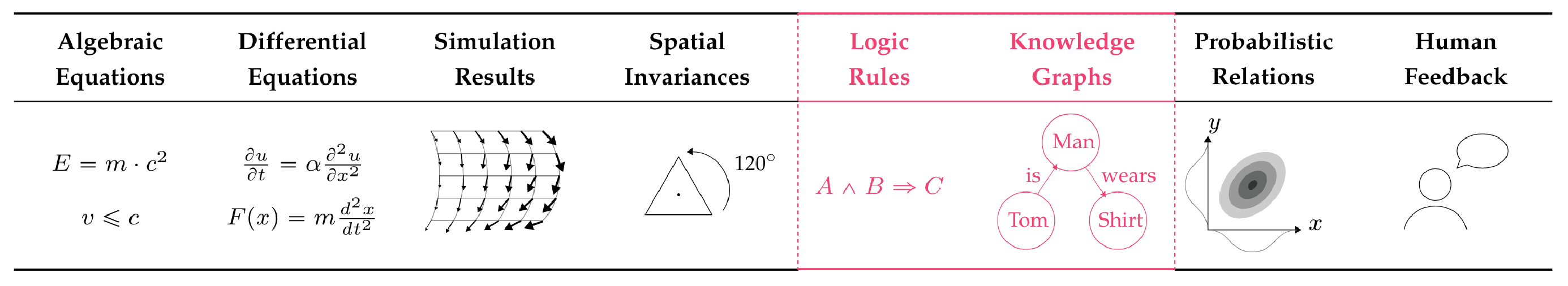} 
    \caption{Types of knowledge representation \cite{von_Rueden_Mayer_Beckh_Georgiev_Giesselbach_Heese_Kirsch_Walczak_Pfrommer_Pick_etal_2021}. Given that our work deals with natural language as input, we are only concerned with Logic Rules (which we subdivide into rules and logic) and Knowledge Graphs (which we subdivide into frames and semantic networks) - see section \ref{sec:kr_taxonomy}}%
    \label{fig:vonrueden}%
\end{figure}

Von Rueden et al. \cite{von_Rueden_Mayer_Beckh_Georgiev_Giesselbach_Heese_Kirsch_Walczak_Pfrommer_Pick_etal_2021} propose a taxonomy for integrating prior knowledge into learning systems. This is an extensive work covering types of knowledge and knowledge representations, neuro-symbolic integration approaches, motivations for each approach, challenges and future directions. The authors categorize knowledge into three types:  scientific knowledge, world knowledge, and expert knowledge. Furthermore, knowledge representations are classified into eight types - Figure \ref{fig:vonrueden}.

Zhang et al. \cite{Zhang_Chen_Zhang_Ke_Ding_2021} survey the area of neuro-symbolic reasoning on Knowledge Graphs (KGs). The authors contribute a unified reasoning framework for Knowledge Graph Completion (KGC) and Knowledge Graph Question Answering (KGQA). Among future directions, the authors advocate for taking inspiration from human cognition for neural-symbolic reasoning in KGs, alluding to the dual model of human reasoning (System 1/System 2). Additional future directions include:
\begin{itemize}
    \item \textit{Few-shot Reasoning} which addresses the issue of few labeled examples.
    \item \textit{Reasoning upon Multi-sources} which incorporates additional information from unstructured text.
    \item \textit{Dynamic Reasoning} which deals with inferring new facts evolving over time.
    \item \textit{Analogical Reasoning (AR)} which involves the use of past experiences to solve problems that are similar to problems solved before. Case Based Reasoning (CBR) is an example of AR \cite{Sriram_1997}.
    \item \textit{Knowledge Graph Pre-training} which enables transfer learning for domain adaptation.
\end{itemize}

\begin{figure}[htbp]%
    \centering
    \includegraphics[scale=0.65]{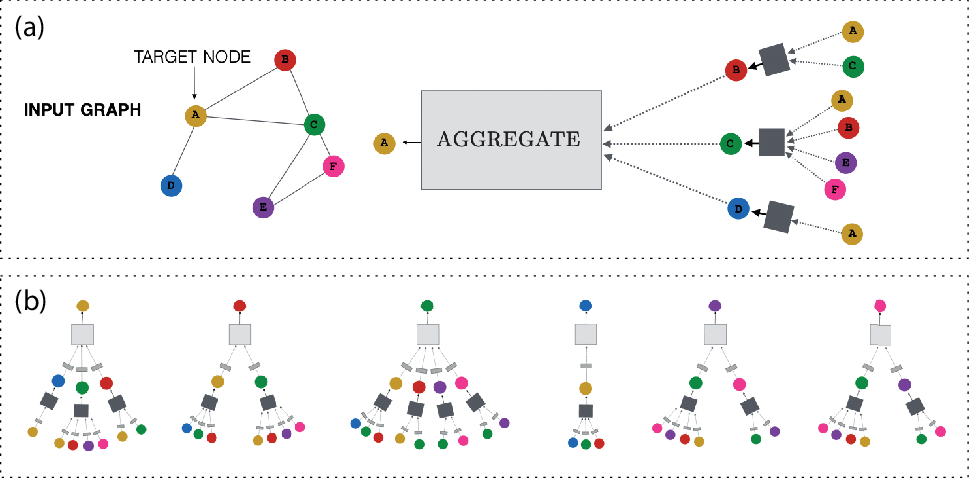} 
    \caption[caption for LOF]{Graph Neural Network (GNN) intuition: generate node embeddings based on local neighborhoods, where nodes aggregate information from their neighbors using neural networks (a). The network neighborhood defines a computation graph such that every node corresponds to a unique computation graph (b). The key distinctions are in how different approaches aggregate information across the layers \cite{Hamilton_Ying_Leskovec_2017}.\protect\footnotemark}
    \label{fig:gnn}%
\end{figure}

Lamb et al. \cite{Lamb_Garcez_Gori_Prates_Avelar_Vardi_2020} review the state of the art on the use of Graph Neural Networks (GNNs) in NeSy (Figure \ref{fig:gnn}). 
\footnotetext{Tutorial slides associated with \cite{Hamilton_Ying_Leskovec_2017}:  \url{http://snap.stanford.edu/proj/embeddings-www/files/nrltutorial-part2-gnns.pdf}}
Similar to \cite{Garcez_Lamb_2020} and our work, this survey is motivated by the AI Debate in Montreal. Henry Kautz's NeSy taxonomy is used as a foundation for describing NeSy systems. A high level overview of state of the art neural architectures (convolutional layers, recurrent layers, and attention) is given, followed by a discussion of each of the following:
\begin{itemize}
    \item Logic Tensor Networks (LTNs) \cite{Serafini_dAvila_Garcez_2016} (Figure \ref{fig:ltn}).
    \item Pointer Networks \cite{vinyals2015pointer}. Pointer networks are based on the encoder/decoder with attention (ie. transformer) architecture, with the modification that the input length can vary. This architecture lends itself to combinatorial optimization problems such as the Traveling Salesperson Problem (TSP).
    \item Graph Convolutional Networks (GCNs) \cite{kipf2017semi} can be thought of as a generalization of Convolutional Neural Networks (CNNs) for non-grid topologies.
    \item Graph Neural Network Model \cite{scarselli2008graph} - early GNN architecture similar to GCN.
    \item Message-passing Neural Networks - similar to GNN with a slightly modified update function \cite{Lamb_Garcez_Gori_Prates_Avelar_Vardi_2020}.
    \item Graph Attention Networks (GATs) \cite{velickovic2017graph}  - implement an attention mechanism enabling vertices to weigh neighbor representations during their aggregation. GATs are known to outperform typical GCN architectures for graph classification tasks.
\end{itemize}
According to the authors, GNNs endowed with attention mechanisms \say{are a promising direction of research towards the provision of rich reasoning and learning in [Kautz's] type 6 neuralsymbolic systems.}  In NLP, GATs have enabled substantial improvements in several tasks through transfer learning over pretrained transformer language models, \footnote{References to relevant works are not provided.} while GCNs have been shown to improve upon the state-of-the-art for seq2seq models \cite{yao2019graph}. GNN models have also been successfully applied to relational tasks over knowledge bases, such as link prediction \cite{schlichtkrull2018modeling}.\footnote{While a detailed review of GNNs in NLP is beyond the scope of this work, we point the interested reader to an online resource dedicated to this topic:  \url{https://github.com/naganandy/graph-based-deep-learning-literature\#computational-linguistics-conferences}.} The authors posit that the application of GNNs in NeSy will bring the following benefits: 
\begin{itemize}
    \item Extrapolation of a learned classification of graphs as Hamiltonian, to graphs of arbitrary size.
    \item Reasoning about a learned graph structure to generalise beyond the distribution of the training data.
    \item Reasoning about the $partOf(X; Y)$ relation (e.g., to make sense of handwritten MNIST digits and non-digits).
    \item Using an adequate self-attention mechanism to make combinatorial reasoning computationally efficient.
\end{itemize}

Belle \cite{Belle_2020} aims to disabuse the reader of the \say{common misconception that logic is for discrete properties, whereas probability theory and machine learning, more generally, is for continuous properties.} The author  advocates for tackling problems that symbolic logic and machine learning might struggle to address individually such as time, space, abstraction, causality, quantified generalizations, relational abstractions, unknown domains, and unforeseen examples.

\begin{figure}[htbp]%
    \centering
    \includegraphics[scale=0.55]{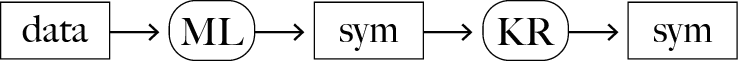} 
    \caption{Schematic diagram using the boxology graphical notation of the AlphaGo system. Ovals denote algorithmic components (i.e. objects that perform some computation), and boxes denote their input and output (i.e. data structures) \cite{van2019boxology}.}%
    \label{fig:boxology}%
\end{figure}

Harmelen \& Teije \cite{van2019boxology} present a conceptual framework to categorize the techniques for combining learning and reasoning via a set of design patterns. \say{Broadly recognized advantages of such design patterns are they distill previous experience in a reusable form for future design activities, they encourage re-use of code, they allow composition of such patterns into more complex systems, and they provide a common language in a community.} A graphical notation is introduced where boxes with labels represent symbolic, and sub-symbolic modules, connected with arrows. Harmelen \& Teije's boxology representation of AlphaGo is given in figure \ref{fig:boxology}.

Earlier surveys \cite{Bader_Hitzler_2005, Hammer_Hitzler_2007, Garcez_Lamb_Gabbay_2009, Gabrilovich_Guha_McCallum_Murphy_2015, Besold_Kuhnberger_2015} tend to focus more on logic and logic programming, and less on learning, which is not surprising given that the ground breaking successes in deep learning are relatively recent. Several themes run through the above listed works, namely, the inherent strengths and weaknesses of symbolic and sub-symbolic techniques when taken in isolation, the types of problems which NeSy promises to solve, and the development of approaches over time. 

Two future directions of particular interest to our work emerge: building systems which take inspiration from human cognition and reasoning, and the integration of unstructured data. To our knowledge there is no survey specifically covering the application of NeSy computing for Natural Language Processing (NLP) where the input data is both unstructured and replete with the ambiguities and inconsistencies of human reasoning. 

\section{Contributions}\label{sec:contributions}

\begin{figure}[htbp]%
    \centering
    \includegraphics[scale=0.75]{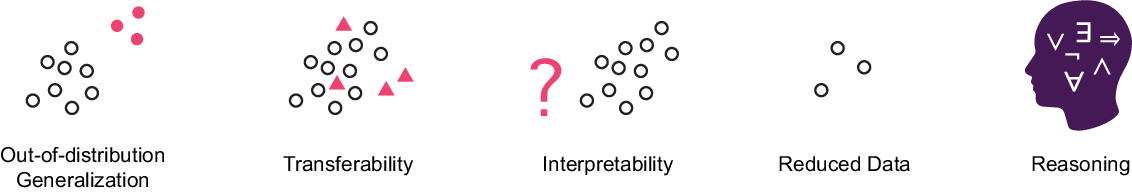} 
    \caption{Neuro-Symbolic Artificial Intelligence Goals}%
    \label{fig:nesygoals}%
\end{figure}

Our aim is to analyze recent work implementing NeSy in the language domain, to verify if the goals of NeSy are being realized, and to identify the challenges and future directions. We briefly describe each of the goals illustrated in figure \ref{fig:nesygoals}, which we have identified based on our synthesis of the related work outlined above.

\subsection{Out-of-distribution (OOD) Generalization}
OOD generalization \cite{Shen_Liu_He_Zhang_Xu_Yu_Cui_2021} refers to the ability of a model to extrapolate to phenomena not previously seen in the training data. The lack of OOD generalization in LLMs is often demonstrated by their inability perform commonsense reasoning, as in the example in Figure \ref{fig:gpt3fumble}.

\subsection{Interpretability}
\label{IandE}
As Machine Learning (ML) and AI become increasingly embedded in daily life, the need to hold ML/AI accountable is also growing. This is particularly true in sensitive domains such as healthcare, legal, and some business applications such as lending, where bias mitigation and fairness are critical. \say{An interpretable model is constrained, following a domain-specific set of constraints that make reasoning processes understandable} \cite{rudin2022interpretable}.

\subsection{Reduced size of training data}
State-of-the-Art (SOTA) language models utilize massive amounts of data for training. This can cost in the thousands or even millions of dollars \cite{Sharir_Peleg_Shoham_2020}, take a very long time, and is neither environmentally friendly nor accessible to most researchers or businesses. The ability to learn from less data brings obvious benefits. But apart from the practical implications, there is something innately disappointing in LLMs' `bigger hammer' approach. Science rewards parsimony and elegance, and NeSy promises to deliver results without the need for such massive scale. While this issue can be partially solved by fine tuning a pre-trained LLM using only a small amount labeled data, these techniques come with their own limitations. For example, Jiang et al. \cite{jiang-etal-2020-smart} discuss issues such as over-fitting the data of downstream tasks and forgetting the knowledge of the pre-trained model.

\subsection{Transferability}
Transferability is the ability of a model which was trained on one domain, to perform similarly well in a different domain. This can be particularly valuable, when the new domain has very few examples available for training. In such cases we might rely on knowledge transfer similar to the way a person might rely on abstract reasoning when faced with an unfamiliar situation \cite{Zhuang_transfer_learning}.

\subsection{Reasoning}\label{sec:intro_reasoning}
According to Encyclopedia Britannica, \say{To reason is to draw inferences appropriate to the situation} \cite{Britanica}. Reasoning is not only a goal in its own right, but also the means by which the other above mentioned goals can be achieved. Not only is it one of the most difficult problems in AI\footnote{As expressed by Luis Lamb at \url{https://video.ibm.com/recorded/131288165}}, it is one of the most contested. Also, a distinction must be made between human-level reasoning, or what is sometimes referred to as commonsense reasoning, and formal reasoning. While human-level reasoning can be ambiguous, error-prone, and difficult to specify, formal reasoning, or logic, follows strict rules and aims to be as precise as possible. The challenge lies in determining when it is appropriate to deploy one or the other or both, and how. In section \ref{discussion:reasoning} we examine the uses of the term reasoning in more depth.

\section{Methods}\label{methods}
Our review methodology is guided by the principles described in \cite{kitchenham2004procedures, Pare_Trudel_Jaana_Kitsiou_2015, Page_McKenzie_Bossuyt_Boutron_Hoffmann_Mulrow_Shamseer_Tetzlaff_Akl_Brennan_et_al._2021}. The data, queries, code, and additional details can be found in our github repository.\footnote{\url{https://github.com/kyleiwaniec/neuro-symbolic-ai-systematic-review}}

\subsection{Research Questions}
\begin{itemize}
\item Is Neuro-symbolic AI meeting its promises in NLP?

    \begin{enumerate}
        \item What are the existing studies on neurosymbolic AI (NeSy) in natural language processing (NLP)?
        \item What are the current applications of NeSy in NLP?
        \item How are symbolic and sub-symbolic techniques integrated and what are the advantages/disadvantages?
    \end{enumerate}

\end{itemize}

\subsection{Search Process}\label{search_process}
We chose Scopus to perform our initial search, as Scopus indexes most of the top journals and conferences we were interested in. In addition to Scopus, we searched the ACL Anthology database and the proceedings from conferences specific to Neuro-symbolic AI. It is possible we missed some relevant studies, but as our aim is to shed light on the field generally, our assumption is that these journals and proceedings are a good representation of the area as a whole. The included sources are listed in Appendix \ref{sec:venues:appendix}. Since we were looking for studies which combine neural and symbolic approaches, our query consists of combinations of neural and symbolic terms as well as variations thereof, listed in table \ref{t1}. The keywords are deliberately broad, as it would be impossible to come up with a complete list of all possible keywords relevant to NeSy in NLP. More importantly, the focus of the work is not on specific subfields, each of which may warrant a review of its own, but rather on the explicit use of neuro-symbolic approaches regardless of subfield. Strictly speaking the only keywords that would cover this would be neuro-symbolic and its syntactic variants, but we relaxed this slightly on the basis that works which explore both symbolic reasoning and deep learning in combination (as per the definition in section \ref{sec:introduction}) may not necessarily have used the term neuro-symbolic.

\begin{table*}[htbp]
\caption{Search Keywords} \label{t1}
\centering
\begin{tabular}{@{}lll}
\toprule
Neural Terms & Symbolic Terms & Neuro-Symbolic Terms\\
\midrule
sub-symbolic & symbolic & neuro-symbolic\\
machine learning & reasoning & neural-symbolic \\
deep learning & logic & neuro symbolic \\
 & & neural symbolic \\
 & & neurosymbolic \\
\bottomrule
\end{tabular}

\end{table*}

The initial query was restricted to peer-reviewed English language journal articles and conference papers from the last 3 years, which produced a total of 21,462 results. 

\subsection{Study selection process}\label{s3}
\begin{figure}[htbp]
\centering
\includegraphics[width=1\textwidth]{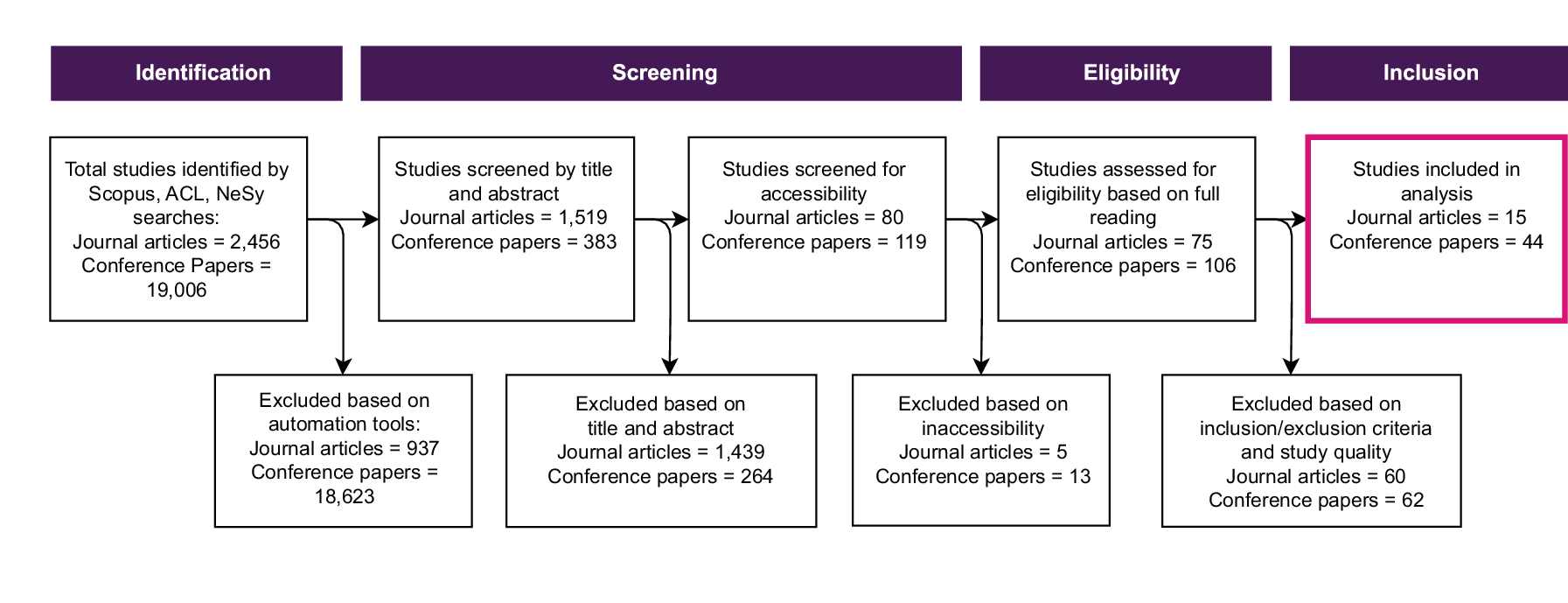}
\caption{Selection Process Diagram}
\label{fig:figure1}
\end{figure}

We further limit the Scopus articles to those published by the top 20 publishers as ranked by Scopus's CiteScore, which is based on number of citations normalized by the document count over a 4 year window\footnote{\url{https://service.elsevier.com/app/answers/detail/a_id/14880/kw/citescore/supporthub/scopus/}}, and SJR (SCImago Journal Rank), a measure of prestige inspired by the PageRank algorithm over the citation network\footnote{\url{https://service.elsevier.com/app/answers/detail/a_id/14883/supporthub/scopus/related/1/}}, the union of which resulted in 29 publishers, and eliminated 19,560 studies, for a total of 1,519 journal articles and 383 conference papers for screening. Two researchers independently screened a sample of each of the 1,902 studies (articles and conference papers), based on the inclusion/exclusion criteria in Table \ref{tbl:inclusion}. The selection process is illustrated in Figure \ref{fig:figure1}.

\begin{table*}[htbp]
\caption{Inclusion/Exclusion Criteria} \label{tbl:inclusion}
\renewcommand{\arraystretch}{1.5}
\begin{tabular}{@{}>{\raggedright\arraybackslash}p{6.5cm}>{\raggedright\arraybackslash}p{6.5cm}}
\toprule
Inclusion & Exclusion \\
\midrule
Input format: unstructured or semi structured text
& Input format: structured query, images, speech, tabular data, categorical data, or any other data type which is not natural language text. \\
Output format: Any
& Application: Theoretical Papers, Position Papers, Surveys, implementations of software pipelines from existing models\\
Application: Implementation of a novel architecture
& The search keywords match, but the actual content does not  \\
Language: English 
& Full text not available (Authors were contacted in these cases) \\
\bottomrule
\end{tabular}
\end{table*}

The inclusion criteria at this stage was intentionally broad, as the process itself was meant to be exploratory, and to inform the researchers of relevant topics within NeSy. As per best practices, this first round is also designed to understand and address inter-annotator disagreement. This unsurprisingly led to some researcher disagreement on inclusion, especially since studies need not have been explicitly labeled as neuro-symbolic to be classified as such. Agreement between researchers can be measured using the Cohen Kappa statistic, with values ranging from [-1,1], where 0 represents the expected kappa score had the labels been assigned randomly, -1 indicates complete disagreement, and 1 indicates perfect agreement. Our score at this stage came to a modest 0.33. We observed that it was not always clear from the abstract alone whether the sub-symbolic and symbolic methods were integrated in a way that meets the inclusion criteria.

To attain inter-annotator agreement and facilitate the next round of review, we kept a shared glossary of symbolic and sub-symbolic concepts as they presented themselves in the literature. We each reviewed all of the 1,902 studies, this time by way of a shallow reading of the full text of each study. Any disagreement at this stage was discussed in person with respect to the shared glossary. This process led to 75 journal articles and 106 conference papers marked for the final round of inclusion/exclusion.

\subsection{Quality Assessment}\label{quality}

During the final round of inclusion/exclusion, the quality of each study was determined through the use of a nine-item questionnaire. Each of the following questions was answered with a binary value, and the study's quality was determined by calculating the ratio of positive answers. Less than a handful of studies were excluded due to a quality score of less than 50\%. 

\begin{enumerate} [Q1.]
    \item Is there a clear and measurable research question?
    \item Is the study put into context of other studies and research, and design decisions justified accordingly (number of references in the literature review/ introduction)? 
    \item Is it clearly stated in the study which other algorithms the study’s algorithm(s) have been compared with?
    \item Are the performance metrics used in the study explained and justified?
    \item Is the analysis of the results relevant to the research question?
    \item Does the test evidence support the findings presented?
    \item Is the study algorithm sufficiently documented to be reproducible (independent researchers arriving at the same results using their own data and methods)?
    \item  Is code provided?
    \item Are performance metrics provided (hardware, training time, inference time)?
\end{enumerate}
\begin{figure}[htbp]
    \centering
    \includegraphics[scale=0.45]{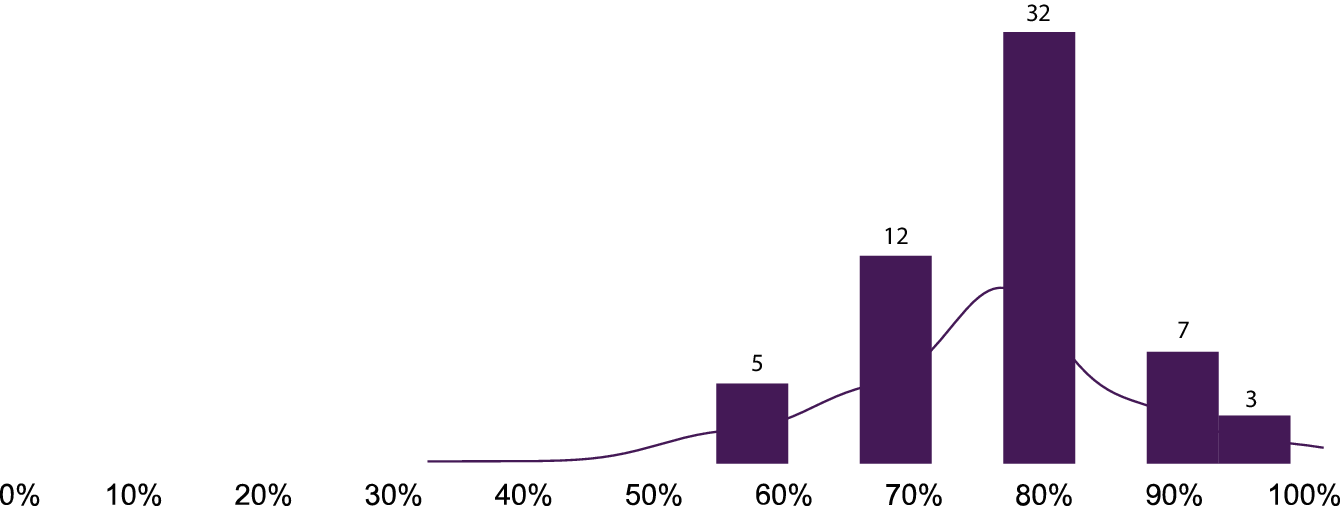}
    \caption{Study quality}
    \label{fig:studyQuality}
\end{figure}
More than 85\% of the studies satisfy the requirements listed from Q1 to Q6. However, over 80\% of the studies fail to provide source code or details related to the computing environment which makes the system difficult to reproduce. This leads to an overall reduction of the average quality score to 76.5\% - Figure \ref{fig:studyQuality}. 

\begin{figure}[htbp]%
    \centering
    \subfloat[\centering Publication type]{{\includegraphics[scale=0.65]{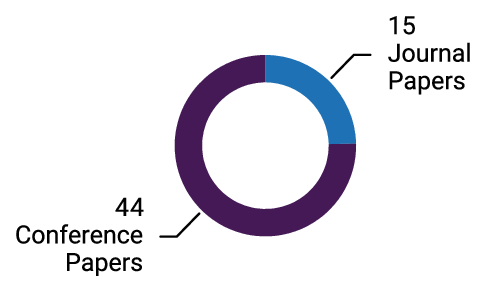} }}%
    \qquad
    \subfloat[\centering Published year]{{\includegraphics[scale=0.55]{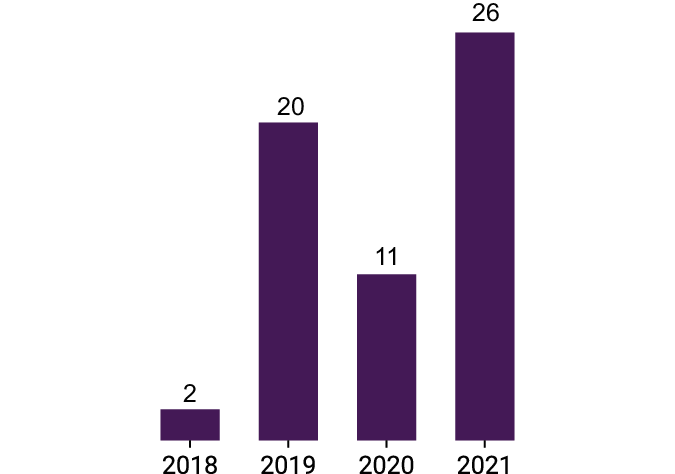} }}%
    \caption{Publications selected for inclusion}%
    \label{fig:publication}%
\end{figure}

Finally, a deep reading of each of the eligible studies led to 59 studies selected for inclusion. Data extraction was performed for each of the features outlined in Table \ref{tab:features}. For acceptable values of individual features see Appendix \ref{sec:allowedvalues:appendix}. The lists of neural and symbolic terms referenced in the table constitute the glossary items learned from conducting the selection process. Figure \ref{fig:publication}(a) shows the breakdown of conference papers vs journal articles, and Figure \ref{fig:publication}(b) shows the number of studies published each year. 

\begin{table}[htbp]
    \footnotesize
    \centering
    \caption{Data extraction features} \label{tab:features}
    \renewcommand{\arraystretch}{1.5}
    \begin{tabular}{@{}>{\raggedright\arraybackslash}p{4.5cm}>{\raggedright\arraybackslash}p{7.5cm}}
    \toprule
       \textbf{Feature}  & \textbf{Description} \\ \midrule
        Business application & The stated objective or application of the proposed study. Often this is an NLP task, but this is not a requirement (i.e., \say{Medical decision support})  \\ 
        Technical application &  Type of model output    \\ 
        Type of learning & Indicates learning method (supervised, unsupervised, etc.) \\ 
        Knowledge representation & One of four categories: Rules, Logic, Frames, and  Semantic networks \\ 
        Type of reasoning & Indicates whether knowledge is represented implicitly (embedded) or explicitly (symbolic) \\
        Language structure & Indicates whether linguistic structure is leveraged to facilitate reasoning \\
        Relational structure & Indicates whether relational structure is leveraged to facilitate reasoning (e.g., part-of-speech tags, named entities, etc.) \\
        Symbolic terms & List of symbolic techniques used by the models\\ 
        Neural terms & List of neural architectures used by the models\\ 
        Datasets & List of all datasets used for evaluation \\ 
        Model description & Describes model architecture schematically \\ 
        Evaluation Metrics & Evaluation metrics reported by the authors \\ 
        Reported score & Model performance reported by the authors \\ 
        Contribution & Novel contribution reported by the authors\\         
        Key-intake & Short description of the study \\ 
        isNeSy & Indicates whether the authors label their study as Neuro-Symbolic \\
        NeSy goals & For each of the goals listed in Section \ref{sec:introduction}, indicates whether the goal is met as reported by the authors  \\
        Kautz category &  List of categories from Kautz's taxonomy \\ 
        NeSy category & List of categories from the proposed nomenclature \\ 
        Study quality & Percentage of positive answers in the quality assessment questionnaire \\ 
    
    \bottomrule
    \end{tabular} 
\end{table}

\section{Results, Data Analysis, Taxonomies}\label{results}
We perform quantitative data analysis based on the extracted features in Table \ref{tab:features}. Each study was labeled with terms from the aforementioned glossary, and each term in the glossary was classified as either symbolic, or neural. A bi-product of this process are two taxonomies built bottom-up of concepts relevant to the set of studies under review. The two taxonomies are a reflection of the definition of NeSy provided earlier: \say{the combination of deep learning and symbolic reasoning.} To make this definition more precise, we limit the type of combination that qualifies as neuro-symbolic. Specifically, the sub-symbolic and symbolic components must be integrated in a way such that one informs the other. By way of counter example, a system which is made up of two independent symbolic and sub-symbolic components would not be considered NeSy if there is no interaction between them. For example, while a system where one component is used to process one type of data, and the other is used to process another type of data may be an effective software pipeline design, we do not consider this type of solution neuro-symbolic as the two components do not interact in any way. Thus the definition becomes \say{the \textit{integration} of deep learning and symbolic reasoning.} It should be noted, that these terms are not always consistently defined in the literature. For example, in a much earlier survey, \cite{Bader_Hitzler_2005} split the interrelation (type of combination) of neuro-symbolic systems into \textit{hybrid} and \textit{integrated}, whereas we use the term \textit{integrated} to cover both. 

On the learning side, we have neural architectures (described in Section \ref{sec:nn_taxonomy}), and on the symbolic reasoning side we have knowledge representation (described in Section \ref{sec:kr_taxonomy}). These results are rendered in Table \ref{table:nexytaxonomy}, with the addition of color representing a simple metric, or \textit{promise score}, for each study. The promise score is simply the number of goals reported to have been satisfied by the solution in the study.

\subsection{Exploratory Data Analysis} 
\label{sec:dataAnalysis}
We plot the relationships between the features extracted from the studies, and the goals from section \ref{sec:contributions} in an effort to identify any correlations between them, and ultimately to identify patterns leading to higher promise scores.

\subsubsection{Business and Technical Applications}
\begin{figure}[htbp]
    \centering
    \includegraphics[scale=0.45]{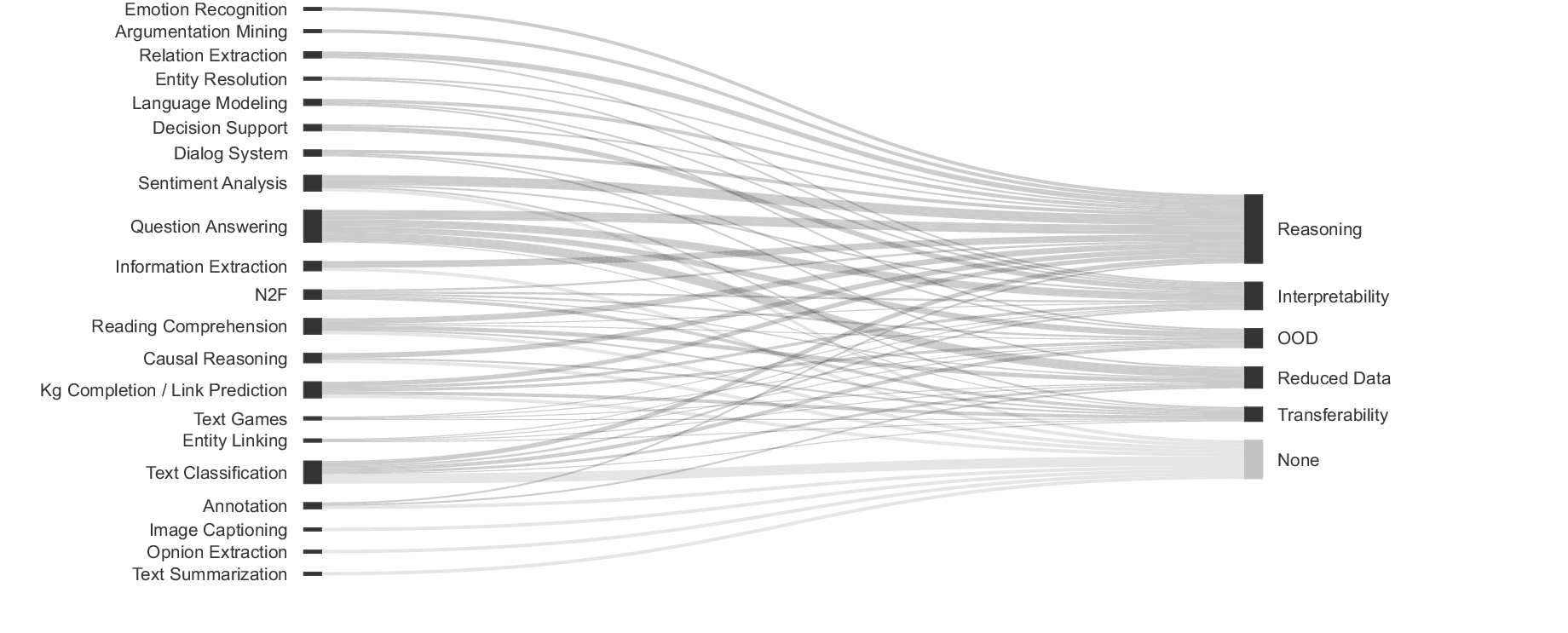}
    \caption{Relationship between Business Applications and NeSy Goals. Question answering is the most frequently occurring task, and is associated mainly with reasoning, reduced data, and to a lesser degree, interpretability.}
    \label{fig:buc-to-goals}
\end{figure}

The \textit{business application} is the stated application, or objective, of a given study. It is often but not always an NLP task, such as \textit{text classification}, or \textit{sentiment analysis}. It should be noted that in this example, sentiment analysis is a type of text classification, but while one author's stated objective is specific to sentiment, another author may be interested in solving for text classification in general. As such there is no particular hierarchy or taxonomy associated with business applications. The relationship between all tasks, or business applications, and NeSy goals is shown in Figure \ref{fig:buc-to-goals}.

The business application largely determines the type of model output, or what we term \textit{technical application}. Most business applications are associated with a single (or at most two) technical applications. The exceptions being \textit{question answering} and \textit{reading comprehension}, which have been tackled as both inference and classification problems, or with the goal of information extraction or text generation. Question answering is the most frequently occurring task, and is associated mainly with reasoning, reduced data, and to a lesser degree, interpretability. On a philosophical level this seems somewhat disappointing, as one would hope that in receiving an answer, one could expect to understand why such an answer was given. 

\begin{figure}[htbp]
    \centering
    \subfloat[\centering Top Business Applications ]{{\includegraphics[scale=0.4]{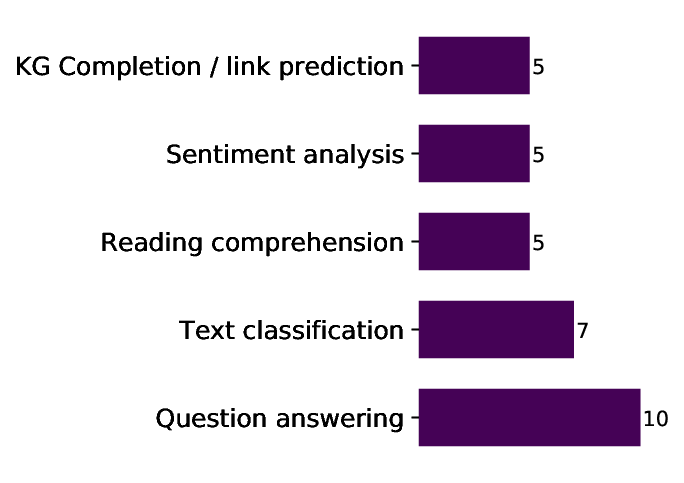} }}%
    \qquad
    \subfloat[\centering Technical Applications (model output)]{{\includegraphics[scale=0.5]{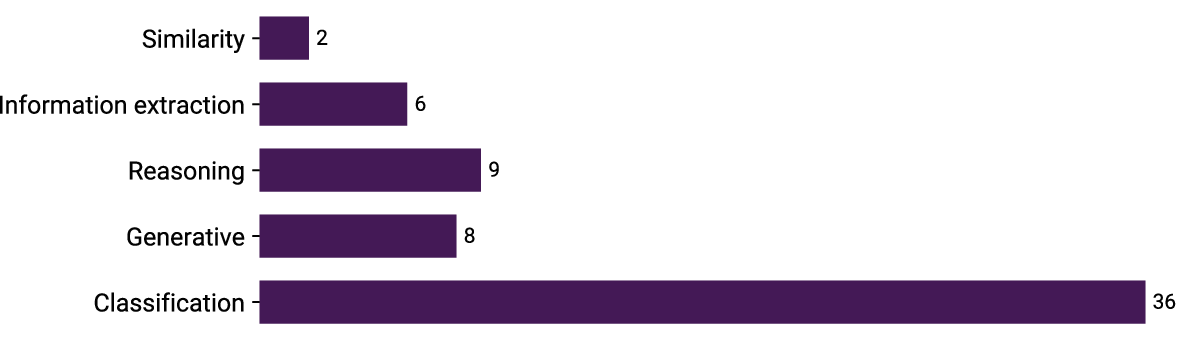} }}%
    \caption{Number of studies in each application category}%
    \label{fig:busines-tech-apps}%
\end{figure}

\begin{figure}[htbp]
    \centering
    \includegraphics[scale=0.45]{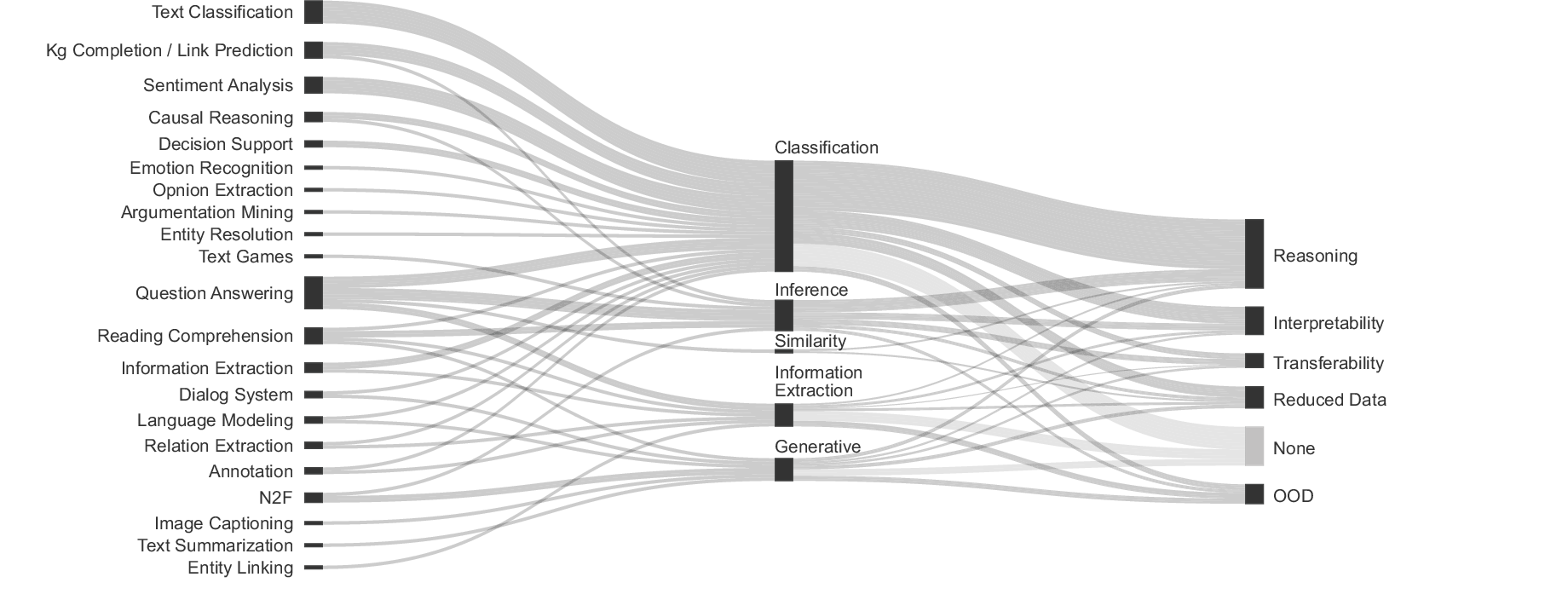}
    \caption{Relationship between Business Applications, Technical Applications, and NeSy Goals}
    \label{fig:buc-techapp-goal}
\end{figure}

For completeness, the number of studies representing the technical applications and most frequently occurring business application is given in Figure \ref{fig:busines-tech-apps}, while Figure \ref{fig:buc-techapp-goal} illustrates the relationship between business applications, technical applications, and goals.


\subsubsection{Type of learning}
\begin{figure}[htbp]
    \centering
    \includegraphics[scale=0.45]{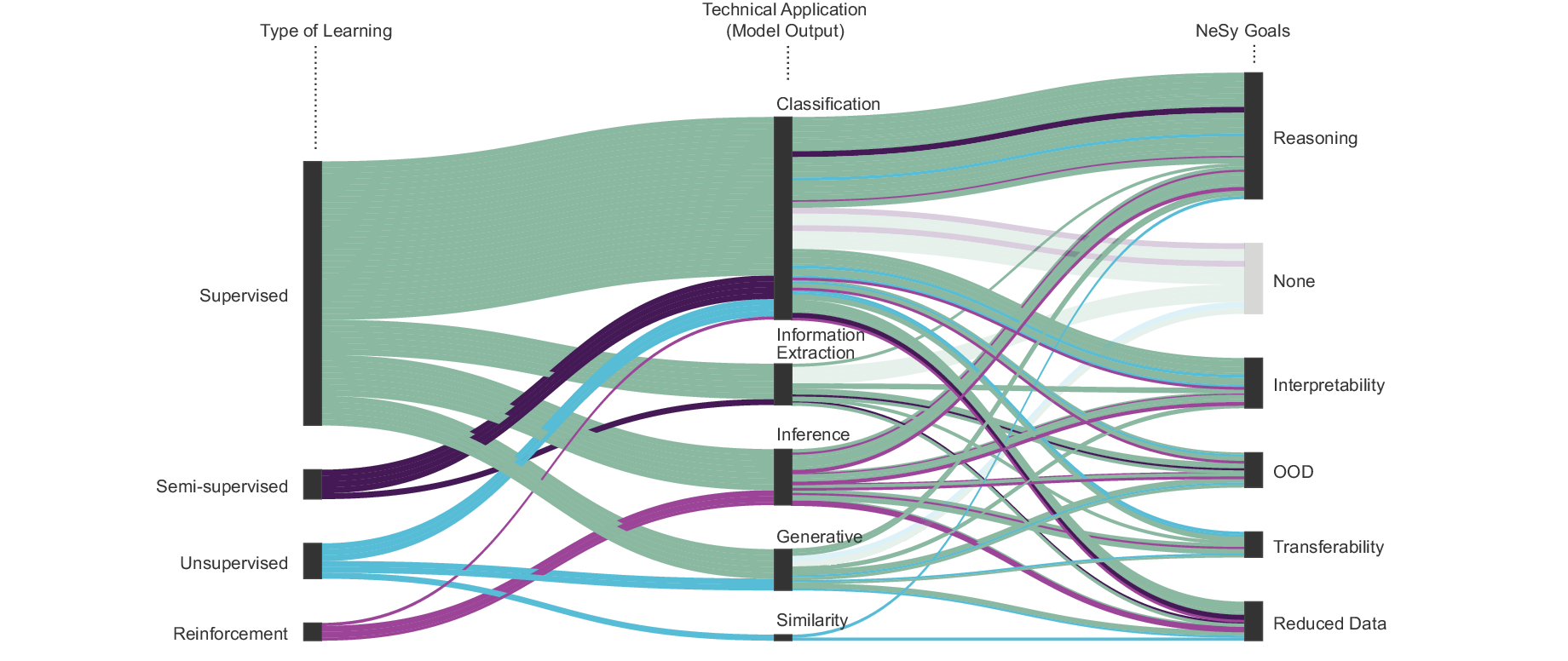} %
    \caption{Relationship between Learning Type, Technical Application, and NeSy Goals. It is clear that supervised approaches dominate the field, are applied across a variety of technical applications, and there is no clear winner when it comes to goals.}%
    \label{fig:learning_type-tech_app-goals}%
\end{figure}

Machine learning algorithms are classified as supervised, unsupervised, semi-supervised, curriculum or reinforcement learning, depending on the amount and type of supervision required during training \cite{kang2018machine, bonaccorso2017machine, bengio2009curriculum}. Figure \ref{fig:learning_type-tech_app-goals} demonstrates that the supervised method outnumbers all other approaches.

\subsubsection{Implicit vs Explicit Reasoning}\label{sec:implicitVSexplit}
The subset of tasks belonging to Natural Language Understanding (NLU) and Natural Language Generation (NLG) are often regarded as more difficult, and presumed to require reasoning.  Given that reasoning was one of the keywords used for search, it is not surprising that many studies report reasoning as a characteristic of their model(s).

How reasoning is performed often depends on the underlying representation and what it facilitates. Sometimes the representations are obtained via explicit rules or logic, but are subsequently transformed into non-decomposable embeddings for learning. As such, we can say that any reasoning during the learning process is done implicitly. Studies utilizing Graph Neural Networks (GNNs) \cite{Saveleva_Petukhova_Mosbach_Klakow_2021,Zhang_Wang_Yu_Wang_Wang_Jiang_Lim_2021,Chen_Xu_Cheng_2020,gu2019local,Lemos2020647,Zhou20212015,Huo2019159} would also be considered to be doing reasoning implicitly. The majority of the studies doing implicit reasoning leverage linguistic and/or relational structure to generate those internal representations. These studies meet 53 out of a possible 180 NeSy goals, where 180 = \#goals * \#studies, or 29.4\%. For reasoning to be considered explicit, rules or logic must be applied during or after training. Studies which implement explicit reasoning perform slightly better, meeting 51 out of 135 goals, or 37.8\% and generally require less training data. Additionally, 4 studies implement both implicit and explicit reasoning, at a NeSy promise rate of 40\%. Of particular interest in this grouping is Bianchi et al. \cite{Bianchi2019161}'s implementation of Logic Tensor Networks (LTNs), originally proposed by Serafini and Garcez in \cite{Serafini_Garcez_2016}. \say{LTNs can be be used to do after-training reasoning over combinations of axioms which it was not trained on. Since LTNs are based on Neural Networks, they reach similar results while also achieving high explainability due to the fact that they ground first-order logic} \cite{Bianchi2019161}. Also in this grouping, Jiang et al.  \cite{Jiang2020} propose a model where embeddings are learned by following the logic expressions encoded in huffman trees to represent deep first-order logic knowledge. Each node of the tree is a logic expression, thus hidden layers are interpretable.

\begin{figure}[htbp]
    \centering
    \includegraphics[scale=0.45]{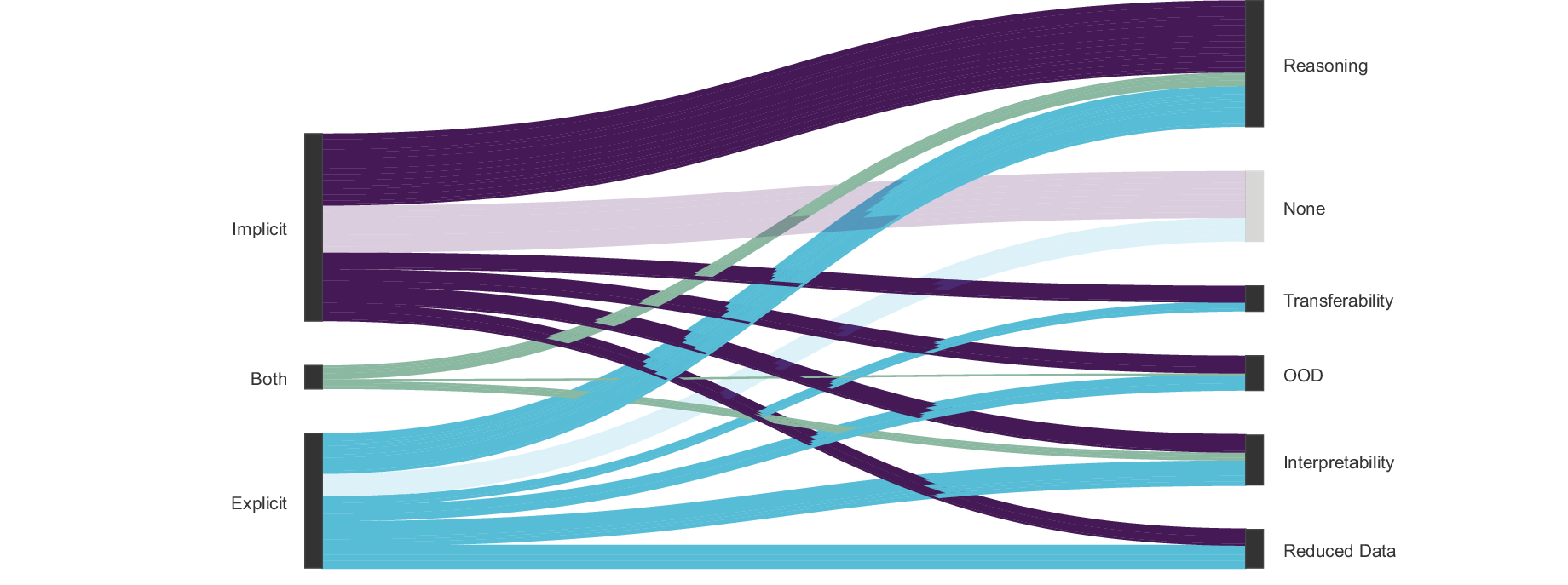}
    \caption{Type of Reasoning and Goals. Around half, 48\%, of studies where reasoning is performed explicitly mention interpretability as a feature. While nearly a third of studies performing reasoning implicitly do not meet any of the NeSy promises identified for this review.}
    \label{fig:implicit_explicit-goals}
\end{figure}

\begin{figure}[htbp]
    \centering
    \includegraphics[scale=0.45]{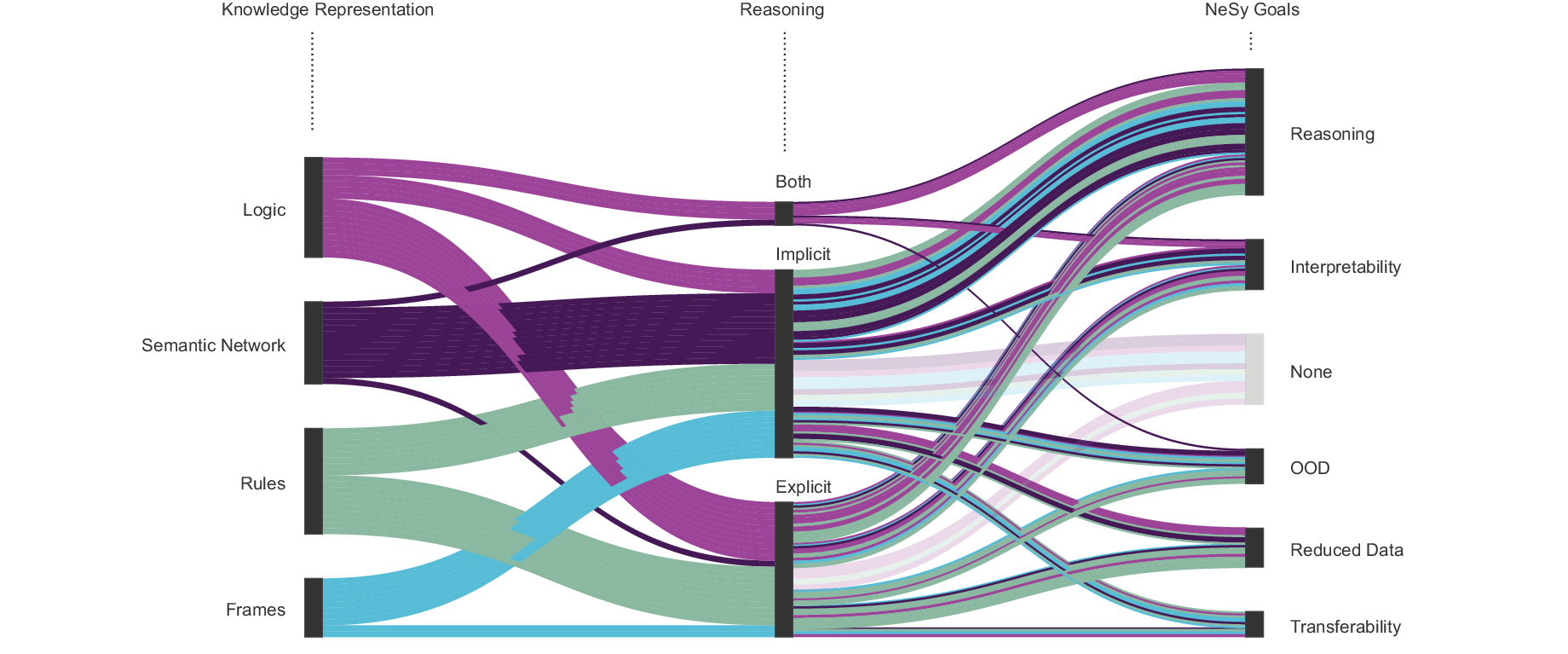}
    \caption{Knowledge Representation, Type of Reasoning, and Goals. What is noteworthy, is that when Semantic Networks are utilized, reasoning is almost always done implicitly. The two exception are \cite{Bianchi2019161}, and \cite{Zhang_Wang_Yu_Wang_Wang_Jiang_Lim_2021}. However, \cite{Bianchi2019161} utilizes FOL for explicit reasoning rather than its network component. On the other hand, \cite{Zhang_Wang_Yu_Wang_Wang_Jiang_Lim_2021} generate a novel interpretable reasoning graph as the output of their model.}
    \label{fig:implicit_explicit-kr-goals}
\end{figure}

Figure \ref{fig:implicit_explicit-goals} shows the relationship between implicit \& explicit reasoning and goals, while the relationship between knowledge representation, type of reasoning, and goals is shown in Figure \ref{fig:implicit_explicit-kr-goals}. 

\subsubsection{Linguistic and Relational Structure}

\begin{figure}[htp]
\centering
\includegraphics[width=1\textwidth]{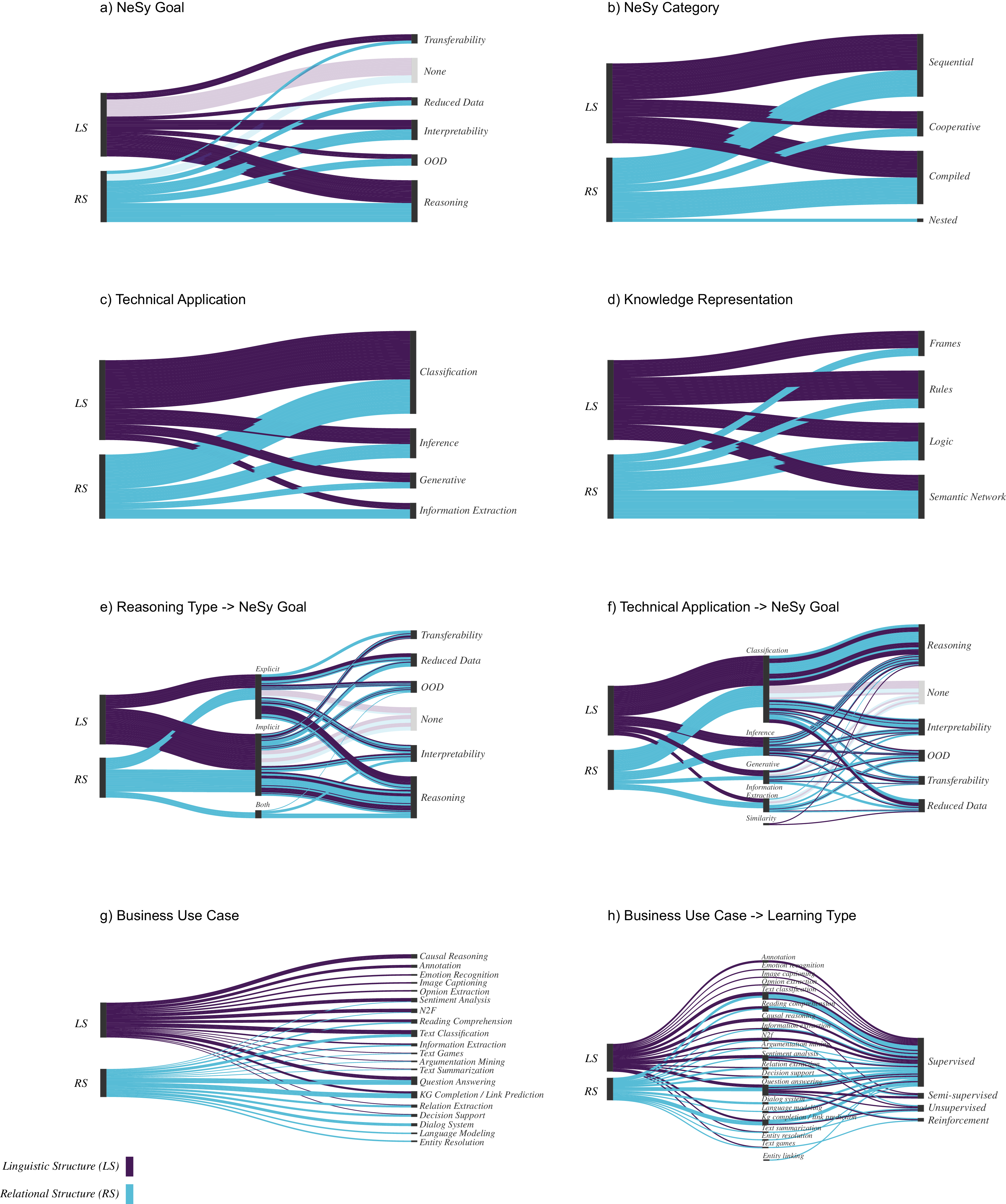}
\caption{Relationships between leveraged structures and extracted features. As can be seen in a), e), and f), studies leveraging linguistic structures often do not meet any NeSy goals, which runs counter to our original hypothesis. Further investigation into this phenomenon may be warranted. \textit{ Note: studies which do no leverage either structure are not shown}}
\label{fig:leverage}
\end{figure}

In the previous section we described how linguistic and relational structures can be leveraged to generate internal representations for the purpose of implicit reasoning. Here we plot the relationships between these structures and other extracted features and their interactions - Figure \ref{fig:leverage}. Perhaps the most telling chart is the mapping between structures and goals, where many the studies leveraging linguistic structure do not meet any of the goals. This runs counter to the intuition that language is a natural fit for NeSy. 

\subsubsection{Datasets and Benchmarks}\label{sec:datasets}


Each study in our survey is based on a unique dataset, and a variety of metrics. Given that there are nearly as many business applications, or tasks, as there are studies, this is not surprising. As such it is not possible to compare the performance of the models reviewed. However, this brings up an interesting question, and that is how one might design a benchmark for NeSy in the first place. A discussion about benchmarks at the IBM Neuro-Symbolic AI Workshop 2022\footnote{\label{ibm}\url{https://video.ibm.com/recorded/131288165}} resulted in general agreement that the most important characteristic of a good benchmark for NeSy is in the diversity of tasks tackled. Gary Marcus pointed out that current benchmarks can be solved extensionally, meaning they can be \say{gamed}.\footnote{\url{https://video.ibm.com/recorded/131288165} time-marker 43:00} In other words, with enough attempts, a model can become very good at a specific task without solving the fundamental reasoning challenge. In essence, this akin to over-fitting on the test set. The phenomenon can be exposed when adversarial examples are introduced such as described in \cite{jia_liang_2017_adversarial}, or through the observation that spurious correlations can be introduced in the annotation process as per \cite{gururangan_etal_2018_annotation}. This leads to models which are not able to generalize out of the training distribution. In contrast, to solve a task intensionally is to demonstrate \say{understanding} which is transferable to different tasks. This view is controversial with advocates of purely connectionist approaches arguing that \say{understanding} is not only ill defined, but also a moving target \cite{Garcez_Lamb_2020} - every time we solve for the current definition of understanding, the definition is revised to have to meet a higher bar. So instead of worrying about the semantics of \say{understanding}, the panelists agreed that to make the benchmarks robust to gaming is to build in enormous variance in the types of tasks they tackle. Taking this a step further, Luis Lamb\footnote{\url{https://video.ibm.com/recorded/131288165} time-marker 50:00} proposed that instead of designing benchmarks for testing models, we should be designing challenges which encourage people to work on important real world problems. For a deeper dive, see the ACL-2021 Workshop on Benchmarking: Past, Present and Future (BPPF)\footnote{\url{https://github.com/kwchurch/Benchmarking_past\_present\_future\#S1}}, where some of the same issues pertaining specifically to NLP and NLU were discussed, as well as the challenges in interpreting performances across datasets, models, and with the evolution of language and context over time.




\subsection{Taxonomies: Neural, Symbolic, \& Neuro-Symbolic}\label{sec:glossary}

\subsubsection{Neural}\label{sec:nn_taxonomy}
\begin{figure}[htp]
    \centering
    \includegraphics[width=1\textwidth]{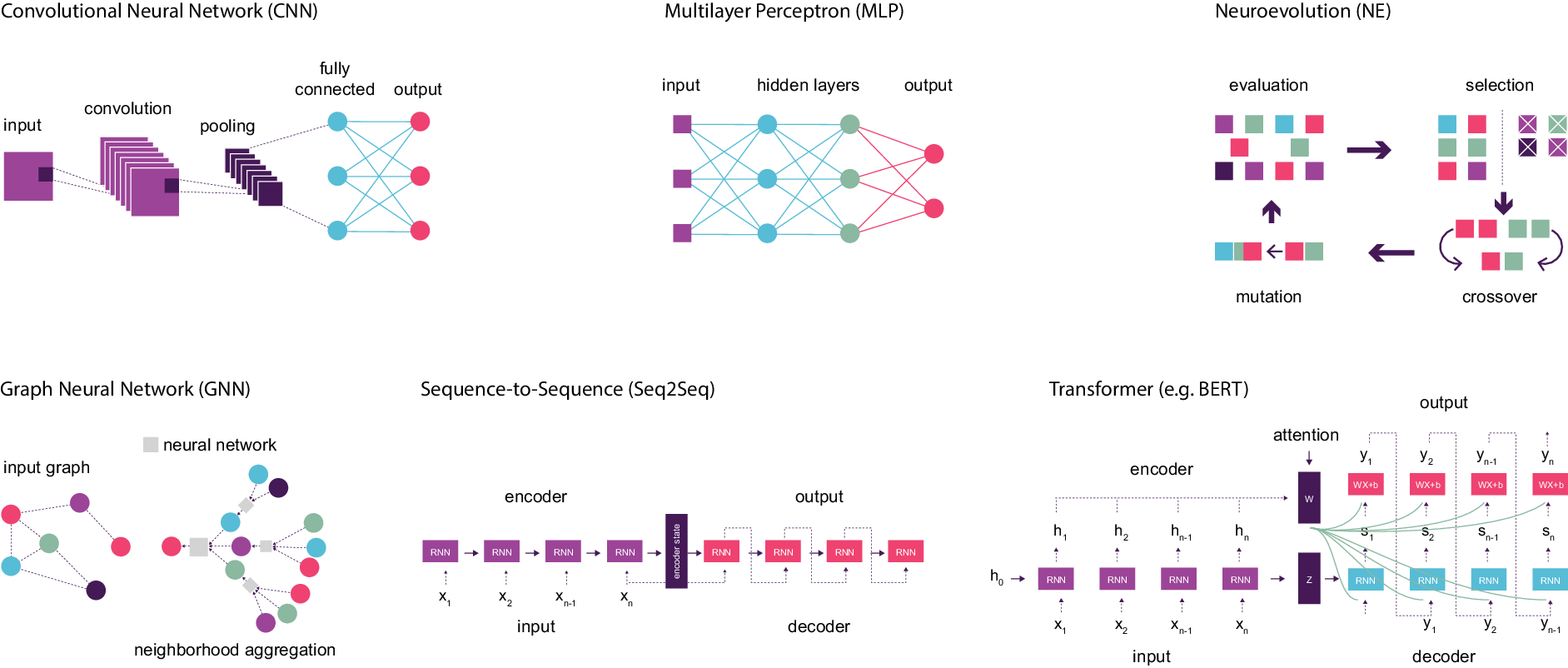}
    \caption{Neural architectures represented in Table \ref{table:nexytaxonomy}}
    \label{fig:Neural-architectures}
\end{figure}

\begin{figure}[htp]
    \centering
    \includegraphics[width=1\textwidth]{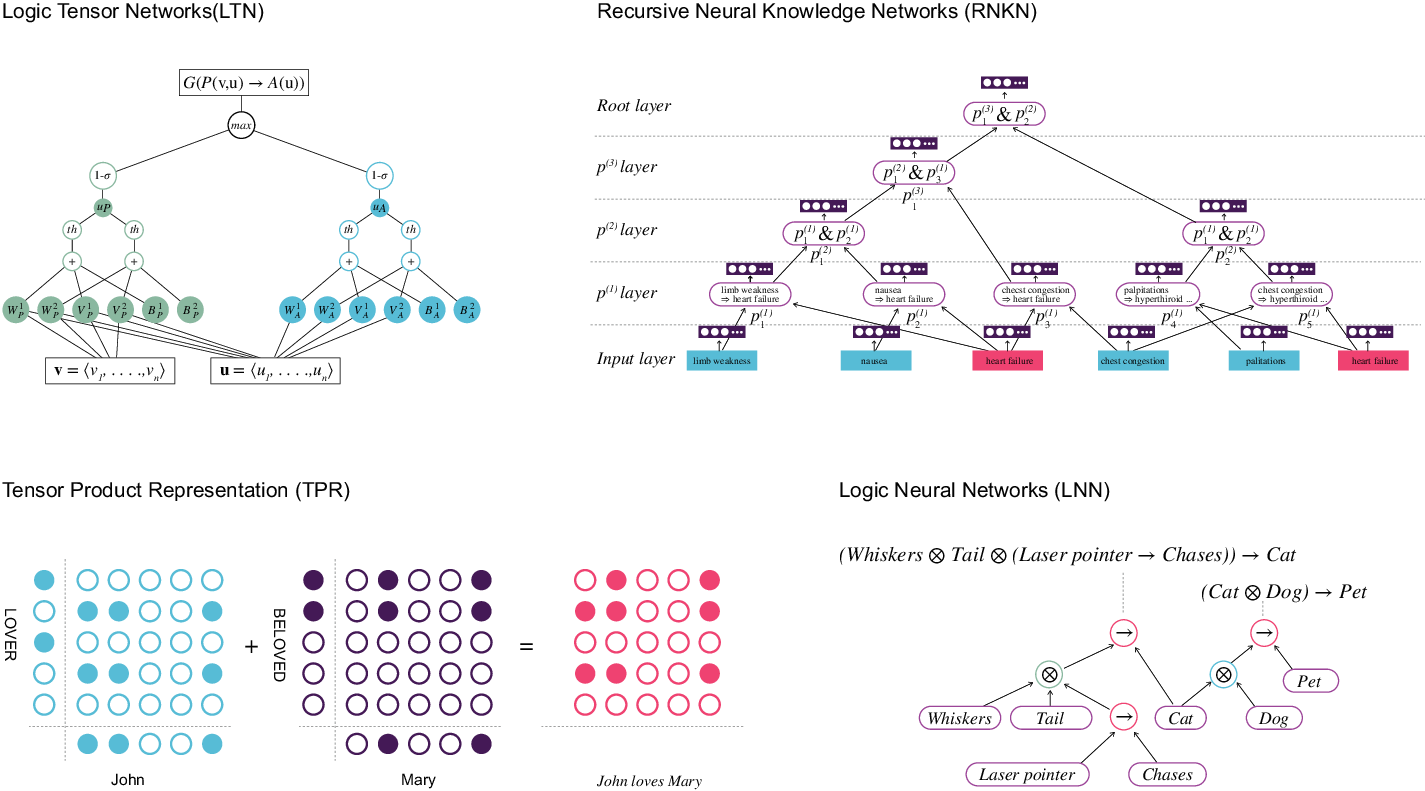}
    \caption{Neuro-symbolic architectures represented in Table \ref{table:nexytaxonomy}}
    \label{fig:nesy-architectures}
\end{figure}

In the main, the extracted neural terms refer to the neural architecture implemented in a given study. We group these into higher level categories such as Linear models, Early generation (which includes CNNs), Graphical models, Sequence-to-Sequence - Figure \ref{fig:Neural-architectures}.
We have included Transformers in the Sequence-to-Sequence category as the original architecture was an encoder/decoder with attention. It should be noted that not all Transformers since then employ both an encoder and decoder, or generate sequences. What they have in common is the attention mechanism described in the seminal paper Attention Is All You Need, by Vaswami et al. \cite{vaswani2017attention} which dramatically advanced NLP research. We also include here Neuro-Symbolic architectures such as Logic Tensor Networks (LTN), Recursive Neural Knowledge Networks (RNKN), Tensor Product Representations (TPRs), and Logical Neural Networks (LNN) because they are suitable to optimization via gradient descent - Figure \ref{fig:nesy-architectures}.
We include one study \cite{Skrlj2021989} which does not implement gradient descent, but rather Neuroevolution (NE). Neuroevolution involves genetic algorithms for learning neural network weights, topologies, or ensembles of networks by taking inspiration from biological nervous systems \cite{miikkulainen:encyclopedia10-ne,Lehman_Miikkulainen_2013}. Neuroevolution is often employed in the service of Reinforcement Learning (RL). Studies which do not specify a particular architecture are categorised as Multilayer Perceptron (MLP).

\subsubsection{Symbolic}\label{sec:kr_taxonomy}
The definition we adopted states that NeSy is
\textit{the  integration  of  deep learning  and  symbolic  reasoning}. Our neural taxonomy described above reflects the \textit{deep learning} component. For the \textit{symbolic reasoning} component we utilize four common Knowledge Representation (KR) categories: 1) production rules, 2) logical representation, 3) frames, and 4) semantic networks \cite{davis1993knowledge, bench2014knowledge, levesque1986knowledge,Brachman_Levesque_2004, travis1990knowledge, Sowa_1991}. The following definitions are merely a glimpse at each of these topics, in order to provide a basic intuition.

\begin{enumerate}
    \item \textit{Production rules} - A production rule is a two-part structure comprising an antecedent set of conditions and a consequent set of  actions \cite{Brachman_Levesque_2004}. We usually write a rule in this form: 
    \vspace{-2\baselineskip}
    \begin{multicols}{2}
      \begin{equation*}
       \begin{aligned}
            IF \ conditions \ THEN \ actions \\
       \end{aligned}
      \end{equation*}
      \vfill\null
      \begin{equation*}
        \begin{aligned}
           ex) \  IF \ Bird \ THEN \ fly \\
       \end{aligned}
      \end{equation*}
    \end{multicols}

    \item \textit{Logical representation} - Logic is the study of entailment relations—languages, truth conditions, and rules of inference. \cite{Brachman_Levesque_2004, Dyer}. A logic includes:
    \begin{itemize}
        \item \textbf{Syntax}: specifies the symbols in the language and how they can be combined to form sentences. Hence facts about the world are represented as sentences in logic.
        \item \textbf{Semantics}: specifies what facts in the world a sentence refers to. Hence, also specifies how you assign a truth value to a sentence based on its meaning in the world. A fact is a claim about the world, and may be true or false.
        \item \textbf{Inference Procedure (reasoning)}: mechanical method for computing (deriving) new (true) sentences from existing sentences.
    \end{itemize}
    The sentence "Not all birds can fly" in First Order Logic (FOL) looks like:
    \begin{equation*}
        \begin{aligned}
        \neg (\forall x Bird(x) \rightarrow Fly(x))
        \end{aligned}
    \end{equation*} 
    FOL is by no means the only choice, but as per \cite{Brachman_Levesque_2004} it is a simple and convenient one for the sake of illustration. Natural Logic (NL) for example, is a formal proof theory built on the syntax of human language, which can be traced to the syllogisms of Aristotle \cite{Byszuk_Wozniak_2020}. \say{For better or worse, most of the reasoning that is done in the world is done in natural language. And correspondingly, most uses of natural language involve reasoning of some sort. Thus it should not be too surprising to find that the logical structure that is necessary for natural language to be used as a tool for reasoning should correspond in some deep way to the grammatical structure of natural language} \cite{Lakoff_1970}. Implementations and extensions include \cite{maccartney2007natural,maccartney_manning_2009_extended,Angeli_Manning_2014,manning-etal-2014-stanford}. Real-valued logics are often utilized in machine learning because they can be made differentiable and/or probabilistic \cite{Serafini_dAvila_Garcez_2016} - first introduced by Łukasiewicz at the turn of the 20th century \cite{McCall_1973,Harder_Besold_2018}). Other, logic-based cognitive modelling approaches such as non-monotonic logic, attempt to deal with the complexities of human reasoning, epistemology, and defeasible inference \cite{sep_logic_nonmonotonic}.

    \item \textit{Frames} - Frames are objects which hold entities, their properties and methods. An individual frame schema looks like this:
    \vspace{-3\baselineskip}
        \begin{multicols}{2}
            \begin{equation*}
                \begin{aligned}
                (Frame-name  \\
                & <slot-name1 \ filler1> \\
                & <slot-name2 \ filler2> \\
                & ...) \\
                \end{aligned}    
            \end{equation*}
            \vfill\null
            \begin{equation*}
                \begin{aligned}
                (Penguin  \\
                & canFly: \ 0 \\
                & isA: \ ''Bird'' \\
                & ...) \\
                \end{aligned}    
            \end{equation*}
        \end{multicols}
    The frame and slot names are atomic symbols; the fillers are either atomic values (like numbers or strings) or the names of other individual frames \cite{Brachman_Levesque_2004}. This is similar to Object Oriented Programming (OOP), where the frame is analogous to the object, and slots and fillers are properties and values respectively.
    
    \item \textit{Semantic networks} - A semantic network is a structure for representing knowledge as a pattern of interconnected nodes and edges \cite{Sowa_1991}. A Frame network is a kind of semantic network where nodes are frames, and edges are the relationships between nodes. An example of a semantic network often used in NLU systems is WordNet\footnote{\url{https://wordnet.princeton.edu/}} - a lexical database of English - Figure \ref{fig:wordnet}. Today semantic networks are more often referred to as Knowledge Graphs (KGs).\footnote{This term was popularized after Google introduced contextual information to search results from their semantic network under the brand name \textit{Knowledge Graph}  \url{https://blog.google/products/search/introducing-knowledge-graph-things-not/}.}
\begin{figure}[htp]
    \centering
    \includegraphics[scale=0.4]{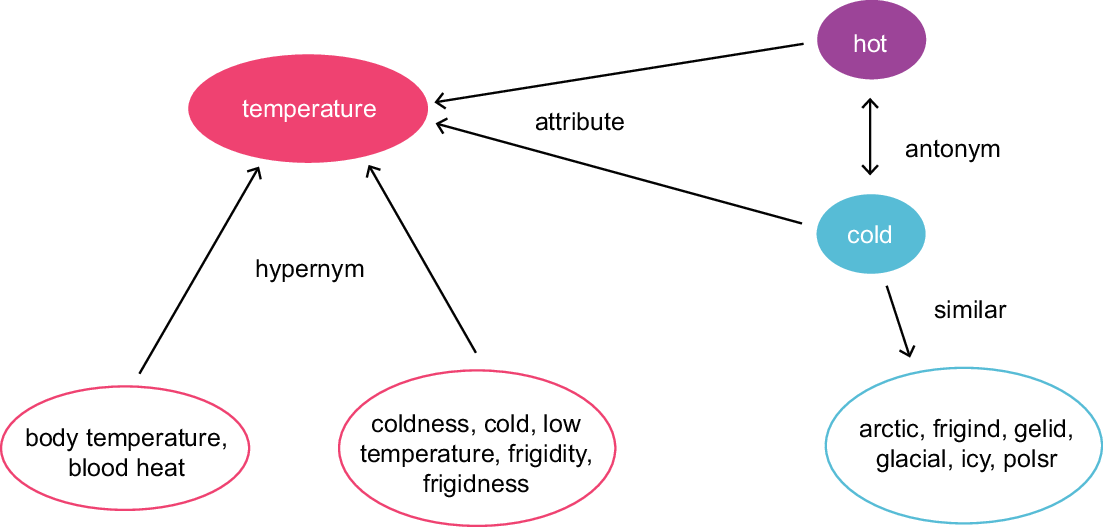}
    \caption{English WordNet subgraph \cite{McCrae_Rudnicka_Bond_2021}}
    \label{fig:wordnet}
\end{figure}
\end{enumerate}


Table \ref{table:nexytaxonomy} shows which studies combine which of the above neural (\ref{sec:nn_taxonomy}) and symbolic (\ref{sec:kr_taxonomy}) categories as well as the number of NeSy goals satisfied.

\begin{table*}[htbp]

\caption{ 
    \centering 
    \text{Neural \& Symbolic Combinations} 
    \hspace{\textwidth}
    \colorbox{purple_1}{\color{black} 1} \colorbox{purple_2}{\color{white} 2} \colorbox{purple_3}{\color{white} 3} \colorbox{purple_4}{\color{white} 4} 
    \colorbox{black}{\color{white} 5}
    \text{Number of NeSy goals satisfied out of the 5 described in Section \ref{sec:contributions}.} 
    \hspace{0.5\textwidth}
    \textit{Note: some studies use multiple techniques.}
    } 
\label{table:nexytaxonomy}
\begin{tabular}{@{}llllll}
\toprule
\multicolumn{2}{@{}l}{}  &  \multicolumn{4}{c}{{Knowledge Representation}}  \\ 
\cmidrule{3-6}
\multicolumn{2}{@{}l}{\multirow{2}{*}{{}}} &
  {Frames} &
  {Logic} &
  {Rules} &
  {\begin{tabular}[c]{@{}l@{}}Semantic \\ network\end{tabular}} \\ 
  \cmidrule[.3pt]{1-6}
    Linear Models &
  {SVM}  &  
  &
  &
  \begin{tabular}[c]{@{}l@{}} \colorbox{purple_1}{\cite{DSouza201990}}   \end{tabular}
  &
  \colorbox{purple_1}{\cite{Hussain20181662}} \colorbox{purple_4}{\color{white}\cite{Skrlj2021989}} \\ 
  \cmidrule[.3pt]{1-6}
  {\multirow{2}{*}{{\begin{tabular}[c]{@{}l@{}}Early  \\ Generation\end{tabular}}}}  &
   {MLP} &
   \begin{tabular}[c]{@{}l@{}}
  \cite{Cui2021419} \\
  \colorbox{purple_2} {\color{white}{\cite{Xu_Li_2019, cowen2019neural}}} \end{tabular} 
  & \begin{tabular}[c]{@{}l@{}}
  \colorbox{purple_1}{\cite{Bounabi2021229, Es-Sabery202117943}} \\ \colorbox{purple_2} {\color{white}{\cite{Lima_Espinasse_Freitas_2019}}}
  \end{tabular} 
  &
  \begin{tabular}[c]{@{}l@{}} \colorbox{purple_1}{\cite{Chen2021328}} \\ \colorbox{purple_3}{\color{white}\cite{Yao201842}} \end{tabular}
  &
  \colorbox{purple_1}{{\cite{Zhou20212015}}}
  \\ 
  \cmidrule[.05pt]{2-6} 
  & {CNN} 
  &
  \cite{Tato2019623} 
  &
  \colorbox{purple_1}{\cite{Es-Sabery202117943}}  \colorbox{purple_2}{\color{white}{ \cite{Chaturvedi2019264}}}
  & 
  \colorbox{purple_1}{\cite{Ayyanar2019}} 
  &  \begin{tabular}[c]{@{}l@{}}
  \cite{Gong202030885} 
   \end{tabular}
  \\ 
  \cmidrule[.3pt]{1-6}  
  
  \multirow{2}{*}{{\begin{tabular}[m]{@{}l@{}}Graphical\\Models\end{tabular}}}  &
  {DBN} &
  &
  \colorbox{purple_2}{\color{white}{ \cite{Chaturvedi2019264}}} &
  &
  \\ 
  \cmidrule[.05pt]{2-6}
  &
  {GNN} &
  \colorbox{purple_1}{\cite{Lemos2020647}} 
  &
  &\colorbox{purple_1}{ \cite{Saveleva_Petukhova_Mosbach_Klakow_2021}} \colorbox{purple_3}{\color{white}{\cite{Huo2019159}}} 
  &  \begin{tabular}[c]{@{}l@{}} 
  \colorbox{purple_1}{{\cite{Chen_Xu_Cheng_2020, Zhou20212015}}} \\ \colorbox{purple_2}{\color{white}{\cite{gu2019local}}} 
  \colorbox{purple_4}{\color{white} \cite{Zhang_Wang_Yu_Wang_Wang_Jiang_Lim_2021}} 
  \end{tabular} \\ 
  \cmidrule[.3pt]{1-6}
  
 \multirow{2}{*}{{\begin{tabular}[m]{@{}l@{}}Sequence-\\to-Sequence\end{tabular}}}  
 
 & {RNN} 
 & 
\begin{tabular}[c]{@{}l@{}}\cite{Brasoveanu2019656,  Tato2019623} \\ \colorbox{purple_1} {  \cite{Hu_Wei_Huai_2021}  }
\colorbox{purple_3}{\color{white}{\cite{Chen20201544}}}
\end{tabular} 
& 
\begin{tabular}[c]{@{}l@{}}\cite{Graziani2019185}, \cite{ Gupta_Ghosal_Ekbal_2021}, \cite{Langton_Srihasam_2021} \\
\colorbox{purple_1}{\cite{Fazlic20191025}}   \colorbox{purple_2}{\color{white}{\cite{Chaturvedi2019264, Schon2019293}}} \\ \colorbox{purple_2}{\color{white}{\cite{Pacheco_Goldwasser_2021}}}
\end{tabular}
& 
\begin{tabular}[c]{@{}l@{}}\colorbox{purple_2}{\color{white}{\cite{Amin2019133}}} \colorbox{purple_2}{\color{white}{\cite{AltszylerBBBV21}}} \colorbox{purple_2}{\color{white}{\cite{Sutherland2019}}} \\ \colorbox{purple_2}{\color{white}{\cite{Demeter20207634}}} \colorbox{purple_2}{\color{white}{\cite{Zhou_Richardson_2021}}} \colorbox{purple_2}{\color{white}{\cite{Qin_Liang_Hong_Tang_Lin_2021}}} \\
\colorbox{purple_3}{\color{white}{\cite{Sen_Danilevsky_Li_2020}}} \colorbox{purple_4}{\color{white}\cite{Mao2019}} 
\end{tabular}
&
\begin{tabular}[c]{@{}l@{}}\cite{ Kouris_Alexandridis_Stafylopatis_2021, Gong202030885} \\
\cite{Pinhanez_Cavalin_Alves_2021} \colorbox{purple_1}{\cite{Liu2021260}} \\  \colorbox{purple_2}{\color{white}{\cite{Manda2020}}}
\end{tabular} \\ 

\cmidrule[.05pt]{2-6} 
& RcNN & 
& \colorbox{purple_2}{\color{white}\cite{Jiang2020}} & \cite{Huang20191344} &  \\ 

\cmidrule[.05pt]{2-6} 
&
Transformer 
&  
\colorbox{purple_4}{\color{white} \cite{Chen_Gao_Moss_2021, Kogkalidis_Moortgat_Moot_2020} }
&  
\begin{tabular}[c]{@{}l@{}} 
\cite{Wu_Wang_Pan_2020} 
\colorbox{purple_1}{\cite{Shi_Ding_Du_Liu_Qin_2021}} \\  \colorbox{purple_1}{\cite{Wang_Pan_2021, Li_Srikumar_2019}} \\
\colorbox{purple_2}{\color{white}\cite{Pacheco_Goldwasser_2021, Honda2019152368}}
\end{tabular}
& 
\begin{tabular}[c]{@{}l@{}}  \cite{Yabloko_2020}, 
\colorbox{purple_2}{\color{white}\cite{Zhou_Richardson_2021}} \\
\colorbox{purple_2}{\color{white}\cite{Das_Zaheer_Thai_2021}} \colorbox{purple_4}{\color{white}{\cite{Jiang_Gurajada_Lu_2021}}}
\end{tabular}

&  \begin{tabular}[c]{@{}l@{}}   
\cite{Kouris_Alexandridis_Stafylopatis_2021} \colorbox{purple_1}{\cite{Dehua_Keting_Jianrong_2021}} \\ \colorbox{purple_1}{{\cite{Chen_Xu_Cheng_2020, Verga_Sun_Baldini_2021}}} \\ \colorbox{purple_1}{{\cite{Zhou20212015}}} \end{tabular} \\

\cmidrule[0.75pt]{1-6}
 
   {\multirow{2}{*}{{\begin{tabular}[c]{@{}l@{}}Neuro-\\ Symbolic\end{tabular}}}}
  &
  {LTN} &
  \colorbox{purple_3}{\color{white}\cite{Bianchi2019161}}&
  &
  &
  \\ 
  \cmidrule[.05pt]{2-6} 
  &
  {RNKN} &
  &
  \colorbox{purple_2}{\color{white}\cite{Jiang2020}} &
  &
  \\ 
  \cmidrule[.05pt]{2-6} 
  & 
  {LNN} &
  &
  \colorbox{black}{\color{white}{\cite{Chaudhury_Sen_Ono_2021}}} & 
  \colorbox{purple_4}{\color{white}{\cite{Jiang_Gurajada_Lu_2021}}}
  
  &
  \\
  
  \cmidrule[.05pt]{2-6} 
  & 
  {TPR} 
  &  \colorbox{purple_3}{\color{white}\cite{Chen20201544}}
  & &   \colorbox{purple_3}{\color{white}{\cite{Huang20191344}}} &
  \\
  \cmidrule[0.75pt]{1-6}
  
  {\multirow{2}{*}{{\begin{tabular}[c]{@{}l@{}}Neuroevolution\end{tabular}}}}
  &  &
  &
  &
  & \colorbox{purple_4}{\color{white}\cite{Skrlj2021989}} 
  \\ 
  & & & & & \\
  
  \bottomrule
  \end{tabular}

\end{table*}

\subsubsection{Neuro-Symbolic} \label{sec:nesycategories}
NeSy systems can be categorized according to the nature of the combination of neural and symbolic techniques. At AAAI-20, Henry Kautz presented a taxonomy of 6 types of Neuro-Symbolic architectures with a brief example of each \cite{Kautz}. While Kautz has not provided any additional information beyond his talk at AAAI-20, several researchers have formed their own interpretations \cite{Sarker_Zhou_Eberhart_Hitzler_2021,Garcez_Lamb_2020,Lamb_Garcez_Gori_Prates_Avelar_Vardi_2020}. We have categorized all the reviewed studies according to Kautz's taxonomy as well as our proposed nomenclature - Figure \ref{fig:NeSy_categories}. Table \ref{table:nexystudies} in Appendix \ref{sec:nesy_kautz:appendix} lists all the studies by category.
\begin{figure}[htbp]
    \centering
    \includegraphics[scale=0.85]{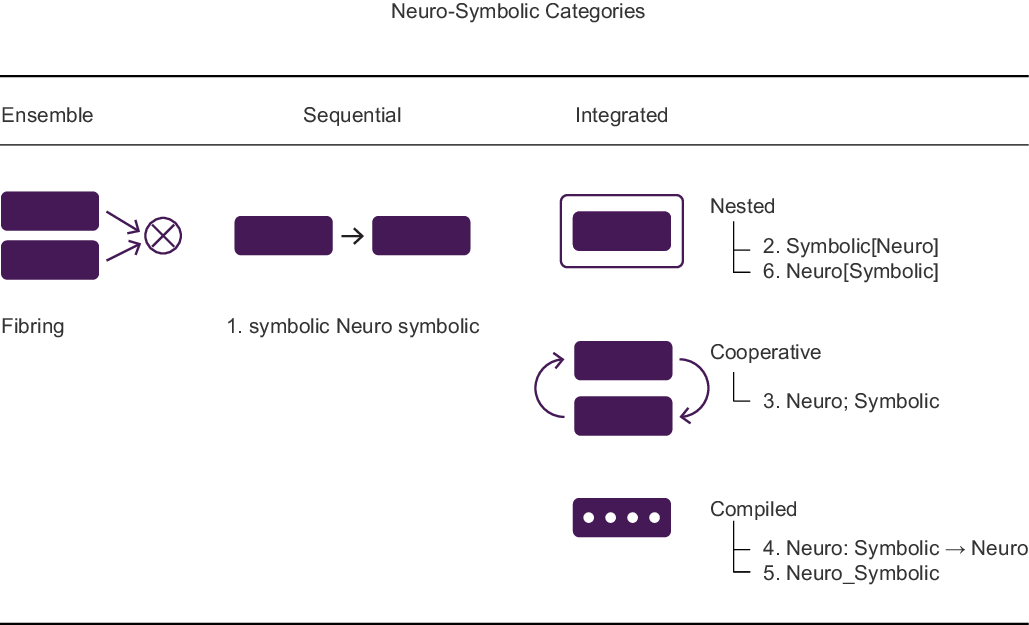}
    \caption{Proposed Neuro-Symbolic Artificial Intelligence categories. Adapted from Henry Kautz.}
    \label{fig:NeSy_categories}
\end{figure}

\begin{figure}[htp]
    \centering
    \includegraphics[width=1\textwidth]{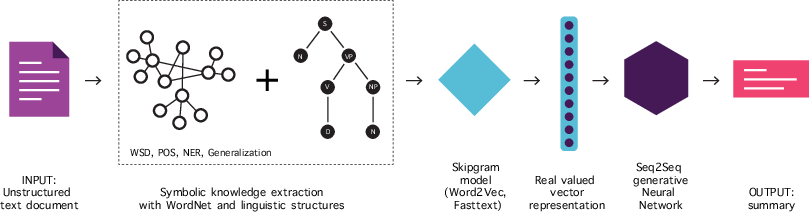}
    \caption{Type 1 \textit{Sequential}. A symbolic knowledge representation module is used to generate rich embeddings for downstream machine learning \cite{Kouris_Alexandridis_Stafylopatis_2021}.}
    \label{fig:kouris}
\end{figure}

Type 1 \textit{symbolic Neuro symbolic} is a special case where symbolic knowledge (such as words) is transformed into continuous vector space and thus encoded in the feature embeddings of an otherwise \say{standard} ML model. We opted to include these studies if the derived input features belong to the set of symbolic knowledge representations described in Section \ref{sec:glossary} - Figure \ref{fig:kouris}. 
One could still argue that this is simply a case of good old fashioned feature engineering, and not particularly special, but we want to explore the idea that deep learning can perform reasoning, albeit implicitly, if provided with a rich knowledge representation in the pre-processing phase. We classify these studies as \textit{Sequential}. Evaluating these studies as a group was particularly challenging as they have very little in common including different datasets, benchmarks and business applications. Half of the studies do not mention reasoning at all, and the ones that do are mainly executing rules on candidate solutions output by the neural models post hoc. In aggregate, only 26 out of a total of 115 (23 studies * 5 goals), or 22.6\%, possible NeSy goals were met.

\begin{figure}[htp]
    \centering
    \includegraphics[width=1\textwidth]{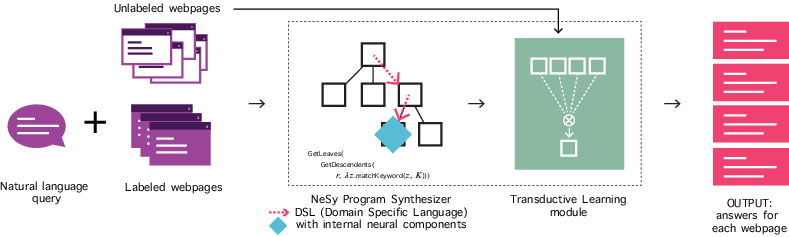}
    \caption{Type 2 \textit{Nested}. Given a natural language query and a set of web pages, the system outputs answers for each page. A symbolic reasoner, which uses a custom Domain Specific Language (DSL) to traverse the HTML, interacts with internal neural modules such as BERT which perform a number of Natural Language Processing tasks. What is learned is a DSL program, using only a few labeled examples, which can generalize to a large number of heterogeneous web pages. The authors report large improvements in precision and recall scores over state-of-the art, in some cases over 50 points \cite{Chen2021328}.}
    \label{fig:chen}
\end{figure}

Type 2 \textit{Symbolic[Neuro]} is what we describe as a \textit{Nested} architecture, where a symbolic reasoning system is the primary system with neural components driving certain internal decisions. AlphaGo is the example given by Kautz, where the symbolic system is a Monte Carlo Tree Search with neural state estimators nominating next states. We found four studies that fit this architecture. We use \cite{Chen2021328} for the purposes of illustration - Figure \ref{fig:chen}. 

\begin{figure}[htp]
    \centering
    \includegraphics[width=1\textwidth]{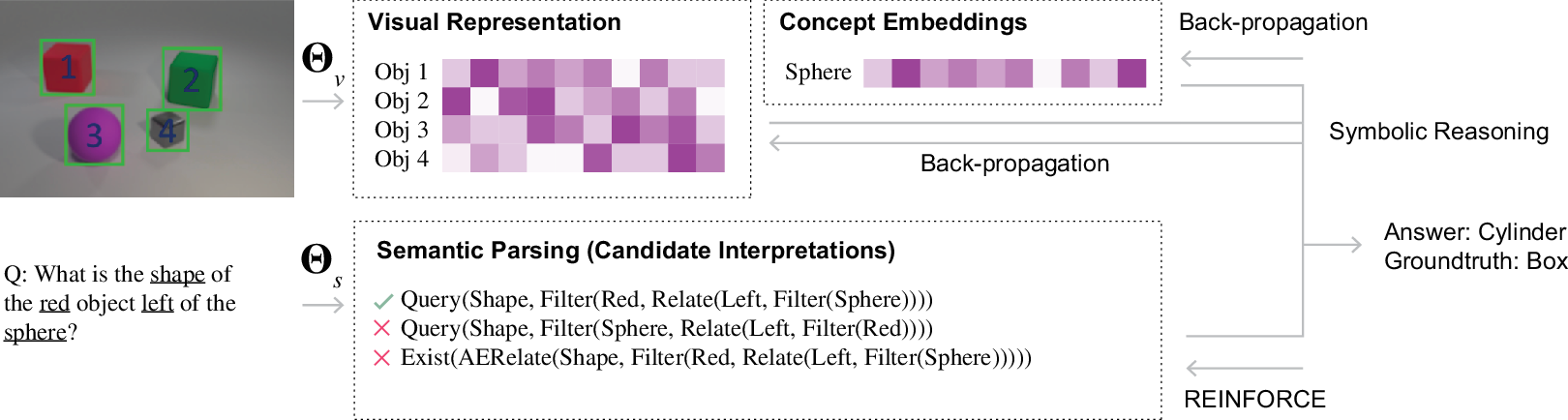}
    \caption{Type 3 \textit{Cooperative}. The Neuro-Symbolic Concept Learner (NS-CL) jointly learns visual concepts, words, and semantic parsing of sentences without any explicit annotations. Given an input image, the visual perception module detects objects in the scene and extracts a deep, latent representation for each of them. The semantic parsing module translates an input question in natural language into an executable program given a domain specific language (DSL). The generated programs have a hierarchical structure of symbolic, functional modules, each fulfilling a specific operation over the scene representation. The explicit program semantics enjoys compositionality, interpretability, and generalizability \cite{Mao2019}.}
    \label{fig:mao}
\end{figure}

Type 3 \textit{Neuro; Symbolic} is what we call \textit{Cooperative}. Here, a neural network focuses on one task (e.g. object detection) and interacts via input/output with a symbolic reasoner specializing in a complementary task (e.g. query answering). Unstructured input is converted into symbolic representations which can be solved by a symbolic reasoner, which in turn informs the neural component which learns from the errors of the symbolic component. This process is iterated until convergence or a satisfactory output is produced. There are nine studies in this category, all but one of which utilize rules and/or logic for knowledge representation. A common theme among the cooperative architectures is the business application of question answering. The Neuro-Symbolic Concept Learner (NS-CL) \cite{Mao2019} - Figure \ref{fig:mao} - is an example of Type 3, meeting 4 out of the 5 NeSy goals. Its ability to perform well with reduced data is particularly impressive: \say{Using only 10\% of the training images, our model is able to achieve comparable results with the baselines trained on the full dataset.}  Similarly, \cite{Yao201842} report perfect performance on small datasets which they also attribute to the use of explicit and precise reasoning. Both studies display similar limitations, the use of synthetic datasets, and the need for handcrafted logic, a DSL (Domain Specific Language) in the case of \cite{Mao2019}, and Image Schemas in \cite{Yao201842}. Six out of the nine studies leverage linguistic structures in some fashion, and in particular, \cite{Shi_Ding_Du_Liu_Qin_2021} utilize \textit{natural logic}, for a model which is both interpretable, and achieves state-of-the-art performance on two QA datasets. This work builds on \cite{Angeli_Manning_2014,maccartney_manning_2009_extended}.


Types 4 and 5, \textit{Neuro: Symbolic → Neuro} and \textit{Neuro\_Symbolic} respectively, were originally presented by Kautz under one heading. After his presentation, Kautz modified the slide deck\footnote{\label{kautz}\url{https://henrykautz.com/talks/index.html}} separating these two types into systems where knowledge is compiled into the network weights, and where knowledge is compiled into the loss function. In Types 4 and 5, reasoning can be performed both implicitly and explicitly, in that it is calculated via gradient descent, but can also be performed post hoc. We have grouped studies belonging to these two categories under the moniker of \textit{Compiled} systems, of which there are sixteen and seven respectively. 

\begin{figure}[htbp]
    \centering
    \includegraphics[scale=0.5]{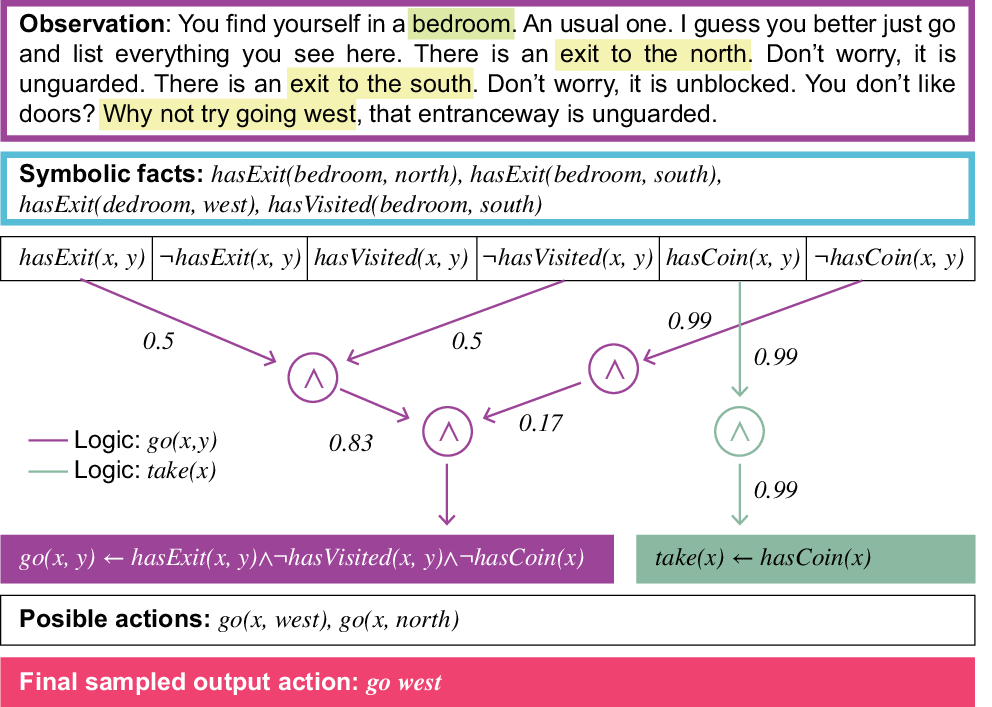}
    \caption{Type 4 \textit{Compiled}. \textbf{S}ymbo\textbf{L}ic \textbf{A}ction policy for \textbf{T}extual \textbf{E}nvironments (\textbf{SLATE}) learns interpretable action policy for each action verb, \textit{go} and \textit{take}, from first-order symbolic states. The goal is to learn symbolic rules as logical connectives for generating action commands by gradient-based training \cite{Chaudhury_Sen_Ono_2021}.}
    \label{fig:slate}
\end{figure}

\begin{figure}[htbp]
    \centering
    \includegraphics[scale=0.8]{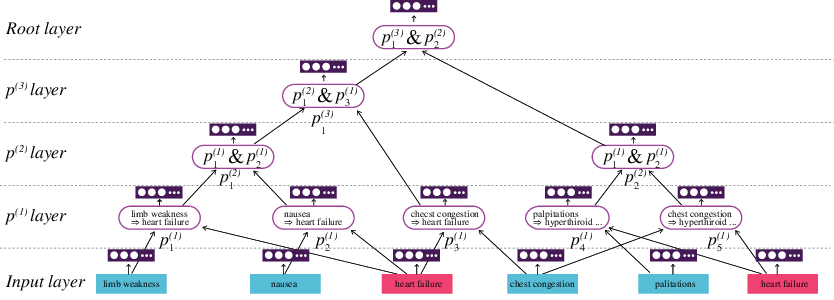}
    \caption{Type 4 \textit{Compiled}. Huffman tree of the Recursive Neural Knowledge Network (RNKN), representing deep first-order logic knowledge. The first layer of the tree consists of entities, the second layer consists of relations \begin{math}(x \rightarrow y)\end{math}. Higher layers compute logic rules. The root node is the final embedding representing a document (in this case a single health record). Back propagation is used for optimization with softmax for calculating class probabilities \cite{Jiang2020}. }
    \label{fig:Jiang2020}
\end{figure}

Deep Learning For Mathematics \cite{lample2019deep} is the canonical example of Type 4, where the input and output to the model are mathematical expressions. The model performs symbolic differentiation or integration, for example, given $x^2$ as input, the model outputs $2x$. The model exploits the tree structure of mathematical expressions, which are fed into a sequence-to-sequence architecture. This seems like a particularly fitting paradigm for natural language applications on the basis that structures such as parse trees can be similarly leveraged to output other meaningful structures such as for example: cause and effect relationships as exemplified in \cite{Zhou_Richardson_2021} and \cite{Yabloko_2020}, or the generation of argument schemes as per \cite{Saveleva_Petukhova_Mosbach_Klakow_2021}. The downside of many of these types of systems is the need for hand-crafted rules and logic \cite{Jiang_Gurajada_Lu_2021,Gupta_Ghosal_Ekbal_2021,Yabloko_2020,Demeter20207634}. In contrast, \cite{Chaudhury_Sen_Ono_2021} learn rules from data (rule induction) by combining Logical Neural Networks (LNN) with text-based Reinforcement Learning (RL). One could argue that this is a combination of Type 4, \textit{compiled} (logic embedded in the network), and Type 3, \textit{cooperative} (symbolic and sub-symbolic modules learning from each other in an iterative fashion). \cite{Chaudhury_Sen_Ono_2021} is the only work we found which meets all five promises, and, it outperforms previous SOTA approaches - Figure \ref{fig:slate}.
Another example of a Type 4 system in our set of studies is proposed by \cite{Jiang2020}. Here, knowledge is encoded in the form of huffman trees made of triples and logic expressions, in order to jointly learn embeddings and model weights - Figure \ref{fig:Jiang2020}. The model is intended for medical diagnosis decision support, where a requisite characteristic is interpretability, and this model meets that goal. 

\begin{figure}[htbp]
    \centering
    \includegraphics[scale=0.6]{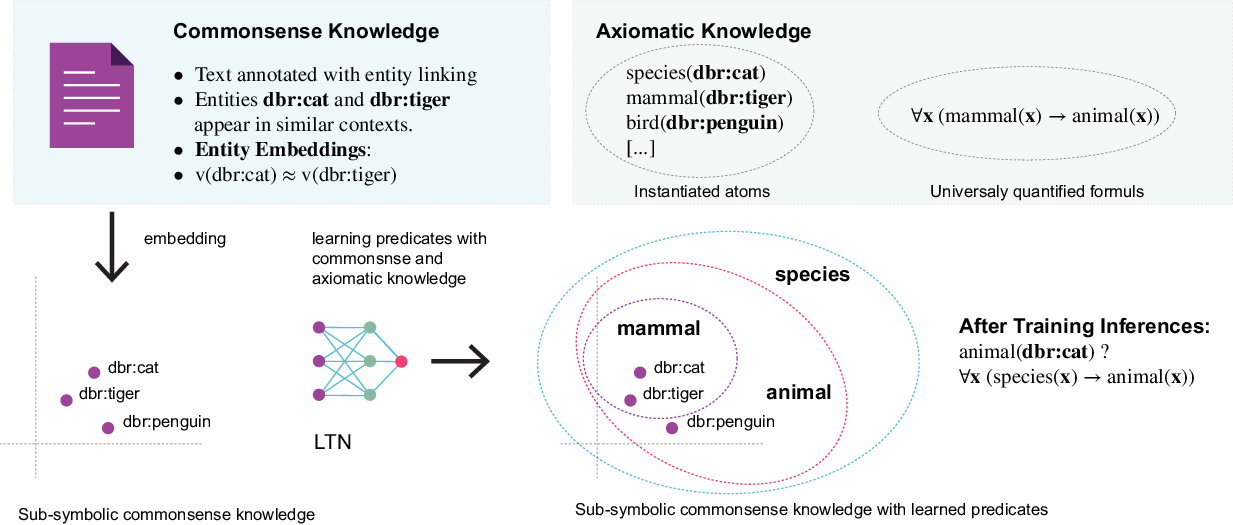}
    \caption{Type 5 \textit{Compiled}. $LTN_{EE}$ - Using Logic Tensor Networks (LTNs) it is possible to integrate axioms and facts (using first-order fuzzy logic to represent terms, functions, and predicates in a vector space) with commonsense knowledge represented in a sub-symbolic form (based on the principle of distributional semantics and implemented with Word2Vec) in one single model performing well in reasoning tasks. The major contribution of this work is to show that combining commonsense knowledge under the form of text-based entity embeddings with LTNs is not only simple, but it is also promising. LTNs can also be used to do after-training reasoning over combinations of axioms on which it was not trained \cite{Bianchi2019161}.}
    \label{fig:Bianchi2019161}
\end{figure}

Type 5 comprises Tensor Product Representations (TPRs) \cite{Smolensky_1990}, Logic Tensor Networks (LTNs) \cite{Serafini_dAvila_Garcez_2016}, Neural Tensor Networks (NTN) \cite{Socher_Chen_Manning_Ng_2013} and more broadly is referred to as tensorization, where logic acts as a constraint.  $LTN_{EE}$ \cite{Bianchi2019161} is an example of a \textit{compiled} Type 5 system - Figure \ref{fig:Bianchi2019161}.

Type 6  \textit{Neuro[Symbolic]} is the most tightly integrated but perhaps the most elusive as there do not appear to be any recent implementations in existence. According to Kautz, this is the ultimate NeSy system which should be capable of efficient combinatorial reasoning at the level of super-intelligence, if not human intelligence.

\begin{figure}[htbp]
    \centering
    \subfloat[\centering NeSy category]{{\includegraphics[scale=0.4]{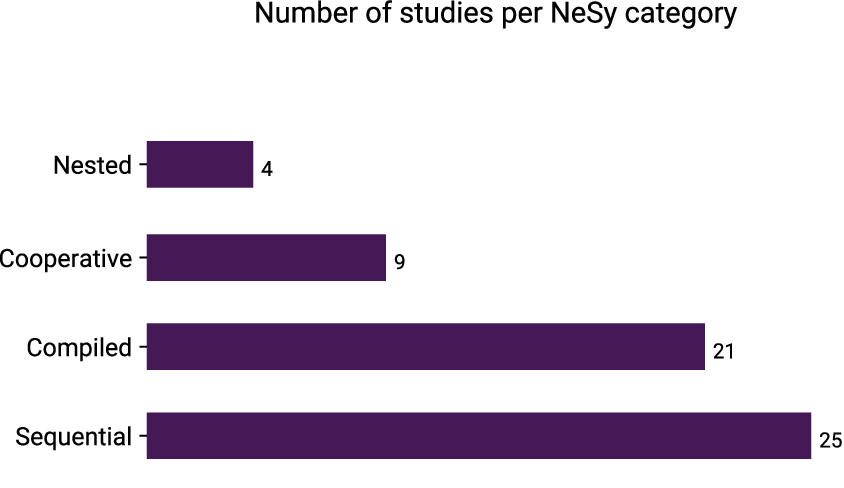} }}%
    \qquad
    \subfloat[\centering Kautz category]{{\includegraphics[scale=0.4]{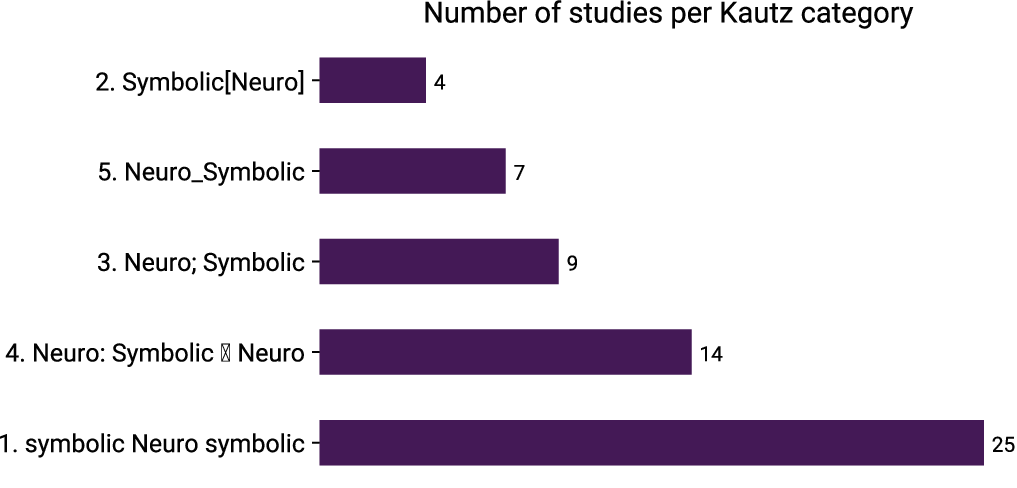} }}%
    \caption{Number of studies per category}%
    \label{fig:studies-per-category}%
\end{figure}

\begin{figure}[htbp]
    \centering
    \includegraphics[scale=0.45]{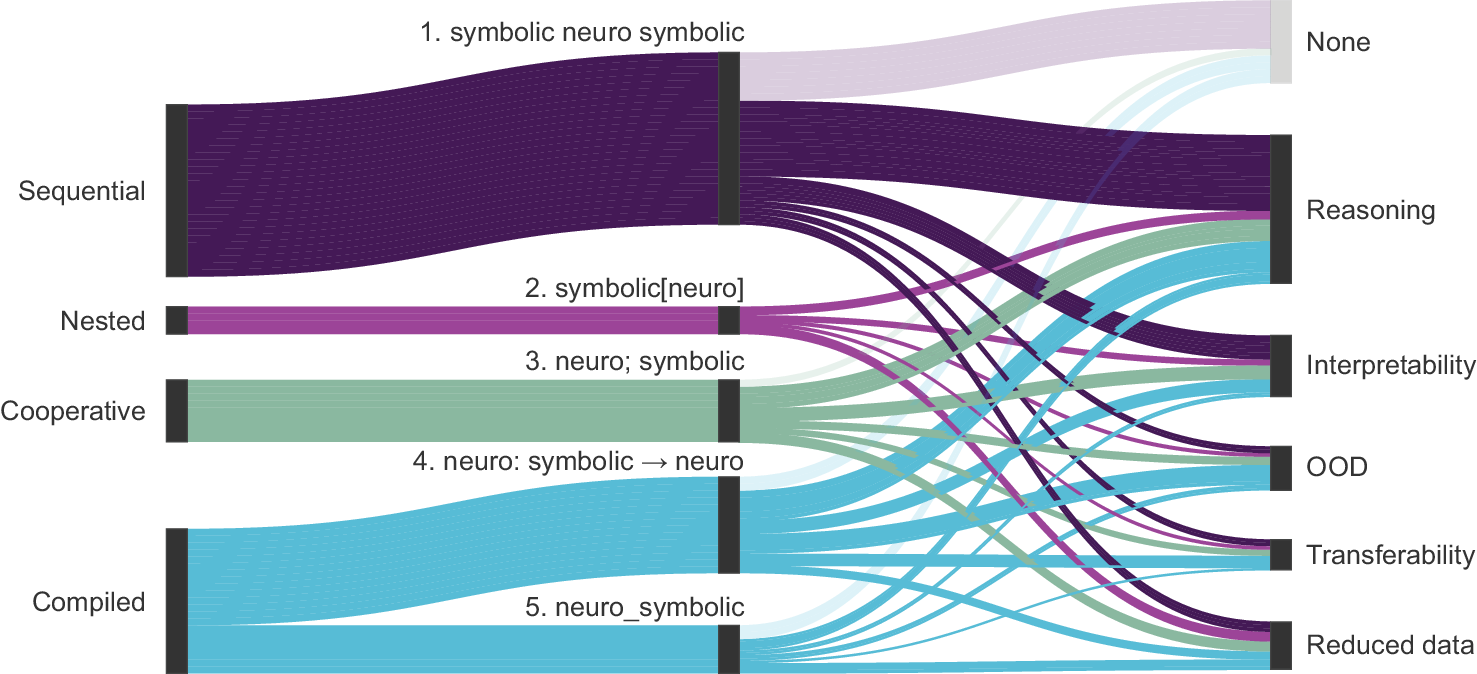}
    \caption{NeSy categories to NeSy Goals. There is no obvious pattern with respect to what types of goals are met within each of the NeSy categories. }
    \label{fig:cat-to-goals}
\end{figure}

Figure \ref{fig:studies-per-category} shows the number of studies per category, and Figure \ref{fig:cat-to-goals} illustrates the relationship between categories and goals. Table \ref{table:counts} shows the number of studies in each category per goal.

\begin{table*}[htbp]
\caption{Number of studies meeting each goal. The \textit{Promise Ratio} represents the percentage of goals reported to have been met out of the total number of possible goals (\# of studies * 5 goals) in each category.} \label{table:counts}
\centering
\footnotesize
\begin{tabular}{lrrrr}
\toprule
{} & {Compiled} & {Cooperative} & {Nested} & {Sequential} \\ 
\hline
Reasoning & {\cellcolor[HTML]{8f6795}} \color[HTML]{F1F1F1} 12 & {\cellcolor[HTML]{c0acbf}} \color[HTML]{000000} 5 & {\cellcolor[HTML]{d7cdd3}} \color[HTML]{000000} 3 & {\cellcolor[HTML]{73407d}} \color[HTML]{F1F1F1} 14 \\
OOD & {\cellcolor[HTML]{8f6795}} \color[HTML]{F1F1F1} 9 & {\cellcolor[HTML]{d7cdd3}} \color[HTML]{000000} 3 & {\cellcolor[HTML]{ffffff}} \color[HTML]{000000} 1 & {\cellcolor[HTML]{edece6}} \color[HTML]{000000} 2 \\
Interpretability & {\cellcolor[HTML]{8f6795}} \color[HTML]{F1F1F1} 8 & {\cellcolor[HTML]{c0acbf}} \color[HTML]{000000} 4 & {\cellcolor[HTML]{edece6}} \color[HTML]{000000} 2 & {\cellcolor[HTML]{c0acbf}} \color[HTML]{000000} 6 \\
Reduced data & {\cellcolor[HTML]{af94b1}} \color[HTML]{F1F1F1} 6 & {\cellcolor[HTML]{c0acbf}} \color[HTML]{000000} 4 & {\cellcolor[HTML]{edece6}} \color[HTML]{000000} 2 & {\cellcolor[HTML]{d7cdd3}} \color[HTML]{000000} 3 \\
Transferability & {\cellcolor[HTML]{af94b1}} \color[HTML]{F1F1F1} 7 & {\cellcolor[HTML]{edece6}} \color[HTML]{000000} 2 & {\cellcolor[HTML]{ffffff}} \color[HTML]{000000} 1 & {\cellcolor[HTML]{edece6}} \color[HTML]{000000} 2 \\
\hline
{Promise Ratio} & {29.5\%} & {40\%} & {45\%} & {21.6\%} \\ 
\bottomrule

\end{tabular}
\end{table*}

\section{Discussion}\label{discussion}
\begin{figure}[h]
    \centering
    \subfloat[\centering All studies]{{\includegraphics[scale=0.4]{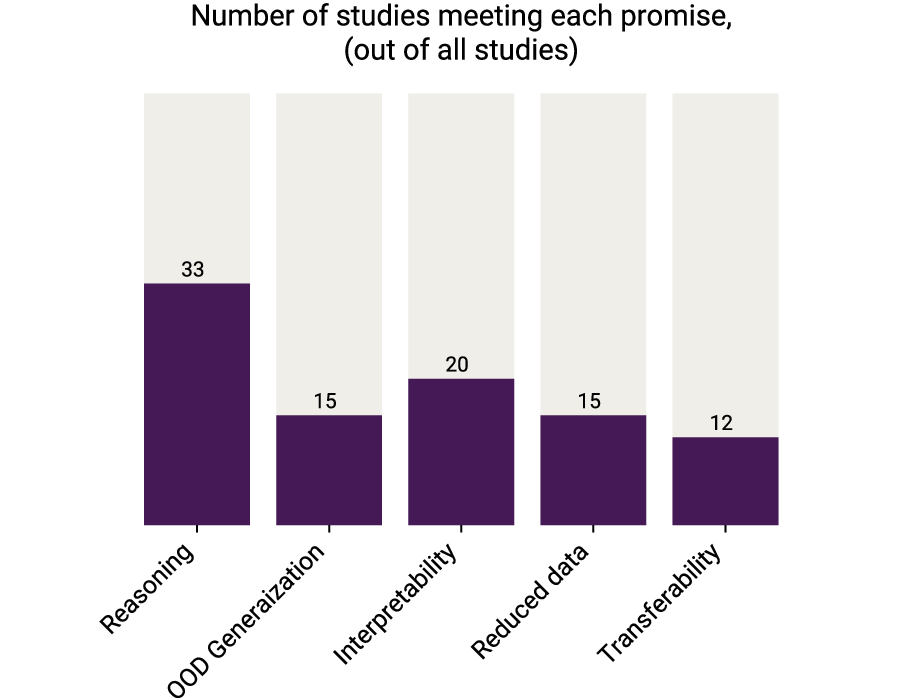} }}
    \qquad
    \subfloat[\centering NeSy studies only]{{\includegraphics[scale=0.4]{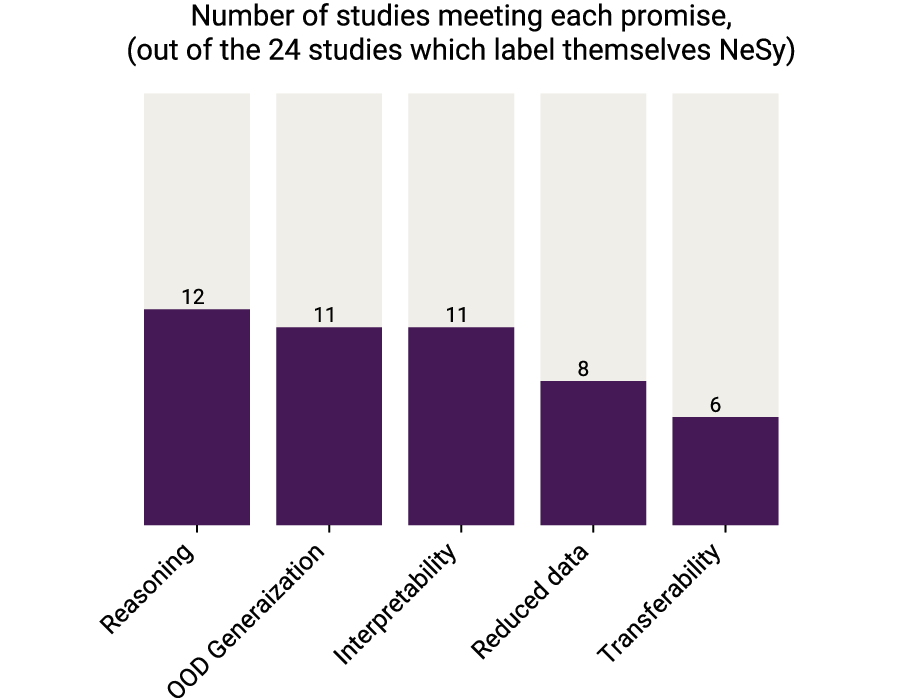} }}
    \caption{Proportion of studies which have met one or more of the 5 goals}
    \label{fig:all_promises}
\end{figure}

All studies report performance either on par or above benchmarks, but we cannot compare studies based on performance as nearly every study uses a different dataset and benchmark as discussed in Section \ref{sec:datasets}. Our focus is instead on whether the goals of NeSy are being met. Our \textit{Promise Score} metric is not necessarily what the studies' authors were optimizing for or even reporting, especially studies which have not labeled themselves as NeSy per se. So we want to make it very clear that our analysis is not a judgement of the success of any particular study, but rather we seek to understand if the hypotheses about NeSy are materializing, namely that the combination of symbolic and sub-symbolic techniques will fulfill the goals described in Section \ref{sec:contributions}: Out-of-distribution (OOD) Generalization, interpretability, tranferability, reduced data, and reasoning. And the short answer is we are not there yet, as can be seen in Figure \ref{fig:all_promises}. For a detailed breakdown of each goal and study see Table \ref{table:promises1}.

\begin{table*}
\caption{NeSy Promises reported as having been met (\colorbox{purple_4}{\color{white} y} = yes, \colorbox{tan_grey}{\color{black} n} = no)} 
\label{table:promises1}
\centering
\begin{tabular}{rccccccc}
\toprule
 \multirow{2}{*}{Ref.} &
 \multirow{2}{*}{Score} &
 \multirow{2}{*}{Reasoning} &
 \multicolumn{1}{p{2cm}}{\centering OOD\\Generalization} & \multirow{2}{*}{Interpretability} & 
 \multicolumn{1}{p{1.5cm}}{\centering Reduced\\Data } &
 \multirow{2}{*}{Transferability} & 
 \multirow{2}{*}{isNeSy} \\
\hline
\cite{Chaudhury_Sen_Ono_2021} & {\cellcolor[HTML]{000000}} \color[HTML]{F1F1F1} 5 & {\cellcolor[HTML]{450256}} \color[HTML]{F1F1F1} y & {\cellcolor[HTML]{450256}} \color[HTML]{F1F1F1} y & {\cellcolor[HTML]{450256}} \color[HTML]{F1F1F1} y & {\cellcolor[HTML]{450256}} \color[HTML]{F1F1F1} y & {\cellcolor[HTML]{450256}} \color[HTML]{F1F1F1} y & {\cellcolor[HTML]{450256}} \color[HTML]{F1F1F1} y \\
\cite{Chen_Gao_Moss_2021,Mao2019} & {\cellcolor[HTML]{450256}} \color[HTML]{F1F1F1} 4 & {\cellcolor[HTML]{450256}} \color[HTML]{F1F1F1} y & {\cellcolor[HTML]{450256}} \color[HTML]{F1F1F1} y & {\cellcolor[HTML]{450256}} \color[HTML]{F1F1F1} y & {\cellcolor[HTML]{450256}} \color[HTML]{F1F1F1} y & {\cellcolor[HTML]{EEEDE7}} \color[HTML]{000000} n & {\cellcolor[HTML]{450256}} \color[HTML]{F1F1F1} y \\
\cite{Kogkalidis_Moortgat_Moot_2020} & {\cellcolor[HTML]{450256}} \color[HTML]{F1F1F1} 4 & {\cellcolor[HTML]{450256}} \color[HTML]{F1F1F1} y & {\cellcolor[HTML]{450256}} \color[HTML]{F1F1F1} y & {\cellcolor[HTML]{450256}} \color[HTML]{F1F1F1} y & {\cellcolor[HTML]{EEEDE7}} \color[HTML]{000000} n & {\cellcolor[HTML]{450256}} \color[HTML]{F1F1F1} y & {\cellcolor[HTML]{450256}} \color[HTML]{F1F1F1} y \\
\cite{Zhang_Wang_Yu_Wang_Wang_Jiang_Lim_2021} & {\cellcolor[HTML]{450256}} \color[HTML]{F1F1F1} 4 & {\cellcolor[HTML]{450256}} \color[HTML]{F1F1F1} y & {\cellcolor[HTML]{EEEDE7}} \color[HTML]{000000} n & {\cellcolor[HTML]{450256}} \color[HTML]{F1F1F1} y & {\cellcolor[HTML]{450256}} \color[HTML]{F1F1F1} y & {\cellcolor[HTML]{450256}} \color[HTML]{F1F1F1} y & {\cellcolor[HTML]{EEEDE7}} \color[HTML]{000000} n \\
\cite{Jiang_Gurajada_Lu_2021,Skrlj2021989} & {\cellcolor[HTML]{450256}} \color[HTML]{F1F1F1} 4 & {\cellcolor[HTML]{EEEDE7}} \color[HTML]{000000} n & {\cellcolor[HTML]{450256}} \color[HTML]{F1F1F1} y & {\cellcolor[HTML]{450256}} \color[HTML]{F1F1F1} y & {\cellcolor[HTML]{450256}} \color[HTML]{F1F1F1} y & {\cellcolor[HTML]{450256}} \color[HTML]{F1F1F1} y & {\cellcolor[HTML]{450256}} \color[HTML]{F1F1F1} y \\
\cite{Sen_Danilevsky_Li_2020,Bianchi2019161} & {\cellcolor[HTML]{805387}} \color[HTML]{F1F1F1} 3 & {\cellcolor[HTML]{450256}} \color[HTML]{F1F1F1} y & {\cellcolor[HTML]{450256}} \color[HTML]{F1F1F1} y & {\cellcolor[HTML]{450256}} \color[HTML]{F1F1F1} y & {\cellcolor[HTML]{EEEDE7}} \color[HTML]{000000} n & {\cellcolor[HTML]{EEEDE7}} \color[HTML]{000000} n & {\cellcolor[HTML]{450256}} \color[HTML]{F1F1F1} y \\
\cite{Yao201842} & {\cellcolor[HTML]{805387}} \color[HTML]{F1F1F1} 3 & {\cellcolor[HTML]{450256}} \color[HTML]{F1F1F1} y & {\cellcolor[HTML]{EEEDE7}} \color[HTML]{000000} n & {\cellcolor[HTML]{450256}} \color[HTML]{F1F1F1} y & {\cellcolor[HTML]{450256}} \color[HTML]{F1F1F1} y & {\cellcolor[HTML]{EEEDE7}} \color[HTML]{000000} n & {\cellcolor[HTML]{EEEDE7}} \color[HTML]{000000} n \\
\cite{Chen20201544} & {\cellcolor[HTML]{805387}} \color[HTML]{F1F1F1} 3 & {\cellcolor[HTML]{450256}} \color[HTML]{F1F1F1} y & {\cellcolor[HTML]{EEEDE7}} \color[HTML]{000000} n & {\cellcolor[HTML]{450256}} \color[HTML]{F1F1F1} y & {\cellcolor[HTML]{EEEDE7}} \color[HTML]{000000} n & {\cellcolor[HTML]{450256}} \color[HTML]{F1F1F1} y & {\cellcolor[HTML]{EEEDE7}} \color[HTML]{000000} n \\
\cite{Huo2019159} & {\cellcolor[HTML]{805387}} \color[HTML]{F1F1F1} 3 & {\cellcolor[HTML]{EEEDE7}} \color[HTML]{000000} n & {\cellcolor[HTML]{450256}} \color[HTML]{F1F1F1} y & {\cellcolor[HTML]{EEEDE7}} \color[HTML]{000000} n & {\cellcolor[HTML]{450256}} \color[HTML]{F1F1F1} y & {\cellcolor[HTML]{450256}} \color[HTML]{F1F1F1} y & {\cellcolor[HTML]{EEEDE7}} \color[HTML]{000000} n \\
\cite{Demeter20207634} & {\cellcolor[HTML]{A586A8}} \color[HTML]{F1F1F1} 2 & {\cellcolor[HTML]{450256}} \color[HTML]{F1F1F1} y & {\cellcolor[HTML]{450256}} \color[HTML]{F1F1F1} y & {\cellcolor[HTML]{EEEDE7}} \color[HTML]{000000} n & {\cellcolor[HTML]{EEEDE7}} \color[HTML]{000000} n & {\cellcolor[HTML]{EEEDE7}} \color[HTML]{000000} n & {\cellcolor[HTML]{450256}} \color[HTML]{F1F1F1} y \\
\cite{Pacheco_Goldwasser_2021,Sutherland2019} & {\cellcolor[HTML]{A586A8}} \color[HTML]{F1F1F1} 2 & {\cellcolor[HTML]{450256}} \color[HTML]{F1F1F1} y & {\cellcolor[HTML]{EEEDE7}} \color[HTML]{000000} n & {\cellcolor[HTML]{450256}} \color[HTML]{F1F1F1} y & {\cellcolor[HTML]{EEEDE7}} \color[HTML]{000000} n & {\cellcolor[HTML]{EEEDE7}} \color[HTML]{000000} n & {\cellcolor[HTML]{450256}} \color[HTML]{F1F1F1} y \\
\cite{Lima_Espinasse_Freitas_2019,gu2019local,Schon2019293,Jiang2020} & {\cellcolor[HTML]{A586A8}} \color[HTML]{F1F1F1} 2 & {\cellcolor[HTML]{450256}} \color[HTML]{F1F1F1} y & {\cellcolor[HTML]{EEEDE7}} \color[HTML]{000000} n & {\cellcolor[HTML]{450256}} \color[HTML]{F1F1F1} y & {\cellcolor[HTML]{EEEDE7}} \color[HTML]{000000} n & {\cellcolor[HTML]{EEEDE7}} \color[HTML]{000000} n & {\cellcolor[HTML]{EEEDE7}} \color[HTML]{000000} n \\
\cite{Amin2019133,Manda2020} & {\cellcolor[HTML]{A586A8}} \color[HTML]{F1F1F1} 2 & {\cellcolor[HTML]{450256}} \color[HTML]{F1F1F1} y & {\cellcolor[HTML]{EEEDE7}} \color[HTML]{000000} n & {\cellcolor[HTML]{EEEDE7}} \color[HTML]{000000} n & {\cellcolor[HTML]{450256}} \color[HTML]{F1F1F1} y & {\cellcolor[HTML]{EEEDE7}} \color[HTML]{000000} n & {\cellcolor[HTML]{EEEDE7}} \color[HTML]{000000} n \\
\cite{Zhou_Richardson_2021,Qin_Liang_Hong_Tang_Lin_2021} & {\cellcolor[HTML]{A586A8}} \color[HTML]{F1F1F1} 2 & {\cellcolor[HTML]{450256}} \color[HTML]{F1F1F1} y & {\cellcolor[HTML]{EEEDE7}} \color[HTML]{000000} n & {\cellcolor[HTML]{EEEDE7}} \color[HTML]{000000} n & {\cellcolor[HTML]{EEEDE7}} \color[HTML]{000000} n & {\cellcolor[HTML]{450256}} \color[HTML]{F1F1F1} y & {\cellcolor[HTML]{450256}} \color[HTML]{F1F1F1} y \\
\cite{Chaturvedi2019264} & {\cellcolor[HTML]{A586A8}} \color[HTML]{F1F1F1} 2 & {\cellcolor[HTML]{450256}} \color[HTML]{F1F1F1} y & {\cellcolor[HTML]{EEEDE7}} \color[HTML]{000000} n & {\cellcolor[HTML]{EEEDE7}} \color[HTML]{000000} n & {\cellcolor[HTML]{EEEDE7}} \color[HTML]{000000} n & {\cellcolor[HTML]{450256}} \color[HTML]{F1F1F1} y & {\cellcolor[HTML]{EEEDE7}} \color[HTML]{000000} n \\
\cite{AltszylerBBBV21} & {\cellcolor[HTML]{A586A8}} \color[HTML]{F1F1F1} 2 & {\cellcolor[HTML]{EEEDE7}} \color[HTML]{000000} n & {\cellcolor[HTML]{450256}} \color[HTML]{F1F1F1} y & {\cellcolor[HTML]{EEEDE7}} \color[HTML]{000000} n & {\cellcolor[HTML]{450256}} \color[HTML]{F1F1F1} y & {\cellcolor[HTML]{EEEDE7}} \color[HTML]{000000} n & {\cellcolor[HTML]{450256}} \color[HTML]{F1F1F1} y \\
\cite{Das_Zaheer_Thai_2021,Honda2019152368} & {\cellcolor[HTML]{A586A8}} \color[HTML]{F1F1F1} 2 & {\cellcolor[HTML]{EEEDE7}} \color[HTML]{000000} n & {\cellcolor[HTML]{450256}} \color[HTML]{F1F1F1} y & {\cellcolor[HTML]{EEEDE7}} \color[HTML]{000000} n & {\cellcolor[HTML]{450256}} \color[HTML]{F1F1F1} y & {\cellcolor[HTML]{EEEDE7}} \color[HTML]{000000} n & {\cellcolor[HTML]{EEEDE7}} \color[HTML]{000000} n \\
\cite{cowen2019neural} & {\cellcolor[HTML]{A586A8}} \color[HTML]{F1F1F1} 2 & {\cellcolor[HTML]{EEEDE7}} \color[HTML]{000000} n & {\cellcolor[HTML]{450256}} \color[HTML]{F1F1F1} y & {\cellcolor[HTML]{EEEDE7}} \color[HTML]{000000} n & {\cellcolor[HTML]{EEEDE7}} \color[HTML]{000000} n & {\cellcolor[HTML]{450256}} \color[HTML]{F1F1F1} y & {\cellcolor[HTML]{EEEDE7}} \color[HTML]{000000} n \\
\cite{Xu_Li_2019} & {\cellcolor[HTML]{A586A8}} \color[HTML]{F1F1F1} 2 & {\cellcolor[HTML]{EEEDE7}} \color[HTML]{000000} n & {\cellcolor[HTML]{EEEDE7}} \color[HTML]{000000} n & {\cellcolor[HTML]{450256}} \color[HTML]{F1F1F1} y & {\cellcolor[HTML]{EEEDE7}} \color[HTML]{000000} n & {\cellcolor[HTML]{450256}} \color[HTML]{F1F1F1} y & {\cellcolor[HTML]{EEEDE7}} \color[HTML]{000000} n \\
\cite{Wang_Pan_2021,Lemos2020647} & {\cellcolor[HTML]{D7CDD3}} \color[HTML]{000000} 1 & {\cellcolor[HTML]{450256}} \color[HTML]{F1F1F1} y & {\cellcolor[HTML]{EEEDE7}} \color[HTML]{000000} n & {\cellcolor[HTML]{EEEDE7}} \color[HTML]{000000} n & {\cellcolor[HTML]{EEEDE7}} \color[HTML]{000000} n & {\cellcolor[HTML]{EEEDE7}} \color[HTML]{000000} n & {\cellcolor[HTML]{450256}} \color[HTML]{F1F1F1} y \\
\multicolumn{1}{r}{
\begin{tabular}{@{}r@{}}
\cite{Saveleva_Petukhova_Mosbach_Klakow_2021,Hu_Wei_Huai_2021,Chen_Xu_Cheng_2020,Hussain20181662} \\ \cite{Liu2021260,Bounabi2021229,Es-Sabery202117943,Ayyanar2019} \\ 
\cite{Zhou20212015,Fazlic20191025,DSouza201990}
\end{tabular}} & 
{\cellcolor[HTML]{D7CDD3}} \color[HTML]{000000} 1 & {\cellcolor[HTML]{450256}} \color[HTML]{F1F1F1} y & {\cellcolor[HTML]{EEEDE7}} \color[HTML]{000000} n & {\cellcolor[HTML]{EEEDE7}} \color[HTML]{000000} n & {\cellcolor[HTML]{EEEDE7}} \color[HTML]{000000} n & {\cellcolor[HTML]{EEEDE7}} \color[HTML]{000000} n & {\cellcolor[HTML]{EEEDE7}} \color[HTML]{000000} n \\
\cite{Verga_Sun_Baldini_2021} & {\cellcolor[HTML]{D7CDD3}} \color[HTML]{000000} 1 & {\cellcolor[HTML]{EEEDE7}} \color[HTML]{000000} n & {\cellcolor[HTML]{450256}} \color[HTML]{F1F1F1} y & {\cellcolor[HTML]{EEEDE7}} \color[HTML]{000000} n & {\cellcolor[HTML]{EEEDE7}} \color[HTML]{000000} n & {\cellcolor[HTML]{EEEDE7}} \color[HTML]{000000} n & {\cellcolor[HTML]{450256}} \color[HTML]{F1F1F1} y \\
\cite{Shi_Ding_Du_Liu_Qin_2021} & {\cellcolor[HTML]{D7CDD3}} \color[HTML]{000000} 1 & {\cellcolor[HTML]{EEEDE7}} \color[HTML]{000000} n & {\cellcolor[HTML]{EEEDE7}} \color[HTML]{000000} n & {\cellcolor[HTML]{450256}} \color[HTML]{F1F1F1} y & {\cellcolor[HTML]{EEEDE7}} \color[HTML]{000000} n & {\cellcolor[HTML]{EEEDE7}} \color[HTML]{000000} n & {\cellcolor[HTML]{450256}} \color[HTML]{F1F1F1} y \\
\cite{Dehua_Keting_Jianrong_2021} & {\cellcolor[HTML]{D7CDD3}} \color[HTML]{000000} 1 & {\cellcolor[HTML]{EEEDE7}} \color[HTML]{000000} n & {\cellcolor[HTML]{EEEDE7}} \color[HTML]{000000} n & {\cellcolor[HTML]{450256}} \color[HTML]{F1F1F1} y & {\cellcolor[HTML]{EEEDE7}} \color[HTML]{000000} n & {\cellcolor[HTML]{EEEDE7}} \color[HTML]{000000} n & {\cellcolor[HTML]{EEEDE7}} \color[HTML]{000000} n \\
\cite{Li_Srikumar_2019,Chen2021328} & {\cellcolor[HTML]{D7CDD3}} \color[HTML]{000000} 1 & {\cellcolor[HTML]{EEEDE7}} \color[HTML]{000000} n & {\cellcolor[HTML]{EEEDE7}} \color[HTML]{000000} n & {\cellcolor[HTML]{EEEDE7}} \color[HTML]{000000} n & {\cellcolor[HTML]{450256}} \color[HTML]{F1F1F1} y & {\cellcolor[HTML]{EEEDE7}} \color[HTML]{000000} n & {\cellcolor[HTML]{450256}} \color[HTML]{F1F1F1} y \\
\cite{Gupta_Ghosal_Ekbal_2021,Pinhanez_Cavalin_Alves_2021,Yabloko_2020, Wu_Wang_Pan_2020} & {\cellcolor[HTML]{EEEDE7}} \color[HTML]{000000} 0 & {\cellcolor[HTML]{EEEDE7}} \color[HTML]{000000} n & {\cellcolor[HTML]{EEEDE7}} \color[HTML]{000000} n & {\cellcolor[HTML]{EEEDE7}} \color[HTML]{000000} n & {\cellcolor[HTML]{EEEDE7}} \color[HTML]{000000} n & {\cellcolor[HTML]{EEEDE7}} \color[HTML]{000000} n & {\cellcolor[HTML]{450256}} \color[HTML]{F1F1F1} y \\
\multicolumn{1}{r}{
\begin{tabular}{@{}r@{}}
\cite{Langton_Srihasam_2021,Kouris_Alexandridis_Stafylopatis_2021,Brasoveanu2019656,Graziani2019185} \\
\cite{Cui2021419,Gong202030885,Huang20191344,Tato2019623}
\end{tabular}} & 
{\cellcolor[HTML]{EEEDE7}} \color[HTML]{000000} 0 & {\cellcolor[HTML]{EEEDE7}} \color[HTML]{000000} n & {\cellcolor[HTML]{EEEDE7}} \color[HTML]{000000} n & {\cellcolor[HTML]{EEEDE7}} \color[HTML]{000000} n & {\cellcolor[HTML]{EEEDE7}} \color[HTML]{000000} n & {\cellcolor[HTML]{EEEDE7}} \color[HTML]{000000} n & {\cellcolor[HTML]{EEEDE7}} \color[HTML]{000000} n \\
\end{tabular}
\end{table*}

In Section \ref{sec:intro_reasoning} we put forward the hypothesis that reasoning is the means by which the other goals can be achieved. This is not evidenced in the studies we reviewed. Some possible explanations for this finding are: 1) The kind of reasoning required to fulfill the other goals is not the kind being implemented; 2) The approaches are theoretically promising, but the technical solutions need further development. Next we look at each of these possibilities.

\subsection{Reasoning Challenges}\label{discussion:reasoning}

Thirty four out of the fifty nine studies mention reasoning as a characteristic of their solution. But there is a lot of variation in how reasoning is described and implemented. Given the overwhelming evidence of the fallibility of human reasoning, to understand language, AI researchers have sought guidance from disciplines such as psychology, cognitive linguistics, neuroscience, and philosophy. The challenge is that there are multiple competing theories of human reasoning and logic both across and within these disciplines. What we have discovered in our review, is a blurring of the lines between various types of logic, human reasoning, and mathematical reasoning, as well as counter-productive assumptions about which theory to adopt. For example, drawing inspiration from \say{how people think}, accepting that how people think is flawed, and subsequently attempting to build a model with a logical component, which by definition, is rooted in validity, seems counter productive to us. Although this does depend somewhat on the business application. For problems like MWP (\textit{Math Word Problems}) \cite{Zhang_Wang_Yu_Wang_Wang_Jiang_Lim_2021,Qin_Liang_Hong_Tang_Lin_2021,Chen20201544}, where answers are precise and unambiguous, less assumptions are needed. Additionally, the justification of \say{because that's how people think} is inconsistent. Some examples from the studies we reviewed include: 

\begin{itemize}
    \item \cite{Bianchi2019161} describe human reasoning in terms of a dual process of \say{subsymbolic commonsense} (strongly correlated with associative learning), and \say{axiomatic} knowledge (predicates and logic formulas) for structured inference.
    \item In \cite{Hussain20181662} humans reason by way of analogy, and commonsense knowledge is represented in  ConceptNet, a graphical representation of common concepts and their relationships.
    \item For \cite{Yao201842} human reasoning can be modeled by Image Schemas (IS). Schemas are made up of logical rules on (Entity1,Relation,Entity2) tuples, such as transitivity, or inversion. 
    \item  \cite{Es-Sabery202117943} explain their choice of fuzzy logic  for \say{its resemblance to human reasoning and natural language.} This is a probabilistic approach which attempts to deal with uncertainty.
    \item  \cite{Ayyanar2019} propose that human thought constructs can be modelled as cause-effect pairs. Commonsense is often described as the ability to draw causal conclusions from basic knowledge, for example: \textit{If I drop the glass, it will break}.
    \item And \cite{Chen20201544} state that \say{when people perform explicit reasoning, they can typically describe the way to the conclusion step by step via relational descriptions.}
\end{itemize}
    
But the most plausible hypothesis in our view is that of Schon et al. \cite{Schon2019293}: in order to emulate human reasoning, systems need to be flexible, be able to deal with contradicting evidence, evolving evidence, have access to enormous amounts of background knowledge, and include a combination of different techniques and logics. Most notably, no particular theory of reasoning is given. The argument put forward by Leslie Kaelbling at IBM Neuro-Symbolic AI Workshop 2022\footnote{\url{https://researcher.watson.ibm.com/researcher/view_group.php?id=10897}} is similarly appealing. Kaelbling points to the over-reliance on the System1/System2 analogy, and advocates for a much more diverse and dynamic approach. We posit that the type of reasoning employed should not be based solely on how we think people think, but on the attendant objective. This is in line with the \say{goal oriented} theory from neuroscience, in that reasoning involves many sub-systems: perception, information retrieval, decision making, planning, controlling, and executing, utilizing working memory, calculation, and pragmatics. But here the irony is not lost on us, and we acknowledge that by resorting to neuroscience for inspiration, we have just committed the same mischief for which we have been decrying our peers! But if we must resort to analogies with human reasoning then it is imperative to be as rigorous as possible. In their recent book, \textit{A Formal Theory of Commonsense Psychology, How People Think People Think} \cite{gordon_hobbs_2017}, Gordon and Hobbs present a \say{large-scale logical formalization of commonsense psychology in support of humanlike artificial intelligence} to act as a baseline for researchers building intelligent AI systems. Santos et al.  \cite{santos_kejriwal_mulvehill_forbush_mcguinness_2021} take this a step in the direction we are advocating, by testing whether there is human annotator agreement when categorizing texts into Gordon and Hobbs' theories. \say{Our end-goal is to advocate for better design of commonsense benchmarks [and to] support the development of a formal logic for commonsense reasoning} \cite{santos_kejriwal_mulvehill_forbush_mcguinness_2021}. It is difficult to imagine a single formal logic which would afford all of Gordon and Hobbs' 48 categories of reasoning tasks. Besold et al. \cite{besold2017neural} dedicate several pages to this topic under the heading of Neural-Symbolic Integration in and for Cognitive Science: Building Mental Models. In short, computational modelling of cognitive tasks and especially language processing is still considered a hard challenge.


\subsection{Technical challenges}\label{discussion:technical}

There is strong agreement that a successful NeSy system will be characterized by compositionality \cite{Garcez_Lamb_2020,Cartuyvels_Spinks_Moens_2021,Garcez_Gori_Lamb_Serafini_Spranger_Tran_2019,besold2017neural,Tsamoura_Hospedales_Michael_2021,Boleda_2020,Chen_Liang_Yu_Song_Zhou_2020,Belle_2020}. Compositionality  allows for the construction of new meaning from learned building blocks thus enabling extrapolation beyond the training data distribution. To paraphrase Garcez et al., one should be able to query the trained network using a rich description language at an adequate level of abstraction \cite{Garcez_Lamb_2020}. The challenge is to come up with dense/compact differentialble representations while preserving the ability to decompose, or unbind, the learned representations for downstream reasoning tasks. 

One such system, proposed by Bianchi et al. \cite{Bianchi2019161} is the $LTN_{EE}$ - Figure \ref{fig:Bianchi2019161} - an extention of Logic Tensor Networks (LTNs), in which pre-trained embeddings are fed into the LTN. They show promising results on small datasets which have the important characteristic of being capable of after-training logical inferences. However, $LTN_{EE}$ is limited by heavy computational requirements as the logic becomes more expressive, for example by the use of quantifiers. 

Other studies \cite{Mao2019,Yao201842} introduce logical inference within their solutions, but all require manually designed rules, and are limited by the domain expertise of the designer.  Learning rules from data, or structure learning \cite{Embar_Sridhar_Farnadi_Getoor_2018} is an ongoing research topic as pointed out by \cite{von_Rueden_Mayer_Beckh_Georgiev_Giesselbach_Heese_Kirsch_Walczak_Pfrommer_Pick_etal_2021}. In \cite{Chaturvedi2019264} Chaturvedi et al. use fuzzy logic for emotion classification where explicit membership functions are learned. However, as stated by the authors, the classifier becomes very slow with the number of functions.

Other (\textit{compiled}) approaches involve translating logic into differentialble functions, which are either directly included as network nodes as in \cite{Jiang2020}, or added as a constraint to the loss function, as in \cite{Diligenti2017143}. To achieve this, First Order Logic (FOL) can be operationalized using t-norms for example. To address the many types of reasoning as discussed in the previous section, we need to be able to incorporate other types of logic, such as temporal, modal, epistemic, non-monotonic, probabilistic, and more, which, presumably, are better able to model human reasoning. 

In summary, formulating logic, or more broadly reasoning, in a differentiable fashion remains challenging.

\section{Limitations \& Future Work}\label{future}

We organized our analysis according to the characteristics extracted from the studies to test whether there were any patterns leading to NeSy goals. Another approach would be to reverse this perspective, and look at each goal separately to understand the characteristics leading to its fulfillment. However, each goal is really an entire field of study in and of itself, and we do not think we could have done justice to any of them by taking this approach. We spent a lot of time looking for signal in a very noisy environment where the studies we reviewed had very little in common. More can be said about what we did not find, than what we did. Another approach might be to narrow the criteria for the type of NLP task, while expanding the technical domain. In particular, a subset of tasks from the NLU domain could be a good starting point, as these tasks are often said to require reasoning. 

We tried to be comprehensive in respect to the selected studies which led to the trade-off of less space dedicated to technical details or additional context from the neuro-symbolic discussion. There are a lot of ideas and concepts which we did not cover, such as, and in no particular order, Relational Statistical Learning (RSL), Inductive Logic Programming (ILP), DeepProbLog \cite{manhaeve2018deepproblog},  Connectionist Modal Logics (CML), Extreme Learning Machines (ELM), Genetic Programming, grounding and proposinalization, Case Based Reasoning (CBR), Abstract Meaning Representation (AMR), to name but a few, some of which are covered in detail in other surveys \cite{Garcez_Gori_Lamb_Serafini_Spranger_Tran_2019,besold2017neural}.

Furthermore, we argued that we need differentiable forms of different types of logic, but we did not discuss how they might be implemented. A comprehensive point of reference such as this would be a very valuable contribution to the NeSy community, especially if the implementations were anchored in cognitive science and linguistics as discussed in \ref{discussion:reasoning}.

Finally, the need for common datasets and benchmarks cannot be overstated. 

\section{Conclusion}\label{conclusion}
We analyzed recent studies implementing NeSy for NLP in order to test whether the promises of NeSy are materializing in NLP. We attempted to find a pattern in a small and widely variable set of studies, and ultimately we do not believe there are enough results to draw definitive conclusions. Only 59 studies met the criteria for our review, and many of them (in the \textit{Sequential} category) we would not consider truly integrated NeSy systems. The one thing studies which meet the most goals \cite{Chen_Gao_Moss_2021,Zhang_Wang_Yu_Wang_Wang_Jiang_Lim_2021,Chaudhury_Sen_Ono_2021,Jiang_Gurajada_Lu_2021,Kogkalidis_Moortgat_Moot_2020,Skrlj2021989,Mao2019} have in common is that they all belong to the tightly integrated set of NeSy categories, \textit{Cooperative} and \textit{Compiled} which is good news for NeSy. Two out of these seven report lower computational cost than baselines, and performance on par or slightly above baselines, though we must reiterate that performance comparisons are not possible as discussed in Section \ref{sec:datasets}. On the down side, we have seen that some studies suffer from high computational cost, and that explicit reasoning still often requires hand crafted domain specific rules and logic which makes them difficult to scale or generalize to other applications. Indeed, of the five goals, transferability to new domains was the least frequently satisfied. 

Our view is that the lack of consensus around theories of reasoning and appropriate benchmarks is hindering our ability to evaluate progress. Hence we advocate for the development of robust reasoning theories and formal logics as well as the development of challenging benchmarks which not only measure the performance of specific implementations, but have the potential to address real world problems. Systems capable of capturing the nuances of natural language (ie., ones that \say{understand} human reasoning) while returning sound conclusions (ie., perform logical reasoning) could help combat some of the most consequential issues of our times such as mis- and dis-information, corporate propaganda such as climate change denialism, divisive political speech, and other harmful rhetoric in the social discourse.

\begin{acks}
This publication has emanated from research supported in part by a grant from Science Foundation Ireland under Grant number 18/CRT/6183. For the purpose of Open Access, the author has applied a CC BY public copyright licence to any Author Accepted Manuscript version arising from this submission. 
\end{acks}


\bibliographystyle{ios1}           
\bibliography{bibliography}        

\begin{appendix}

\section{NeSy and Kautz Categories}\label{sec:nesy_kautz:appendix}

\begin{table*}[htbp]

\caption{ 
    \centering 
    \text{NeSy and Kautz Categories} 
    } 
\label{table:nexystudies}
\begin{tabular}{lll}
\toprule
NeSy (Ours) & Kautz & Refs. \\
\cmidrule[.3pt]{1-3}
Sequential & {1. symbolic Neuro symbolic} &
\begin{tabular}[l]{@{}l@{}}
\cite{Langton_Srihasam_2021,Kouris_Alexandridis_Stafylopatis_2021,Hu_Wei_Huai_2021,Chen_Xu_Cheng_2020,Lima_Espinasse_Freitas_2019,Xu_Li_2019,cowen2019neural,gu2019local,Brasoveanu2019656,Cui2021419,Schon2019293,Bounabi2021229,Honda2019152368,Gong202030885,Es-Sabery202117943,Amin2019133,Sutherland2019} \\
\cite{Ayyanar2019,Zhou20212015,Manda2020,Fazlic20191025,DSouza201990,Tato2019623,Dehua_Keting_Jianrong_2021,Pinhanez_Cavalin_Alves_2021} \\
\end{tabular} \\
Nested &  {2. Symbolic[Neuro]} &
\cite{Pacheco_Goldwasser_2021,Chen_Gao_Moss_2021,Chaturvedi2019264,Chen2021328} \\
Cooperative &  {3. Neuro; Symbolic} &
\cite{Shi_Ding_Du_Liu_Qin_2021,Das_Zaheer_Thai_2021,Wang_Pan_2021,Qin_Liang_Hong_Tang_Lin_2021,Wu_Wang_Pan_2020,Lemos2020647,Skrlj2021989,Yao201842,Mao2019} \\
Compiled &  {4. Neuro: Symbolic → Neuro} &
\cite{Saveleva_Petukhova_Mosbach_Klakow_2021,Gupta_Ghosal_Ekbal_2021,Zhou_Richardson_2021,Verga_Sun_Baldini_2021,Zhang_Wang_Yu_Wang_Wang_Jiang_Lim_2021,Chaudhury_Sen_Ono_2021,Jiang_Gurajada_Lu_2021,Yabloko_2020,Sen_Danilevsky_Li_2020,Kogkalidis_Moortgat_Moot_2020,Jiang2020,Liu2021260,Huo2019159,Demeter20207634} \\
&  {5. Neuro\_Symbolic} &
\cite{Li_Srikumar_2019,AltszylerBBBV21,Graziani2019185,Bianchi2019161,Hussain20181662,Huang20191344,Chen20201544} \\
\bottomrule
\end{tabular}
\end{table*}

\newpage
\section{Allowed Values}\label{sec:allowedvalues:appendix} 

\begin{table*}[h]
   \centering 
    \caption{Allowed values}
    \label{table:allowedvalues:appendix}
    \renewcommand{\arraystretch}{1}
    \begin{tabular}{ll}    
    \toprule
    \textbf{Feature} & \textbf{Allowed values} \\ 
    \midrule
    Business application  &
    {\begin{tabular}[m]{@{}l@{}}  
    
    Annotation, Argumentation mining, Causal Reasoning, Decision support, Dialog system, Emotion recognition, \\ Entity Linking, Entity Resolution, Image captioning, Information extraction,  KG Completion / link prediction, \\ Language modeling, N2F, Opinion extraction, Question answering,  Reading comprehension, Relation extraction, \\ Sentiment analysis, Text classification, Text games,  Text summarization
    
 \end{tabular}}
     \\ 
     \\
    Technical application  & 
    {\begin{tabular}[]{@{}l@{}} Clustering, Generative,  Inference,  Classification, Information extraction, Similarity  \end{tabular}} \\
    \\
    Type of learning &  {\begin{tabular}[]{@{}l@{}} Supervised, Unsupervised, Semi-supervised, Reinforcement, Curriculum  \end{tabular}} \\ 
    \\
    Type of reasoning &  Implicit, Explicit, Both
    \\ 
    \\
    Language structure & Yes, No\\ 
    \\
    Relational structure & Yes, No\\ 
    \\
    NeSy goals &    {\begin{tabular}[]{@{}l@{}}
    Reasoning, OOD Generalization, Interpretability, Reduced data, Transferability 
    \end{tabular}}
    \\ \\
   Kautz category & {\begin{tabular}[]{@{}l@{}}
    1. symbolic Neuro symbolic, 
    2. Symbolic[Neuro], 3. Neuro; Symbolic, \\
    4. Neuro: Symbolic → Neuro, 
    5. Neuro\_Symbolic, 6. Neuro[Symbolic] 
 \end{tabular}}  \\ \\  
   NeSy category & Sequential, Nested, Cooperative, Compiled  \\ 
   \bottomrule 
    \end{tabular}  
\end{table*}

\newpage

\section{Venues}
\label{sec:venues:appendix}

\begin{table*}[h]
    \centering 
    \caption{Venues referred in the study}
    \label{table:venues}
    \renewcommand{\arraystretch}{1}
    \begin{tabular}{l} 
    \toprule
 American Association for the Advancement of Science \\
 American Chemical Society \\
 American Institute of Physics \\
 American Society for Microbiology \\
 Association for Computing Machinery (ACM) \\
 Association for Computational Linguistics (ACL) \\
 Cairo University \\
 Chongqing University of Posts and Telecommunications \\
 Elsevier \\
 Emerald \\
 IEEE \\
 IOS Press \\
 Institute for Operations Research and the Management Sciences \\
 King Saud University \\
 MIT Press\\
 Mary Ann Liebert \\
 Morgan \& Claypool Publishers \\
 Now Publishers Inc \\
 Optical Society of America \\
 Oxford University Press \\
 Public Library of Science \\
 SAGE \\
 Society for Industrial and Applied Mathematics \\
 Springer Nature \\
 Taylor \& Francis \\
 University of California Press \\
 University of Minnesota \\
 Wiley-Blackwell \\
   \bottomrule
    \end{tabular}
\end{table*}

\newpage

\section{Acronyms}
\label{sec:acronyms:appendix}

\begin{table*}[h]
    \centering 
    \caption{Acronyms and Abbreviations}
    \label{table:acronyms:appendix}
    \renewcommand{\arraystretch}{0.95}
    \begin{tabular}{ll} 
    \toprule
    AAAI & Association for the Advancement of Artificial Intelligence \\
    ACL & Association for Computational Linguistics \\
    AI & Artificial Intelligence \\
    AR & Analogical Reasoning \\ 
    CBR & Case based reasoning \\ 
    CNN & Convolutional Neural Network \\
    DBN	& Deep Belief Network \\
    DL	& Deep Learning \\
    DLs & Description Logic \\ 
    GAT & Graph Attention Network \\ 
    GCN & Graph Convolutional Network \\ 
    GNN	& Graph Neural Network  \\
    GPT3 & Third generation Generative Pre-trained Transformer \\
    IJCAI & International Joint Conference on Artificial Intelligence \\ 
    ILP & Inductive Logic Programming \\
    KG & Knowledge Graphs \\
    KGC & Knowledge Graph Completion \\ 
    KGQA & Knowledge Graph Question Answering \\ 
    KR & Knowledge Representation \\ 
    KRR & Knowledge Representation \& Reasoning \\ 
    LNN & Logical Neural Networks \\ 
    LLM	&Large Language Models \\
    LSTM & Long Short Term Memory \\ 
    LTN	&Logic Tenson Network \\
    ML	& Machine Learning \\
    MLN & Markov Logic Network \\ 
    MLP & Multilayer Perceptron\\
    MWP & Math Word Problem \\
    NE	& Neuroevolution  \\
    NeSy &	Neuro-Symbolic AI  \\
    NL & Natural Logic \\
    NLI & Natural Language Inference \\ 
    NLG & Natural Language Generation \\ 
    NLM & Neural Logic Machine \\
    NLP	&Natural Language Processing \\
    NLU &	Natural Language Understanding \\
    NS-CL & Neuro-Symbolic Concept Learner \\
    NTP & Neural Theorem Prover \\ 
    NN	&Neural Network \\
    OOD	& Out-of-distribution \\
    OOP	& Object-oriented programming(paradigm) \\
    OWL & Web Ontology Language \\ 
    ProbLog & Probabilistic Logic Programming  \\
    RcNN&	Recursive Neural Network \\
    RL	&Reinforcement Learning \\
    RNKN&	Recursive Neural Knowledge Network \\
    RNN	&Recurrent Neural Network \\
    SOTA & State of the Art \\
    SVM	& Support Vector Machine \\
    TPR	& Tensor Product Representation \\
    TSP &  Traveling Salesperson Problem \\
    (\(\partial\)ILP) & Differentiable Inductive Logic Programming \\
    \bottomrule
    \end{tabular}
\end{table*}

\end{appendix}

%

\end{document}